\documentclass[10pt,journal,compsoc]{IEEEtran}

\usepackage{multirow}
\usepackage{xcolor}
\usepackage{amsmath,amssymb,amsfonts}
\usepackage{textcomp}
\usepackage{manyfoot}
\usepackage{booktabs}
\usepackage{listings}
\usepackage{graphicx}
\usepackage{adjustbox}
\usepackage{makecell}
\usepackage{array}
\usepackage[nocompress]{cite}
\usepackage[pagebackref,breaklinks,colorlinks]{hyperref}
\usepackage{enumitem}

\begin{document}

\title{Survey on video anomaly detection in dynamic scenes with moving cameras}

\author{Runyu Jiao, Yi Wan, Fabio Poiesi, Yiming Wang
\IEEEcompsocitemizethanks{
\IEEEcompsocthanksitem Runyu Jiao and Yi Wan are with Key Laboratory of High Efficiency and Clean Manufacturing, School of Mechanical Engineering, Shandong University, Jinan, China. E-mails: $<$jiao,wanyi$>$@mail.sdu.edu.cn.
\IEEEcompsocthanksitem Fabio Poiesi and Yiming Wang are with Fondazione Bruno Kessler, Trento, Italy. E-mails: $<$poiesi,ywang$>$@fbk.eu.
}
\thanks{
This work was supported by the National Natural Science Foundation of China (No.~51975336), Key Technology R\&D Program of Shandong (No.~2020JMRH0202, No.~2022CXGC020701), Major Industrial Research Projects in Shandong Province for the Transition to New from Old Economic Engines (2021-13). 
We also acknowledge the support of the PNRR ICSC National Research Center for High Performance Computing, Big Data and Quantum Computing (CN00000013), under the NRRP MUR program funded by the NextGenerationEU.
This work was supported by the PNRR project FAIR - Future AI Research (PE00000013), under the NRRP MUR program funded by the NextGenerationEU.}
}

\IEEEtitleabstractindextext{%
\begin{abstract}
The increasing popularity of compact and inexpensive cameras, e.g.~dash cameras, body cameras, and cameras equipped on robots, has sparked a growing interest in detecting anomalies within dynamic scenes recorded by moving cameras. However, existing reviews primarily concentrate on Video Anomaly Detection (VAD) methods assuming static cameras.
The VAD literature with moving cameras remains fragmented, lacking comprehensive reviews to date. To address this gap, we endeavor to present the first comprehensive survey on Moving Camera Video Anomaly Detection (MC-VAD). We delve into the research papers related to MC-VAD, critically assessing their limitations and highlighting associated challenges. Our exploration encompasses three application domains: security, urban transportation, and marine environments, which in turn cover six specific tasks. We compile an extensive list of 25 publicly-available datasets spanning four distinct environments: underwater, water surface, ground, and aerial. We summarize the types of anomalies these datasets correspond to or contain, and present five main categories of approaches for detecting such anomalies. Lastly, we identify future research directions and discuss novel contributions that could advance the field of MC-VAD. With this survey, we aim to offer a valuable reference for researchers and practitioners striving to develop and advance state-of-the-art MC-VAD methods.
\end{abstract}

}

\maketitle

\IEEEdisplaynontitleabstractindextext

\IEEEpeerreviewmaketitle

\IEEEraisesectionheading{\section{Introduction}\label{sec:introduction}}

\IEEEPARstart{V}{ideo} anomaly detection (VAD) aims to locate unusual activities or behaviors from normal or expected ones, within a video frame as well as in the sequence of video frames.
Detecting anomalies in the real world is a complex and challenging task. 
One challenge is that there exists no clear and unified definition of anomalies of interest. 
The definition greatly depends on the application context, where the normal activities in one context may be deemed abnormal in another context~\cite{surveypami}. 
For example, the circulation of cars and bikes on roads is normal~\cite{DAD}, while it would be abnormal in pedestrian zones~\cite{UCSD}. 
Activities like eating, drinking, or exercising are abnormal in an office setting but may be entirely normal in others~\cite{ICRA17}. 
Another challenge is due to the rare and sporadic occurrences of such anomalies~\cite{zaheer2022generative}. 
This makes the collection of large and well-curated anomaly videos a very demanding task, thus limiting the learning of anomalous patterns~\cite{UCF}.
Existing works often formulate VAD as an out-of-distribution detection or one-class classification problem, and address it by modeling the normal patterns using a large amount of normal videos (i.e.~the unsupervised setting)~\cite{7780455, liu2018ffp, ShanghaiTech, Ravanbakhsh2017icip, Sabokrou2017tip, Xia2015iccv, Zaigham2020cvpr, zhang2016video}, together with a limited amount of abnormal videos with either video-level supervision (i.e.~the semi-supervised setting)~\cite{liu2019completeness,liu2019weakly,narayan20193c,shou2018autoloc,wang2017untrimmednets,yu2019temporal,tian2021weakly,purwanto2021dance,zaheer2020claws,zaheer2020self}, or frame-level supervision (i.e.~the supervised setting) thanks to synthetic data generation~\cite{ubnormal}.
Significant progress has been made in Video Anomaly Detection (VAD) for \textit{static cameras} (SC-VAD) within the application domain of video surveillance.
Benchmark datasets, including UCSD Anomaly Detection \cite{UCSD}, ShanghaiTech Campus \cite{ShanghaiTech}, and CUHK Avenue \cite{CUHK}, encompass videos captured in urban environments that present a range of anomalies. 
These datasets pose challenges under different supervision settings, as discussed earlier, while also accounting for minor camera motion resulting from pan-tilt-zoom operations.
There exist large datasets that feature anomalies in scenes captured in diverse environments, such as UCF-Crime that is composed of long-hour real-world surveillance videos with 13 anomaly types~\cite{UCF}, and UBnormal that is composed of synthetically-generated videos with 22 anomaly types~\cite{ubnormal}.
Surveys on SC-VAD have been conducted, categorizing the methods into two main areas: \textit{single-scene} anomaly detection, and \textit{multi-scene} anomaly detection \cite{surveypami, survey2021multimedia}. 
Single-scene methods focus on detecting anomalies within videos where the scene remains constant throughout \cite{UCSD, raghavendra2006unusual, CUHK, Subway, ramachandra2020street, Poiesi2015}. 
Multi-scene methods aim to detect anomalies across different scenes by utilizing videos captured from multiple static cameras \cite{UCF, ShanghaiTech}.

\begin{figure*}[t]
    \centering
    \includegraphics[width=.9\textwidth]{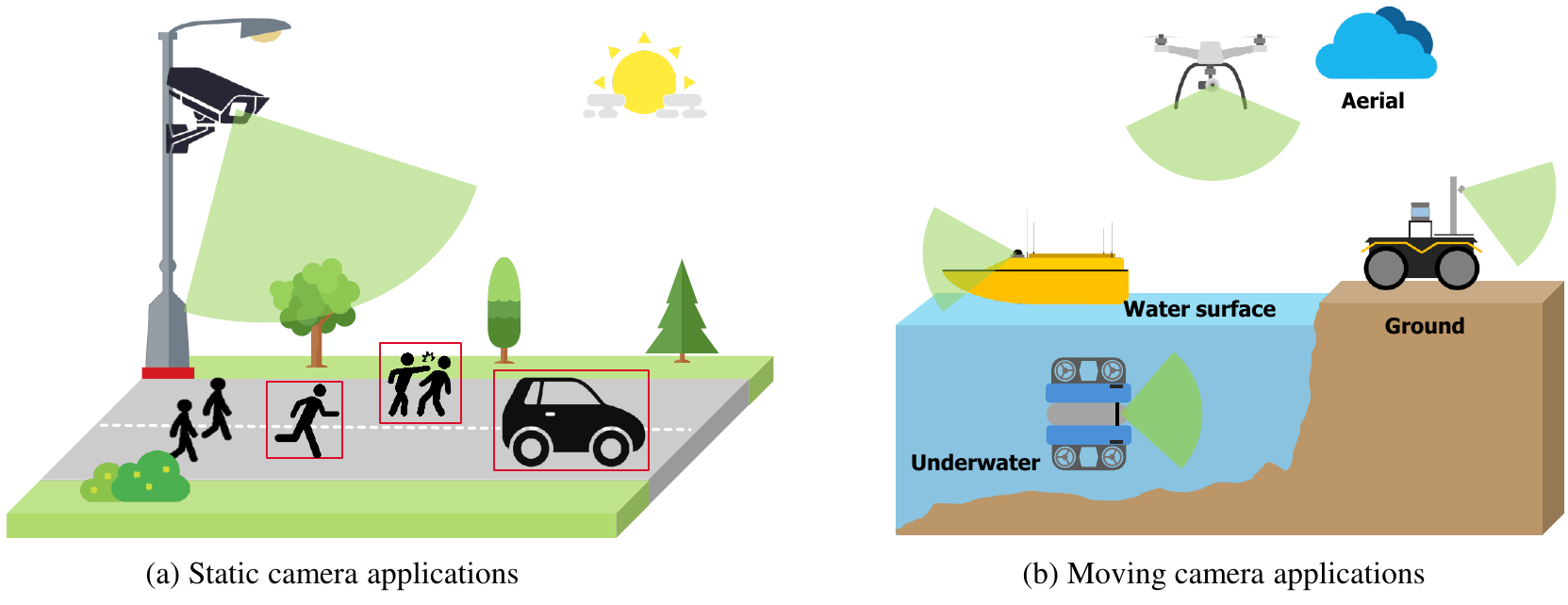}
    \vspace{-4mm}
    \caption{Difference between (a) static camera applications and (b) moving camera applications. We review papers that tackle video anomaly detection in videos captured from moving cameras.
    Unlike static camera algorithms, moving camera algorithms have to handle situations where both background and foreground dynamically change over time.}
    \label{MCVAD}
\end{figure*}

Unlike previously mentioned surveys on SC-VAD, we focus on presenting the first comprehensive survey on video anomaly detection in dynamic scenes captured by \emph{moving cameras}, which we refer to as Moving Camera Video Anomaly Detection (MC-VAD) (Fig.~\ref{MCVAD}).
By moving cameras, we mean cameras that are mounted on mobile platforms, such as unmanned aerial vehicles (UAVs)~\cite{minidrone,li2021uav,seagull,seadronesee}, ground robots~\cite{chakravarty2007anomaly, gehring2021anymal}, road vehicles \cite{pinggera2016lost, kragh2017fieldsafe, singh2020lidar, roadanomal}, water surface vehicles~\cite{SMD,MODS2022}, or underwater vehicles~\cite{UUD,zhou2022discovering}.
As the types of moving cameras expose a large diversity, MC-VAD naturally spans a much larger number of application domains as compared to SC-VAD.
In this survey, we identify three broad application domains including \emph{security}, \emph{urban transportation}, and \emph{marine environments}, covering various anomalies, such as traffic accidents~\cite{DAD,HEVI,A3D,dada2000,CCD,DoTA}, criminal activities~\cite{Eyeinsky2018,ipatch}, or hazardous actions/objects~\cite{HTA,TIP2010}.

In MC-VAD, the camera is no longer a neutral ``observer'' (as in SC-AVD), but it can be a potential anomaly ``participator''. 
Such enrichment in the camera role introduces a new range of anomalies that do not exist in SC-VAD.
For example, the vehicle that equips the camera can participate in car accidents either actively (i.e.~the responsible) or passively (i.e.~the involved)~\cite{DoTA}.
Moreover, any objects or contexts that might cause abrupt motion deviation of the equipping platforms, such as drones and unmanned vehicles should also be considered anomalies, as they pose a potential threat to the mobile platforms or other traffic participants. 
Such new anomaly definition leads to the inclusion of works related to obstacle detection for safe autonomous navigation, such as early collision detection for mobile robots~\cite{poiesi2016detection}, obstacle detection for unmanned surface vehicles (USVs)~\cite{kristan2015fast,modd2}, and detection of hazardous environments for mobile robots~\cite{mantegazza2022challenges}. 
In total, we identify six specific application tasks relevant to MC-VAD, including \emph{static object detection}, \emph{dynamic objects and behaviors analysis}, \emph{traffic accident detection}, \emph{obstacle detection}, \emph{environmental surveillance}, and \emph{underwater exploration}.

In the following, we first discuss surveys related to the topic of VAD in Sec.~\ref{sec:related_surveys}, followed by Sec.~\ref{sec:lit_stats} with the statistics of the research papers covered in this survey in terms of the method for literature searching, and their distribution over publication venues and publication years. 
Lastly, Sec.~\ref{sec:motivation} motivates the need for this survey.

\subsection{Related surveys}\label{sec:related_surveys}

Because there are no specific surveys about MC-VAD in the literature, we cover recent surveys on the topic of VAD~\cite{deepvad2021, surveyAD, surveypami, AIreview22, survey2021multimedia}, as well as related surveys covering topics about abnormal human behavior understanding~\cite{popoola2012video, Spagnolo2014, Poiesi2015, surveyHAR} and hazard detection in autonomous driving~\cite{10038646}. 
Tab.~\ref{tab0} summarizes the surveys covered in this section.

\begin{table*}
\caption{List of recent surveys related to multi-camera video anomaly detection approaches.
The categories are arranged based on the scope and relevance within each domain. 
Within each category, the surveys are sorted chronologically.}
\label{tab0}
\vspace{-3mm}
\begin{tabular}{lllp{11cm}}
\toprule
Survey & Category & Year & Content\\
\midrule

\cite{surveyAD}  & Anomaly detection  & 2021  & Deep learning for anomaly detection, covering key intuitions, objective functions, underlying assumptions, challenges, and future opportunities \\

\cite{survey2021multimedia} & Multimedia anomaly detection & 2021 & Multimedia datasets for anomaly detection, covering a variety of video, audio, and audio-visual datasets \\

\cite{deepvad2021} & Video anomaly detection & 2021  & Deep learning for video anomaly detection, discussing future opportunities in various domains and challenges \\

\cite{surveypami} & Video anomaly detection & 2022 & Single-scene video anomaly detection in surveillance, focusing on problem formulations, datasets, evaluation criteria, and past research \\

\cite{AIreview22} & Video anomaly detection & 2022 & Anomaly detection in surveillance videos, including a thematic taxonomy of deep models, review and performance analysis  \\

\cite{liu2023generalized} & Video anomaly detection & 2023 & Generalized video anomaly event detection with systematic taxonomy and comparison of deep models \\

\cite{ullah2023comprehensive} & Violence detection & 2023 &Vision-based violence detection in surveillance videos, covering new taxonomy, working flow of the methods, datasets, challenges and future directions\\

\cite{mumtaz2023overview} & Violence detection & 2023 & Violence detection in surveillance videos, including achievements, datasets, performance evaluation, challenges, and future directions\\

\cite{popoola2012video} & Human behavior recognition & 2012 & Video-based scene contextual understanding and abnormal human behavior detection in surveillance applications \\

\cite{gowsikhaa2014automated} & Human behavior analysis & 2014  & Human behavior analysis in surveillance videos, covering related background modeling, foreground detection, object classification, motion tracking, and behavior analysis methods \\

\cite{Poiesi2015} & Human interactions recognition   & 2015 & Human interactions prediction and recognition in public spaces, covering taxonomy and extensive survey of methods\\

\cite{surveyoldpeople} & Human activity recognition & 2020  & Non-intrusive human activity recognition and abnormal behavior detection on elderly people\\

\cite{surveyHAR} & Human action recognition  & 2021 & Video-based Human Action Recognition (HAR), including machine learning and deep learning techniques for HAR, datasets, and applications \\

\cite{santhosh2020anomaly} & Traffic anomaly detection  & 2020 & Anomaly detection in road traffic using visual surveillance, covering problem statements, learning methods, datasets, and applications  \\

\cite{breitenstein2021corner} & Autonomous driving & 2021 & Corner cases for visual perception in automated driving, including systematic classification and detection methods \\

\cite{10038646} & Autonomous driving & 2023  & Graph-based hazardous event detection methods for autonomous driving\\

\cite{nguyen2022state} & Aerial surveillance & 2022 & Human-centric aerial surveillance tasks such as detection, tracking, identification, and action recognition using drones, UAVs, and other airborne platforms\\

\bottomrule
\end{tabular}
\end{table*}

VAD-related surveys focus on describing applications~\cite{deepvad2021}, categorizing methods~\cite{surveyAD, surveypami, AIreview22}, or organizing publicly available datasets~\cite{survey2021multimedia}.
The survey by Ren et al.~\cite{deepvad2021} reviews challenges and application opportunities in the domains of intelligent transportation, digital education, smart home, public health, and digital twins. 
While intelligent transportation is an important application area for MC-VAD, this survey lacks details in many aspects including datasets, approaches, and challenges that are specific to each application.
The survey by Pang et al.~\cite{surveyAD} focuses on reviewing deep learning-based VAD methods, formulating a comprehensive taxonomy and discussing key intuitions, objective functions, underlying assumptions, advantages, and disadvantages.
The survey by Chandrakala et al.~\cite{AIreview22} is structured similarly to that of \cite{surveyAD}, with an updated coverage on more recent deep learning-based VAD methods.
The survey by Ramachandra et al.~\cite{surveypami} first assesses the differences between anomalies captured in single-scene and multi-scene setups, and then focuses on the research progress of single-scene VAD methods.
This survey also includes a comprehensive quantitative experimental comparison of several VAD algorithms on popular benchmarks.
Lastly, the survey by Kumari et al.~\cite{survey2021multimedia} focuses on multimedia datasets for the study of VAD.
Datasets are divided into video anomaly datasets, audio anomaly datasets, and audio-visual anomaly datasets.
For video anomaly datasets, the authors listed 38 publicly available video datasets published between 2007 and 2021. 
While this list is comprehensive for the SC-VAD problem, only two datasets involve moving cameras.
In general, the above-mentioned surveys are primarily for VAD with static cameras, there is no dedicated analysis of specific challenges for VAD with moving cameras.

We also consider the literature about human activity recognition (HAR), or human abnormal behavior detection, as relevant to VAD, because human activities are often present and analyzed for anomaly detection.
For example, the survey by Popola and Wang~\cite{popoola2012video} focuses on video-based abnormal human behavior recognition. 
This survey starts with an analysis of human motion and then discusses the contextual characteristics of anomalies, scene modeling, and behavior abstraction and representation. It also argues that the definition of anomaly can be often ambiguous, which is similar to the context of MC-VAD.
The survey by Poiesi and Cavallaro~\cite{Poiesi2015} reviews methods for the analysis of human interactions in public places, with particular focus on the prediction of rendezvous areas where people are likely to interact with each other or with static objects, e.g.~a door or a meeting point. Such methods can be relevant for analyzing anomalies in crowded scenes.
The survey by Pareek and Thakkar~\cite{surveyHAR} discusses a general framework for action recognition tasks comprising of feature extraction, feature encoding, dimensionality reduction, and action classification.
None of these HAR surveys specifically address the VAD problem from moving cameras.
The survey by Xiao et al.~\cite{10038646} provides a comprehensive analysis of hazardous event analysis for autonomous driving scenarios.
In particular, it provides an overview of graph-based methods, and summarizes the related datasets and evaluation metrics.
However, their discussion does not cover the VAD problem of moving cameras.

To the best of our knowledge, ours is the first survey on the topic of MC-VAD.
We define the concept of MC-VAD, which builds upon the fundamental principles of other surveys regarding VAD, and discuss the challenges posed by camera motion.
Our survey provides an overview of research progress regarding MC-VAD ranging from 2001 to 2023, covering six tasks across three broad application domains. 
We provide an organized summary of 25 publicly available datasets collected from moving cameras for the study of MC-VAD. 
We also organize related methods addressing MC-VAD into five categories with a discussion of their respective advantages and limitations.

\begin{figure*}[t]
    \centering
    \includegraphics[width=.95\textwidth]{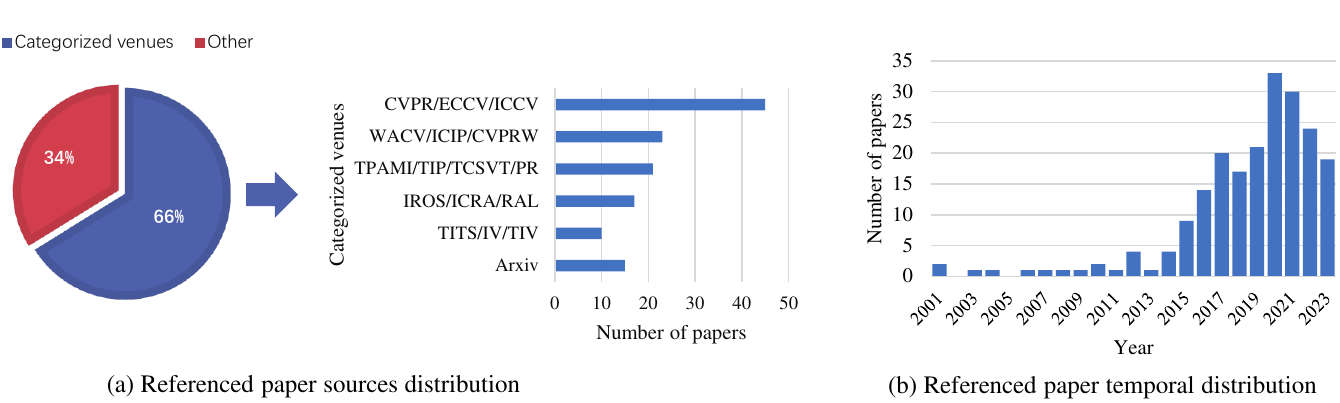}
    \vspace{-3mm}
    \caption{Overview of MC-VAD publication statistics.
    (a) Publication sources can be divided into six categories. 
    Vision conferences: 
    CVPR (Computer Vision and Pattern Recognition), 
    ECCV (European Conference on Computer Vision), 
    ICCV (International Conference of Computer Vision), 
    WACV (Winter Conference on Applications of Computer Vision), 
    ICIP (International Conference of Image Processing),
    CVPRW (Computer Vision and Pattern Recognition Workshops). 
    Vision journals: 
    TPAMI (Transactions on Pattern Analysis and Machine Intelligence),
    TIP (Transactions on Image Processing),
    TCSVT (Transactions on Circuits and Systems for Video Technology),
    PR (Pattern Recognition).
    Robotics conferences and journals:
    IROS (international conference on Intelligent Robots and Systems),
    ICRA (International Conference on Robotics and Automation),
    RAL (Robotics and Automation Letters).
    Transportation and vehicle conferences and journals:
    TITS (Transactions on Intelligent Transportation Systems),
    IV (Intelligent Vehicles Symposium),
    TIV (Transactions on Intelligent Vehicles).
    (b) Histogram of discussed papers over the publication year.}
    \label{statistic1}
\end{figure*}

\subsection{Literature statistics}\label{sec:lit_stats}

We conducted an extensive search for papers within the context of MC-VAD using Google Search, Google Scholar, Web of Science, and ScienceDirect platforms.
We used Papers With Code, Google Dataset Search, and Kaggle platforms to collect relevant datasets.
Given the absence of a clear definition of MC-VAD literature, and considering the plethora of related topics that often use different terminology, such as hazard detection, violence detection, obstacle detection, anomaly recognition, among others, it can be challenging for traditional database search strategies to capture all relevant content.
Therefore, we performed multiple rounds of searches using different keywords, aiming to filter and select works closely aligned with MC-VAD based on our definition. 
We also used citations of each paper as a way to find research works advancing the underlying literature.
Lastly, we counted the academic papers cited in this survey after an iterative screening process following the practice in the surveys of \cite{ullah2023comprehensive,mumtaz2023overview}.

Fig.~\ref{statistic1} provides an overview of the MC-VAD publication statistics, reporting the publication sources and temporal distributions of the research works.
Similarly to \cite{khan2019survey}, these statistics are compiled based on their publication venues.
Fig.~\ref{statistic1}a shows that most works related to MC-VAD have been published in computer vision and image processing conferences such as CVPR, ICCV, ECCV, WACV, ICIP, and their associated workshops.
We also found contributions in image processing and pattern recognition journals, such as TPAMI, TIP, TCSVT, and PR.
Because MC-VAD involves mobile robotic platforms, and finds wide applications in the fields of robotics and intelligent transportation, a portion of the research work also comes from robotics conferences and journals such as IROS, ICRA, and RAL, as well as conferences and journals in the smart transportation and vehicle research domain such as IEEE TITS, IV, and TIV.

Fig.~\ref{statistic1}b shows the number of selected papers per year from 2001 to 2023. 
VAD research using mobile cameras can be traced back as early as 2001.
However, due to the lack of large-scale datasets and effective data-driven techniques, only a few studies were conducted in this area prior to 2012.
The growing adoption of moving cameras has boosted the collection of large-scale datasets, which in turn has sped up the development of MC-VAD over the past decade.
Yet, we also observe a decrease in publications between 2020 and 2023, the reasons for which are challenging to pinpoint.
One possibility could be a shift in research interest, likely influenced by various factors such as new emerging technologies or funding priorities due to urging societal need, e.g. the pandemic.
Also, the increasing concern on data privacy and AI ethics may also slow down the research progress of MC-VAD as the relevant datasets and method development requires more regulatory and ethical considerations. 

\begin{figure}[t]
    \centering
    \includegraphics[width=1\columnwidth]{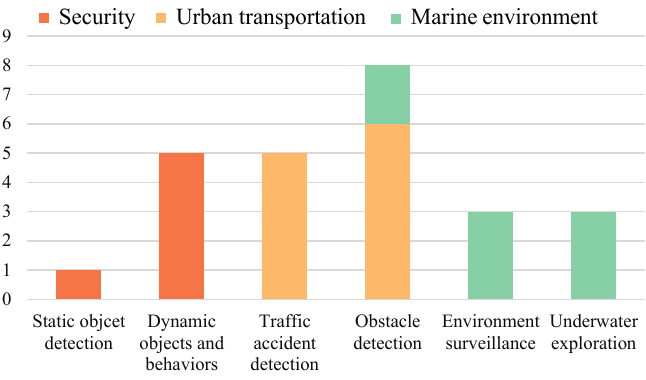}
    \vspace{-7mm}
    \caption{Distribution of public MC-VAD datasets (25 in total) over six tasks across three application domains: static object detection, dynamic objects and behaviors, traffic accident detection, obstacle detection, environment surveillance and underwater exploration.
    }
    \label{statistic2}
\end{figure}

Fig.~\ref{statistic2} presents the number of publicly available datasets per task across three application domains.
Within the security domain, public datasets for static object detection are few.
This scarcity is likely due to the niche nature of scenarios typically encountered in this domain, such as indoor industrial settings, which requires ad-hoc methods for effective resolution. 
In contrast, datasets for studying the detection of dynamic objects and behaviors are more prevalent, likely attributed to the growing interest in research about to human behavior understanding.
Within the urban transportation domain, there are abundant datasets concentrating on obstacle detection and traffic accident detection. 
This trend can be primarily linked to the surging interest in autonomous driving and mobile robot navigation.
Lastly, within the marine domain, the growing interest in the development of techniques for obstacle detection, environmental monitoring, and underwater exploration has led to the creation of relevant datasets.

\subsection{Motivation}\label{sec:motivation}

Our analysis of the related literature shows that there is a growing interest in the problem of VAD with moving cameras.
However, no comprehensive literature organization exists due to the high diversity of the camera-equipped platforms and the application domains.
We are thus motivated to compile this survey in order to bridge this gap and provide insights to both interested readers and domain experts.
Our survey begins by delineating a comprehensive taxonomy, reflecting the current research \textit{status quo} in the realm of VAD with moving cameras.
We proceed by identifying potential application domains and discussing the unique technical challenges each domain presents.
We then present and categorize an extensive list of datasets based on the mobile platforms equipped with cameras.
These datasets are either specifically designed for or have the potential to be utilized for MC-VAD research.
We also classify existing methods according to the nature of the anomalies they address, followed by a detailed discourse on their respective advantages and limitations.
Lastly, we conclude this survey by offering recommendations for future research directions.
In summary, the main contributions of our survey are:

\begin{itemize}

\item[\textbullet]\textit{Problem definition and new challenges}.
We define the problem and scope of MC-VAD and we identify a thorough list of novel challenges introduced by the use of moving cameras.

\item[\textbullet]\textit{Comprehensive literature review.}
We review a significant number of research studies from leading conferences and journals in the fields of computer vision, artificial intelligence, and robotics. Such comprehensive coverage presents a complete picture of the research progress regarding MC-VAD in the relevant application domains.

\item[\textbullet]\textit{Thorough dataset summary.} 
We identify and summarize 25 publicly available VAD datasets captured from moving cameras that are suitable for the study of MC-VAD. 
We provide detailed descriptions for these datasets, covering aspects such as the application context, number of videos/anomalies, typical anomalies, participants, and their advantages and disadvantages. 
We also provide their respective download links to directly access them.

\item[\textbullet]\textit{Categorized anomaly and methods.} 
We categorize the anomalies present in MC-VAD into five distinct categories and present a taxonomy to classify methods addressing MC-VAD accordingly.  

\item[\textbullet]\textit{Insightful outlook.}
We outline the limitations of existing work and discuss promising future directions to tackle MC-VAD-related challenges.

\end{itemize}

\begin{figure}[t]
    \centering
    \includegraphics[width=1\columnwidth]{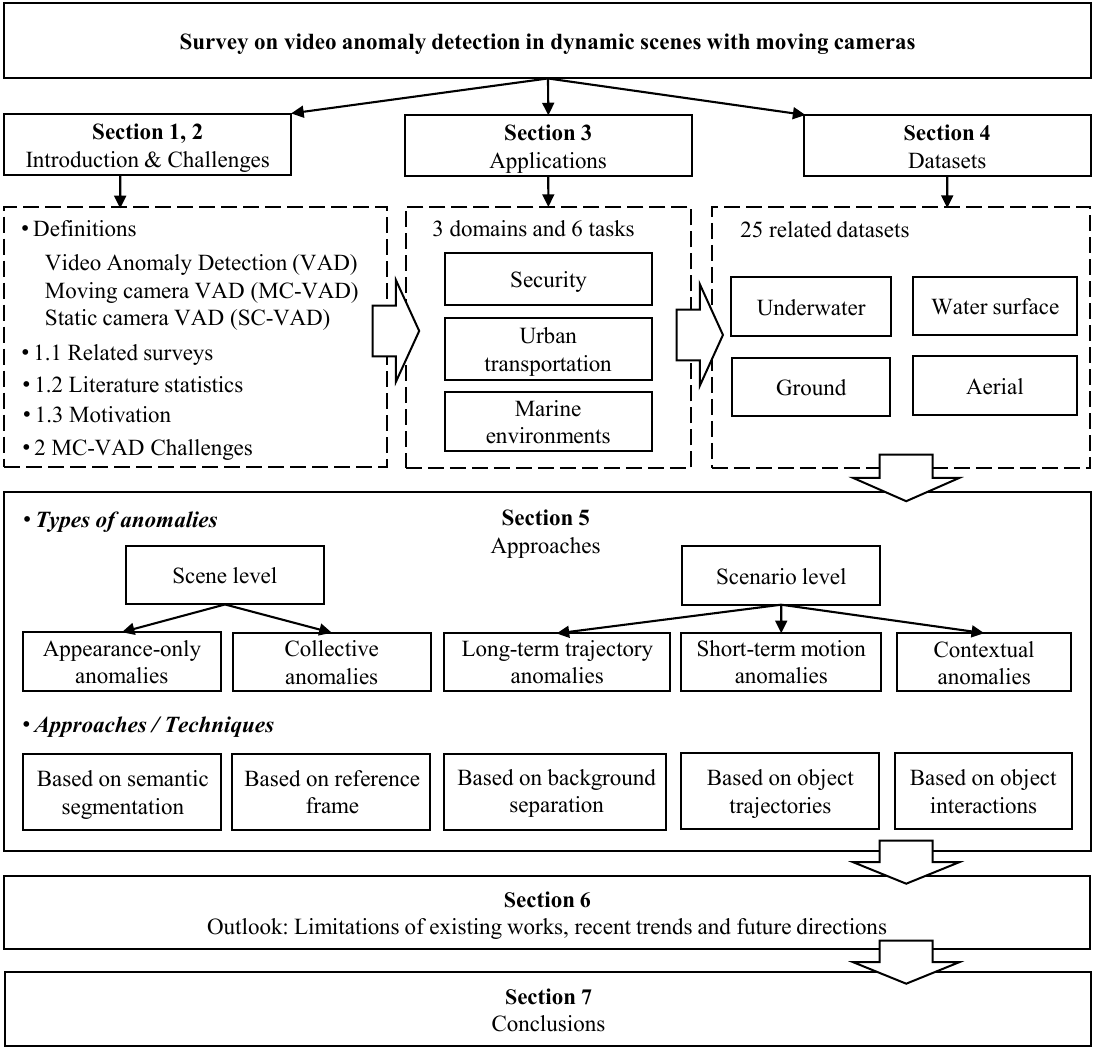}
    \vspace{-6mm}
    \caption{
    Survey overview. 
    Sec.~\ref{sec:introduction} introduces and defines the Video Anomaly Detection (VAD) problem, and relates the study of Moving Camera VAD (MC-VAD) with respect to Static Camera VAD (SC-VAD). 
    We position our survey among related surveys and compile literature statistics.
    Sec.~\ref{sec:challenges} lists MC-VAD-related challenges.
    Sec.~\ref{sec:Applications} presents three application areas of MC-VAD, i.e.~security, urban transportation, and marine environment, as well as six specific tasks across different domains. 
    Sec.~\ref{sec:Datasets} presents 25 related datasets, spanning four environments including underwater, water surface, ground, and aerial.
    Sec.~\ref{sec:Approaches} classifies the anomalies contained in the datasets into two levels, i.e.~scene level and scenario level, and five types of anomalies.
    We further categorize MC-VAD approaches into five classes.
    Sec.~\ref{sec:outlook} presents limitations of existing works and future directions.
    Sec.~\ref{sec:conclusions} concludes our survey.
    }
    \label{overview}
\end{figure}

Fig.~\ref{overview} shows the overall paper organization.
Sec.~\ref{sec:challenges} discusses the new challenges that camera motion poses for VAD. 
Sec.~\ref{sec:Applications} describes the primary applications of MC-VAD. 
Sec.~\ref{sec:Datasets} details the publicly available datasets for studying MC-VAD, with relevance to the applications discussed in Sec.~\ref{sec:Applications}. 
In Sec.~\ref{sec:Approaches}, we first classify the types of anomalies present in the dataset and then present a taxonomy for categorizing various MC-VAD approaches.
In Sec.~\ref{sec:outlook}, we discuss the limitations of existing methods and delve into potential future research directions.
Lastly, in Sec.~\ref{sec:conclusions}, we draw our conclusions.

\section{MC-VAD challenges}\label{sec:challenges}

Recent deep learning-based methods have shown compelling results on public SC-VAD datasets, such as ShanghaiTech~\cite{ShanghaiTech}.
However, MC-VAD still remains a challenging problem due to the presence of camera ego-motion. The broader range of applications makes generalized solutions more challenging to emerge. 
In this section, we discuss MC-VAD challenges in detail.

\noindent \textbf{Ego-motion.}
The optical flow captured by moving cameras entangles both the motion of scene objects and the camera ego-motion. 
This makes the task of detecting anomalies based on abnormal motion patterns more challenging. 
To address this issue, camera motion can be estimated and compensated for \cite{poiesi2016detection}.
Early works focused on studying MC-VAD consider simple camera motions, such as a camera steadily moving along a fixed trajectory~\cite{VDAO,TIP2010}.
Such limited motion can exploit planar homography to account for the camera motion. However, it is still challenging to handle scenarios with unconstrained camera motions, such as high speeds, and sudden direction variations~\cite{mur2017orb,D3vo,9741458}.

\begin{figure*}[t]
    \centering
    \includegraphics[width=1\textwidth]{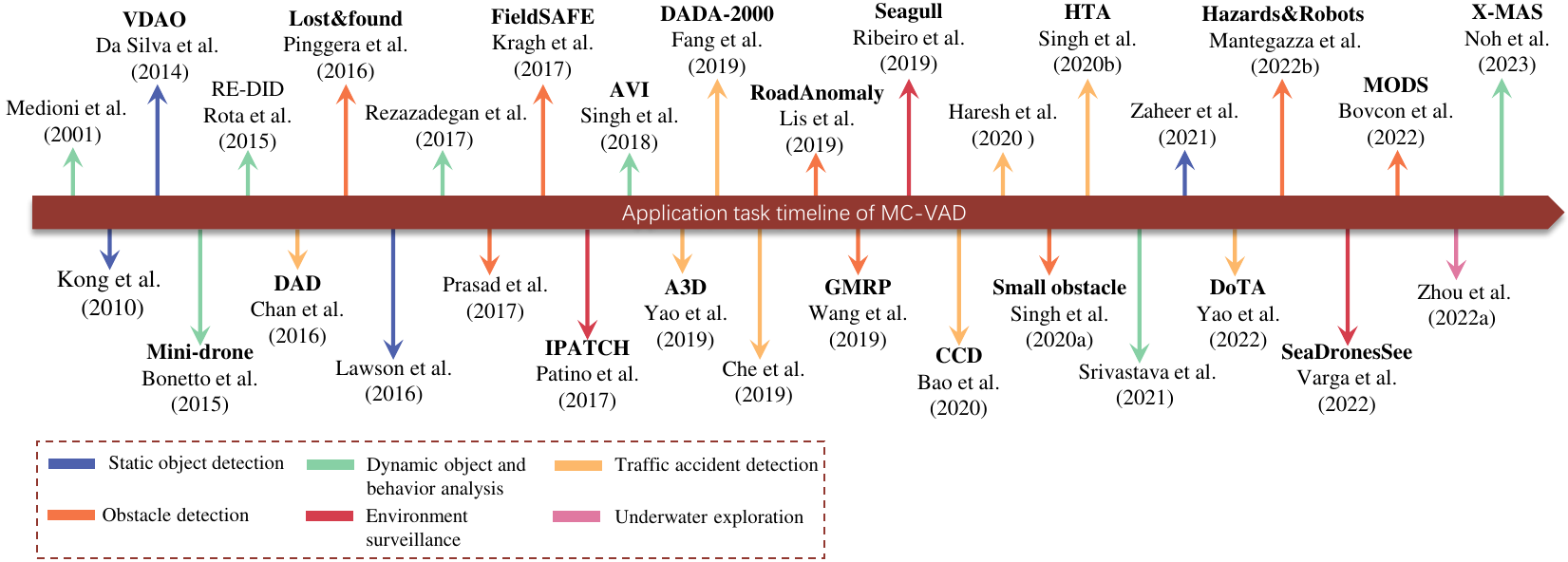}
    \vspace{-6mm}
    \caption{
    Timeline of relevant research works covering six main categories of tasks in MC-VAD, from 2001 to 2023. 
    Entries in \textbf{bold} highlight the datasets that were proposed for each application.
    Early works were predominantly in the security sector, focusing on both static object detection and dynamic object and behavior analysis, a trend that also continues today. 
    MC-VAD has later emerged in the field of urban transportation, specifically in traffic accident detection and obstacle detection, since 2016, with a noticeable uprise around 2019.
    Lastly, the applications of MC-VAD in the field of marine environments started in 2017 and have recently gained more research attention.}
    \label{timeline}
\end{figure*}

\noindent \textbf{Changing background.}
Numerous studies have shown evidence that the majority of VAD algorithms heavily rely on background cues \cite{backgroundbias,zhou2019attention}. This makes MC-VAD with changing background a challenging task due to the dynamics in the background and camera ego-motion.
Methods that rely on a static background \cite{ICRA17,chapel2020moving,erez2022deep} are highly likely to fail. Thus, how to limit the impact from the background remains one of the challenges in the study of MC-VAD. 

\noindent \textbf{Presence of abrupt motion.}
Abrupt changes in the camera ego-motion in terms of speed and direction present a challenge that is unique to MC-VAD.
Even in fixed surveillance datasets such as UCF-Crime~\cite{UCF}, the presence of camera shake or switch in videos can make most existing SC-VAD methods based on frame prediction or frame reconstruction ineffective.
This situation becomes more prevalent in videos captured by moving cameras, where abrupt motion can occur in many unpredictable manners, being normal or abnormal. While ad-hoc solutions exist in the literature to address particular scenarios, it is challenging to obtain generalizable solutions to handle such abrupt camera motion. 

\noindent \textbf{First-person perspective.}
The shift in viewpoint from a third-person perspective, as in SC-VAD, to a first-person perspective as in MC-VAD introduces new challenges to MC-VAD, in particular for the patterns of occlusions.
In SC-VAD, surveillance cameras are typically installed in locations with optimal views for better coverage of the scene. Instead, in the case of first-person view from a moving camera, occlusions occur more frequently and severely as compared to the third-person view in the case of SC-VAD, making the video analysis in MC-VAD more challenging.

\noindent \textbf{Dynamic context priors.}
There can be rich context priors exploited for VAD, such as jaywalking, riding a bike on a pedestrian sidewalk, or driving a car in the wrong direction~\cite{surveypami}. Such context prior can be easily extracted in SC-VAD as the scene has a static and a known structure. However, in MC-VAD, the scene is continuously changing due to camera motion, thus requiring methods that are capable of understanding scene semantics regardless of the viewpoint. 

\noindent \textbf{Varying anomaly definition.}
As the scene context changes with camera motion, the definition of normality and anomaly also changes accordingly. 
For instance, for a mobile robot that patrols an office area, actions such as eating food or holding a knife might be recognized as anomalies, while such actions in a kitchen would be deemed as normal activities. 
Such varying anomaly definition requires context-aware modeling in order to account for the context changes in dynamic scenes.

\noindent \textbf{Low-quality data.}
It is also noticeable that current datasets for the study of MC-VAD have limited data quality. A large proportion of MC-VAD datasets are acquired through crowdsourcing \cite{RE-DID, DAD, UCF, A3D, dada2000, HTA, CCD, DoTA}, and the quality is limited by factors such as camera hardware, installation conditions, compression techniques, camera motion, and external environmental conditions, leading to low resolution, blurriness, and sometimes significant noises. Such datasets increase the difficulty of detecting anomaly detection, hence multimodal techniques might be helpful to address such challenge~\cite{SMD, kragh2017fieldsafe, GMRPD, seadronesee, XMAS}.

\section{Applications}\label{sec:Applications}

\begin{figure*}[t]
\centering
\includegraphics[width=1\textwidth]{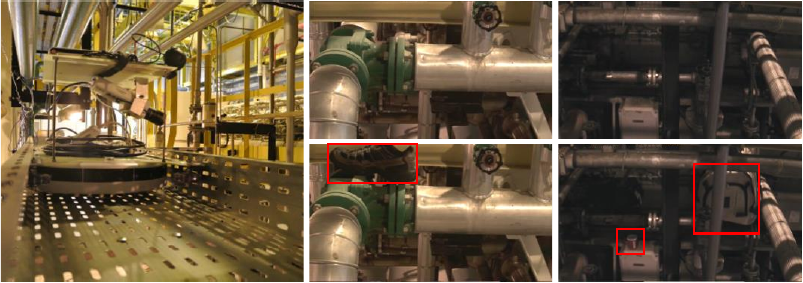}
\vspace{-6mm}
\caption{Mobile camera platform for abnormal object detection in industrial scenes~\cite{VDAO}. 
It captures both reference videos featuring normal scenes and target videos featuring abnormal scenes with the presence of unusual objects, such as shoes, cups, and bags (highlighted in red boxes).}
\label{VDAO}
\end{figure*}

MC-VAD is of importance to several application domains, including security, urban transportation, and marine environments. 
Among these domains, we identify six application tasks.
For security, the tasks are \textit{static object detection}, and \textit{dynamic object and behavior analysis}.
For urban transportation, the tasks are \textit{traffic anomaly detection} and \textit{obstacle detection}.
For marine environments, the tasks are \textit{obstacle detection}, \textit{environmental surveillance}, and \textit{underwater exploration}.
In Fig.~\ref{timeline}, we present a timeline of relevant research works that were published between 2001 and 2023. 
In bold we highlight datasets that were introduced along with their published work.
We can observe that methods published before 2016 are predominantly in the security sector, focusing on both static object detection, and dynamic object and behavior analysis. 
Since 2016, works in the field of urban transportation have emerged with a noticeable uprise starting from 2019. The application of MC-VAD in the field of marine environments started in 2017 and has gradually garnered more attention in recent years.
In the following sections, we dive into each application domain with thorough coverage of each detailed task.

\subsection{Security}

Early research on VAD within the field of security started with the detection of abnormal events in videos captured by static surveillance cameras.
With advancements in mobile robot technologies, the interest in video monitoring via patrol or surveillance robots has increased.
Anomalies within the security domain can be about \textit{static objects}, such as suspiciously abandoned objects in public spaces, or \textit{dynamic behaviors}, such as people fighting.

\subsubsection{Static object detection}

Works that focus on the detection of static objects can model anomalies as the unusual presence of objects, like abandoned objects~\cite{filonenko2016unattended,luna2018abandoned}, illegally parked vehicles~\cite{jo2017cumulative}, road obstructions~\cite{TIP2010,Zaham}, or unexpected objects in industrial scenes~\cite{VDAO}.

Abandoned objects can be detected in urban scenarios and in industrial environments using the systems proposed in \cite{TIP2010} and \cite{VDAO}, respectively.
Both systems employ wheeled vehicles equipped with cameras.
Because abandoned object can have different shapes, colors and textures, object detection algorithms trained on a finite set of classes may fail.
Therefore, frame-based approaches are preferred. 
A reference video can be recorded when no objects are present in the scene.
Then, given a target video, anomalous objects can be detected by matching the frames of the reference video with those of the target video.
The work by Silva et al.~\cite{VDAO} introduces a dataset composed of reference and target videos of a cluttered industrial environment.
This dataset was captured by using various cameras under different lighting conditions, containing different anomalous objects, such as backpacks, umbrellas, and bottle caps (Fig.~\ref{VDAO}).
This dataset has been used as a benchmark by several reference video approaches \cite{VDAO18TCS, VDAO18MSSP, VDAO20TIP}. 
These types of approaches are typically designed for cases where the vehicles navigate along a fixed path with a highly controlled motion.
Hence, their application is limited to environments with minor structural scene variations and they may fail if new normal objects are added to the scene.
Anomalous static objects can also be detected with a mobile robot in industrial environments by building a database of normal frames along with their geographic tags, and by comparing these normal frames with frames captured during the robot navigation \cite{Zaham}.
The system includes a human-in-the-loop module, where if the anomaly detection model predicts an incoming frame as an anomaly, the operator will assess whether the alert is a genuine one. 
If the operator deems the frame as normal, this will be added to the database of normal frames, thus making the database incrementally richer over time.
However, this system is bound to the human-in-loop input, which can render the system impractical to apply to large-scale applications.
The works by Lawson et al.~\cite{patrolrobot16,patrolrobot17} focus on the detection of anomalous objects using autonomous robots within known office environments. 
The anomaly detection model is configured to learn normal conditions in an unsupervised manner.
Their proposed solution employs a Generative Adversarial Network to compare the robot's current views with the learned model of normality to detect anomalies. 
However, if objects are moved from one place to another, due to their specific affordance, this may trigger false anomaly alarms.

\begin{figure*}[t]
    \centering
    \includegraphics[width=1\textwidth]{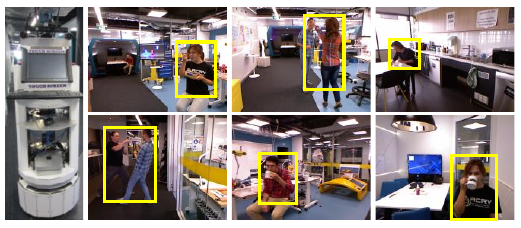}
    \vspace{-8mm}
    \caption{Guiabot robot (shown on the left) is used to generate action region proposals in images and detect abnormal behaviors in the campus environment, including eating, drinking, and doing sports. 
    Figure taken from \cite{ICRA17}.}
    \label{guaibot}
\end{figure*}

\subsubsection{Dynamic object and behavior analysis}

Dynamic abnormal behavior detection has consistently been a primary research focus in VAD \cite{UCSD,CUHK,ShanghaiTech,ubnormal}, as human is one of the main concerning contributors to anomalies.
Several related applications have gradually emerged in MC-VAD with the use of Unmanned Aerial Vehicles (UAVs) \cite{PAMI2001, MDV2018, Eyeinsky2018, MDV2020, MDV2021, srivastava2021recognizing} and wheeled robots \cite{RE-DID, ICRA17} for monitoring human behavior.
The system presented by Medioni et al.~\cite{PAMI2001} features road monitoring by analyzing video sequences captured from UAVs. 
The system detects and tracks road vehicles, and models the speed of cars as Gaussian distributions, taking into account both scene context (such as road and checkpoint areas) and task context (like avoiding checkpoints and passing checkpoints). 
An alarm is triggered if the speed of a vehicle falls outside the distribution.
This system is highly customized to a specific application scenario as it makes use of prior knowledge in terms of the properties of dynamic objects, the scene context, and the task context, making its application easily ineffective in the case of scenario changes.
Another application of MC-VAD with UAV-equipped cameras is for the detection of anomalous activities in parking lots \cite{minidrone}, such as fighting or stealing \cite{MDV2018, MDV2020, MDV2021}. 
These videos involve nearly-continuous camera motion, except when the drone is in a hovering position.
However, the development of generalized data-driven solutions is challenging because the dataset contains numerous small targets.
Moreover, videos can be noisy, and the errors they generate are likely to stem from the frame-level feature extraction approaches used for anomaly detection \cite{backgroundbias}.
The system proposed by Singh et al.~\cite{Eyeinsky2018} features real-time detection of violent individuals in public areas using drones.
This system uses a Feature Pyramid Network \cite{lin2017feature} to detect humans and to estimate their pose.
The angles between the limbs are utilized by a Support Vector Machine to classify individuals engaging in violent activities, such as punching, kicking, falling, strangling, throwing, swinging pipe, pushing, and SOS signals \cite{srivastava2021recognizing}.
Face detection and identification techniques can be further integrated into the system to identify violent individuals \cite{srivastava2022uav}. 
However, dealing with individual identities requires careful ethical considerations to be compliant with privacy-related regulations.
There also exist mobile robot systems to detect actions like eating, drinking, and sports activities \cite{ICRA17}.
The system can operate by first classifying the scene, then recognizing the action, and finally calculating the occurrence probability of certain actions in a scene (Fig.~\ref{guaibot}).
The system can then be tested on both static and moving camera datasets.
When the system has to process unseen actions or actions that exhibit a high degree of similarity to others, the accuracy of anomaly detection can significantly decrease.

\subsection{Urban transportation}

In the context of urban transportation, autonomous driving applications are most prevalent. 
These involve detecting long-tailed unexpected activities, such as animals darting across the road or confronting oncoming drivers in the wrong lane. 
It is not only crucial to detect these activities reliably, but also to do so in a timely manner.

\subsubsection{Traffic accident detection}

Training a model to anticipate accidents from dashcam videos can facilitate the generation of timely warnings, potentially a few seconds prior to the occurrence of and accident \cite{DAD}.
Dashcams are affordable cameras that can be mounted inside a vehicle and that can record street-level visual observations from the driver's point of view.
In recent years, dashcams have become nearly ubiquitous in cars, making them among the most widely deployed mobile cameras. 
A substantial number of dashcam videos, including those involving accidents, have been shared on websites such as YouTube, and several researchers have leveraged this wealth of data to build traffic accident datasets for their studies \cite{A3D, DoTA, DAD, HTA, CCD, dada2000, d2city, truckVAD}.
Traffic anomalies can be categorized as collision or loss of control \cite{DoTA}.
Each of these can be further divided into non-ego and ego-involved anomalies.
Non-ego anomalies involve situations where the ego-vehicle is just an ``observer" of the anomaly, hence capturing a collision between two vehicles at a distance.
Ego-involved anomalies are situations where the ego-vehicle is directly involved, such as in collision with other vehicles, in which the ego-vehicle is a ``participant" in the anomaly. 
This classification broadens the scope of anomaly detection tasks, and extends anomaly detection to video action recognition and online action detection.
Other anomalous traffic patterns, such as speeding motorcycles, halted vehicles, and close merges, are also of interest to be detected from dashcam videos of vehicles on highways \cite{HTA}.
Compared to the setting of traffic accident detection, the featured anomalous events in this work represent more common and general cases where drivers should pay attention to while driving.
However, this work only tackles five types of anomalies with 25 abnormal traffic videos, which is rather limited compared to the diversity of traffic anomalies that one can find in the real world.
Driver attention prediction can be integrated with the information extracted from dashcam accident videos, in order to better understand the relationship between driver attention and accident prediction \cite{dada2000}. 
The annotation of the driver attention can be added to dashcam accident videos to facilitate the training of the prediction models for driver attention. 
Such study assists in the development of more precise and effective driver attention monitoring devices, e.g.~eye-tracking technology and driver fatigue detection systems. 
However, as it is difficult to obtain real driver attention data during real accidents, the attention annotation is often collected from volunteers after the incident, which might introduce potential biases.

\subsubsection{Obstacle detection}

According to the statistics provided in \cite{national2021traffic}, road debris and obstacles are the cause of a significant number of accidents.
Obstacles can lead to crashes, hence mission failures. 
Hence, obstacle detection is critical in both autonomous driving and robotic navigation.
A stereo vision-based method can be used to detect small but important road hazards for autonomous vehicles, such as debris and lost cargos \cite{pinggera2016lost}. 
The method incorporates photometric cues, considering patches that deviate from the expected road appearance as potential hazards. 
As only photometric features are sensitive to lighting conditions, the method may fail in detecting small obstacles in badly lit areas or at night.
A multi-modal framework that combines sparse LiDAR and monocular vision can help mitigate the previously mentioned shortcoming in systems based only on photometric information \cite{singh2020lidar}.
The accurate alignment between LiDAR and camera is key for precise detection and can be achieved by using an external parameter refinement method based on the Hausdorff distance.
This allows the segmentation of obstacles that are smaller than 15 centimeters at a distance of 50 meters.
In robotic navigation, humans are often considered as dynamic obstacles. 
Hence, a stereo camera-based system for multi-person tracking in crowded spaces can be used \cite{obstacle2009}. 
The system jointly estimates camera position, stereo depth, object detection, and trajectories based only on visual information, then predicts future motion for dynamic objects, and fuses this information with a static occupancy map estimated from dense stereo. 
Such controlled navigation is particularly important for robotic platforms that assist people.
For example, the work by Wang et al.~\cite{8786197} features robotic wheelchairs and proposes a pipeline to automatically label drivable areas and road anomalies with self-supervised learning. 
Such labels are then used to train RGB-D semantic segmentation neural networks.
Unlike self-driving cars that primarily focus on pedestrians and cars on roadways \cite{kitti2012, cordts2016cityscapes}, the robotic wheelchair \cite{8786197} focuses on the detection of small obstacles, such as stones and tin cans in both indoor and outdoor scenes.

\subsection{Marine environments}

Open water covers over 70\% of the Earth, and over 80\% of world merchandise trade by volume is carried out by sea \cite{sirimanne2019review}. 
Traditional monitoring is performed by human, and thus is limited due to constraints such as monitoring range, time, and perspective.
The use of robotic platforms equipped with cameras is rapidly expanding in this field, as evidenced by recent surveys and workshops \cite{ipatch,8911242,kiefer20231st}.
Relevant applications to MC-VAD can be divided into 
\textit{obstacle detection}, which can be carried out by Unmanned Surface Vehicles (USVs), 
\textit{environment surveillance}, which can be carried out from Unmanned Aerial Vehicles (UAVs), and 
\textit{underwater exploration}, which can be carried out by Unmanned Underwater Vehicles (UUVs).

\subsubsection{Obstacle detection}

USVs are frequently employed robotic platforms designed to navigate across oceanic environments for environmental monitoring or water rescue.
USVs generally function by autonomously navigating along pre-set waypoints or pre-programmed paths, thus detecting and avoiding obstacles is key to ensure safe operations.
Recently, dynamic camera-based visual methods have been explored for obstacle detection in marine environment, including semantic segmentation \cite{MODS2022}, object detection and tracking \cite{SMD}. 
While these techniques facilitate more accurate and efficient obstacle identification, most of such works have not considered detection in nighttime scenarios, which is equally important for long-duration navigation of USVs.

\subsubsection{Environment surveillance}

Maritime surveillance is key for both law enforcement and environmental protection. 
The implementation of an intelligent maritime surveillance system using UAVs equipped with cameras, can enhance the awareness of maritime incidents, effectively replacing the human effort required for extensive and long-term ocean monitoring.
UAVs can aid in detecting oil spills and hazardous substances, tracking marine objects like vessels and debris, and recognizing behavioral patterns, such as vessel rendezvous and high-speed vessels \cite{seagull}. 
The versatility and swift operational capability of UAVs can also be leveraged for search and rescue missions \cite{seadronesee}. 
The primary targets for detection in these scenarios are swimmers, floating objects, and vessels in open waters. 
Currently, most of these works only focus on basic detection, tracking, and segmentation. 
However, we believe the utilization of contextual information, e.g.~scene context or temporal context, can also be important for future research in maritime surveillance.

\begin{figure*}[t]
\centering
\includegraphics[width=.9\textwidth]{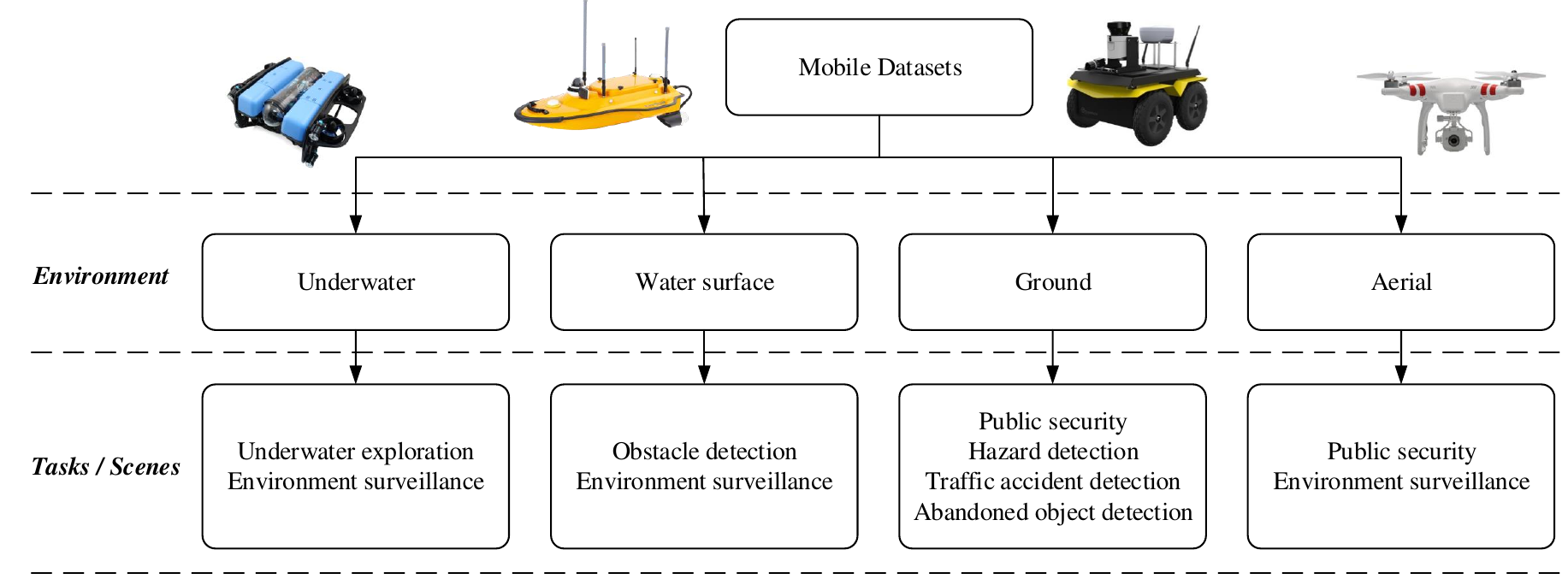}
\vspace{-3mm}
\caption{
MC-VAD dataset taxonomy.
We categorize datasets into four types of environments: underwater, water surface, ground, and aerial. 
(top) Typical mobile camera platforms that are used for capturing these datasets, i.e.~unmanned underwater vehicle, unmanned surface vehicle, unmanned ground vehicle, and unmanned aerial vehicle, respectively. 
(bottom) Different environments correspond to different tasks or detection scenes.}
\label{taxonomy}
\end{figure*}

\subsubsection{Underwater exploration}

The application of MC-VAD for underwater exploration primarily focuses on the detection of novel objects. 
This exploration task is challenging due to the low visibility and unpredictable nature of underwater environments. 
We observe that research in this area is currently limited, with few dedicated datasets available for the development and evaluation.
The work by Zhou et al.~\cite{zhou2022discovering} represents a promising attempt in this task. 
It aims to visually locate unknown objects, such as underwater sculptures, to facilitate autonomous exploration in underwater environments. 
The study utilizes both simulated and real video sequences where the sandy seafloor, rocks, and oscillating seagrasses are treated as normal data. 
In contrast, anomalies or targets are represented by objects such as shipwrecks, statues, and ship anchors.
Moreover, there are many more specific applications in underwater exploration, such as underwater garbage cleaning, observation of marine flora and fauna \cite{hong2020trashcan}, and underwater robot harvesting in marine farms \cite{UUD}.
However, the significant variations in water quality and visibility in different marine regions and depths pose challenges to the utilization and development of relevant applications.

\section{Datasets}\label{sec:Datasets}

We review publicly available RGB and RGB-D datasets that can serve for the development and evaluation of MC-VAD methods. 
Specifically, we review 25 datasets captured by moving cameras, offering a rich variety of ego-motion and complex background changes for anomaly detection in different scenarios.
As shown in Fig.~\ref{taxonomy}, we categorize these datasets based on the environment in which the mobile platform is situated, and the application scenario or tasks they feature. 
In addition, we compile various attributes of these datasets in Tab.~\ref{tab2}, such as background, number of videos, number of anomalies, types of sensors, resolution, types of anomalies or participants, occlusion, and whether the anomalies are acted. 
Tab.~\ref{tab1} presents an overview of the advantages and drawbacks of these datasets.

\begin{table*}
\caption{Summary of MC-VAD datasets and their properties. Key. *: dataset contains images instead of videos, I: Images, T: Type, U: Unspecified, Occ: Occlusion, Act: Acted, EO: Electro-Optics sensor, IR: Infrared Radiation, NV: Night-vision, NI: Near Infrared.
The dataset name is clickable and links to the dataset webpage.}
\vspace{-3mm}
\setlength{\tabcolsep}{4pt}
\label{tab2}
\begin{tabular}{lllllllll}

\toprule

    \multicolumn{1}{l}{Dataset} &Background & Number & Number of &Modality & Resolution & Typical of anomalies  & Occ & Act  \\ 
    &&of videos &anomalies &&&or participants \\
    &&/images \\
\midrule
    \href{https://irvlab.cs.umn.edu/resources/suim-dataset}{SUIM*} \cite{SUIM2020} & Dynamic  &- &-   & RGB  & Multi  & underwater obstacle   & $\checkmark$ & \\ 

    \href{https://conservancy.umn.edu/handle/11299/214865}{TrashCan*} \cite{hong2020trashcan}   &Static  & 7212 I  & 7212 I &RGB & Multi  &underwater trash and fish  &  & \\
    
    \href{https://github.com/chongweiliu/UDD\_Official}{UUD*} \cite{UUD}  &Static  &-  &-   &RGB   &Multi &seacucumber, seaurchin  &$\checkmark$  &\\

    \href{mailto:j.l.patinovilchis@reading.ac.uk}{IPATCH} \cite{ipatch} &Dynamic  &24   &17   &RGB, thermal   &Multi  &pirate surveillance  &   &$\checkmark$ \\
    
    \href{https://sites.google.com/site/dilipprasad/home/singapore-maritime-dataset}{SMD}\cite{SMD} &Dynamic  &11  &11   &RGB, EO   &1920$\times$1080 & vessels  &$\checkmark$  &\\
    
    \href{https://www.vicos.si/resources/}{MODS} \cite{MODS2022} &Dynamic   & 94  &94    &RGB (stereo)    & 1278$\times$958  & marine obstacle    &$\checkmark$  &    \\

    \href{https://cvgl.stanford.edu/projects/uav_data/}{DAD} \cite{DAD}   & Dynamic  & 1730    & 620    &RGB  & 1280$\times$720   & traffic accident     &$\checkmark$  &\\
    
    \href{https://github.com/JWFangit/LOTVS-DADA}{DADA-2000} \cite{dada2000}  &Dynamic  & 2000   & 2000   &RGB  & 1456$\times$660   & traffic accident    &$\checkmark$  &\\
    
    \href{https://github.com/harpreets652/highway-traffic-anomaly}{HTA} \cite{HTA}  & Dynamic & 389   & 25    &RGB  & 1280$\times$720   & traffic anomalies    &$\checkmark$  &\\
    
    \href{https://github.com/Cogito2012/CarCrashDataset}{Car Crash Dataset} \cite{CCD} & Dynamic & 4500   & 1500   &RGB  & U   & traffic accident   &$\checkmark$  &\\
    
    \href{https://github.com/MoonBlvd/Detection-of-Traffic-Anomaly}{DoTA} \cite{DoTA}  & Dynamic & 4677   & 4677  &RGB  & 1280$\times$720  & traffic accident    &$\checkmark$  &\\
    
    \href{http://www.lehre.dhbw-stuttgart.de/~sgehrig/lostAndFoundDataset/index.html}{Lost\&Found} \cite{pinggera2016lost} &Hybrid &112   &112  &RGB (stereo)  &2048$\times$1024   & small road hazards  &  &$\checkmark$\\
    
    \href{https://vision.eng.au.dk/fieldsafe/}{FieldSAFE} \cite{kragh2017fieldsafe}  &Hybrid  &5  &5  &RGB (stereo), LiDAR &1920$\times$1080     &obstacle in agriculture    &  &$\checkmark$\\

    &  &  &  &thermal &     &    &  &\\
    
    \href{https://small-obstacle-dataset.github.io/}{Small Obstacle} \cite{singh2020lidar} &Static &15 &15 & RGB (stereo), LiDAR  &1920$\times$1080   &off-road and small-obstacle    &  &$\checkmark$\\

    \href{https://github.com/hlwang1124/GMRPD}{GMRP*} \cite{GMRPD}  &Static &3896 I &3896 I &RGB-D &1280 x 720 &road anomalies  & &\\
    
    \href{https://github.com/idsia-robotics/hazard-detection}{Hazards\&Robots} \cite{mantegazza2022outlier} &Static  &U   & 16 T  &RGB  & 512$\times$512  & hazard environment  &  &$\checkmark$ \\ 
    
    \href{http://cvlab.epfl.ch/data/road-anomaly/}{RoadAnomaly*} \cite{roadanomal} &Static &60 I &60 I &RGB  &1280 $\times$ 720  &road obstacles  & &$\checkmark$\\
    
    \href{http://www.smt.ufrj.br/~tvdigital/database/objects}{VDAO}  \cite{VDAO} &Static   & 66   & 62  &RGB   & 1280$\times$720   & industrial abandoned-object     &$\checkmark$  &$\checkmark$\\
    
    \href{http://loki.disi.unitn.it/~ReDID/}{RE-DID} \cite{RE-DID} &Dynamic  & 30   & 30   &RGB    & 1280$\times$720   & fighting     &$\checkmark$  &\\
    
    \href{https://webpages.charlotte.edu/cchen62/dataset.html}{UCF-Crime} \cite{UCF} &Dynamic & 1900   & 1900  &RGB & Multi   & abuse, arrest,accident,etc    &  &\\

    \href{uty@kiro.re.kr}{X-MAS} \cite{XMAS} &Dynamic & 2624   & 1220  &RGB-D, IR, LiDAR & U   & fallen, convulsions, etc    &$\checkmark$  &$\checkmark$\\

    &  &  &  & thermal&     &    &  &\\
    
    \href{https://www.epfl.ch/labs/mmspg/downloads/mini-drone/}{mini-drone video} \cite{minidrone} &Dynamic & 38   & 38  &RGB    & 960$\times$540   & attacking,stealing,etc    &  &$\checkmark$\\
    
    \href{https://github.com/sutdcv/UAV-Human}{UAV-Human} \cite{li2021uav}  &Dynamic  & 22,476  & U  &RGB-D, IR, NV  & 1920$\times$1080  & human behaviour    &  &$\checkmark$\\
    
    &  &  &  & skeleton&     &    &  &\\
    
    \href{https://vislab.isr.tecnico.ulisboa.pt/seagull-dataset/}{Seagull} \cite{seagull} &Dynamic &19   &19  &RGB-D, IR, NV   & Multi  & vessels, hydrocarbon slick     & & \\
    
    \href{https://seadronessee.cs.uni-tuebingen.de/home}{SeaDronesSee} \cite{seadronesee}  &Dynamic  &U  &U &RGB, RedEdge, NI &Multi &swimmer, floater  & &$\checkmark$\\
\bottomrule
\end{tabular}
\end{table*}

\subsection{By platforms under water}

While existing datasets captured by underwater mobile platforms are primarily designed for underwater image enhancement~\cite{8917818,berman2020underwater}, there are datasets introduced for object detection or semantic segmentation, that contain valuable information about the underwater environment and the types of objects one might encounter (see Fig.~\ref{uuvusv}a-c). 
These can be useful for developing and evaluating MC-VAD methods in the context of underwater exploration~\cite{zhou2022discovering}.

The SUIM dataset~\cite{SUIM2020} is a large-scale dataset that contains over 1,500 images with pixel-level annotations for eight object categories: fish, reefs, aquatic plants, wrecks/ruins, human divers, robots, and seafloor. 
These images are captured with high visibility, allowing for the clear observation of a variety of underwater objects. 
Although the primary use of the SUIM dataset is for underwater image semantic segmentation, it can also be repurposed for MC-VAD.
For example, the SUIM dataset is used in \cite{zhou2022discovering} to validate detection of unknown objects with dynamic backgrounds as they were anomalies in underwater scenarios. 
The algorithm employs images without sea fauna as the training set (normal) and images with marine animals (e.g.~fish, turtle) as the test set (anomalies).

The UUD dataset~\cite{UUD} is an underwater (open-sea farm) object detection dataset comprising three categories of objects, i.e.~sea cucumber, sea urchin, and scallop, which can be used as anomaly samples.
UUD contains 2,227 annotated images, which are captured by high-definition cameras carried by divers and robots. 
The image resolution varies and it is 720p, 1080p, and 4K.
UUD presents typical dynamic backgrounds and motion blur, but its utility for anomaly detection is somewhat limited by the small number of object categories.

The TrashCan dataset~\cite{hong2020trashcan} is an instance-segmentation dataset of underwater trash. 
It consists of 7,212 annotated images and includes observations of trash, Remotely Operated underwater Vehicles (ROVs), and a diversity of underwater flora and fauna scenarios.
Images are extracted from nearly 1K real ROV-captured underwater videos, with the captured trash items varying significantly in shape.
However, the complex structure of the seabed and low visibility present challenges for detection.

\subsection{By platforms on the water surface}

\begin{table*}
\caption{Advantages and disadvantages of MC-VAD datasets organized based on their application domains, anomaly categories and data collection methods. Paper citations corresponding to the dataset abbreviations are detailed in Tab.~\ref{tab2}.}
\vspace{-3mm}
\label{tab1}
\tabcolsep 3pt
\resizebox{\textwidth}{!}{%
\begin{tabular}{p{3.5cm}p{6cm}p{6cm}p{4cm}}
\toprule
Datasets & Advantage & Disadvantage &Application domain\\
\midrule
SUIM, TrashCan, UUD & Natural representation of underwater scenes & Image dataset without temporal information & Underwater exploration \\

IPATCH & Five complete maritime anomaly scenarios & Limited number of events & Abnormal behavior in marine \\

SMD, MODS  & Large amount of marine scenes and obstacles & Few types of obstacles & Obstacle detection in marine \\

DAD, DADA-2000, CCD, DoTA & Great amount of diversity in terms of real motion anomalies, a large variation in camera ego-motion, and rich annotations &  Low-resolution and noisy videos & Traffic accident detection \\

HTA & High-definition and smooth ego-motion videos & Few types of anomalies & Traffic anomalous behavior \\  

Lost\&Found, Small Obstacle, FieldSAFE, GMRP, RoadAnomaly & Multi-modal data and several small obstacles of different types & Lack of samples of diverse moving anomalous objects & Obstacle detection on road \\

Hazard\&Robots & Various types of hazard events & Few numbers of scenes & Hazard environment detection \\

VDAO & Target and reference videos with and without anomalies in the same scene & Camera motion is simple and slow & Abandoned object detection \\

RE-DID & Real-world fights captured from moving cameras & Small number of samples  & Human abnormal behavior \\

UCF-Crime & Great number of diverse types of real-world anomalous events & Few videos containing PTZ camera motion & Abnormal behavior \\

minidrone & Abnormal events captured from aerial moving cameras & Few abnormal event categories and samples & Human behavior surveillance \\

UAV-Human & Large number of anomalous actions captured from aerial moving cameras & Short and indistinct duration of actions & Human behavior surveillance \\

Seagull, SeaDroneSee & Natural marine scenes captured from aerial moving cameras & Few object categories and scenes & Marine surveillance \\

\bottomrule
\end{tabular}
}
\end{table*}


Datasets captured by USVs often encompass data of multiple modalities acquired from cameras, radars, and Electro-Optical (EO) sensors. 
Several such datasets have been proposed to test and validate algorithms designed for maritime surveillance and robotic navigation~\cite{7301727, 9381638, Cheng_2021_ICCV, liu2021efficient}. 
However, these datasets do not directly cater to the needs of MC-VAD due to factors such as small camera motion, limited types of detection targets, or simplistic scenarios.

\begin{figure*}[t]
\centering
\includegraphics[width=.9\textwidth]{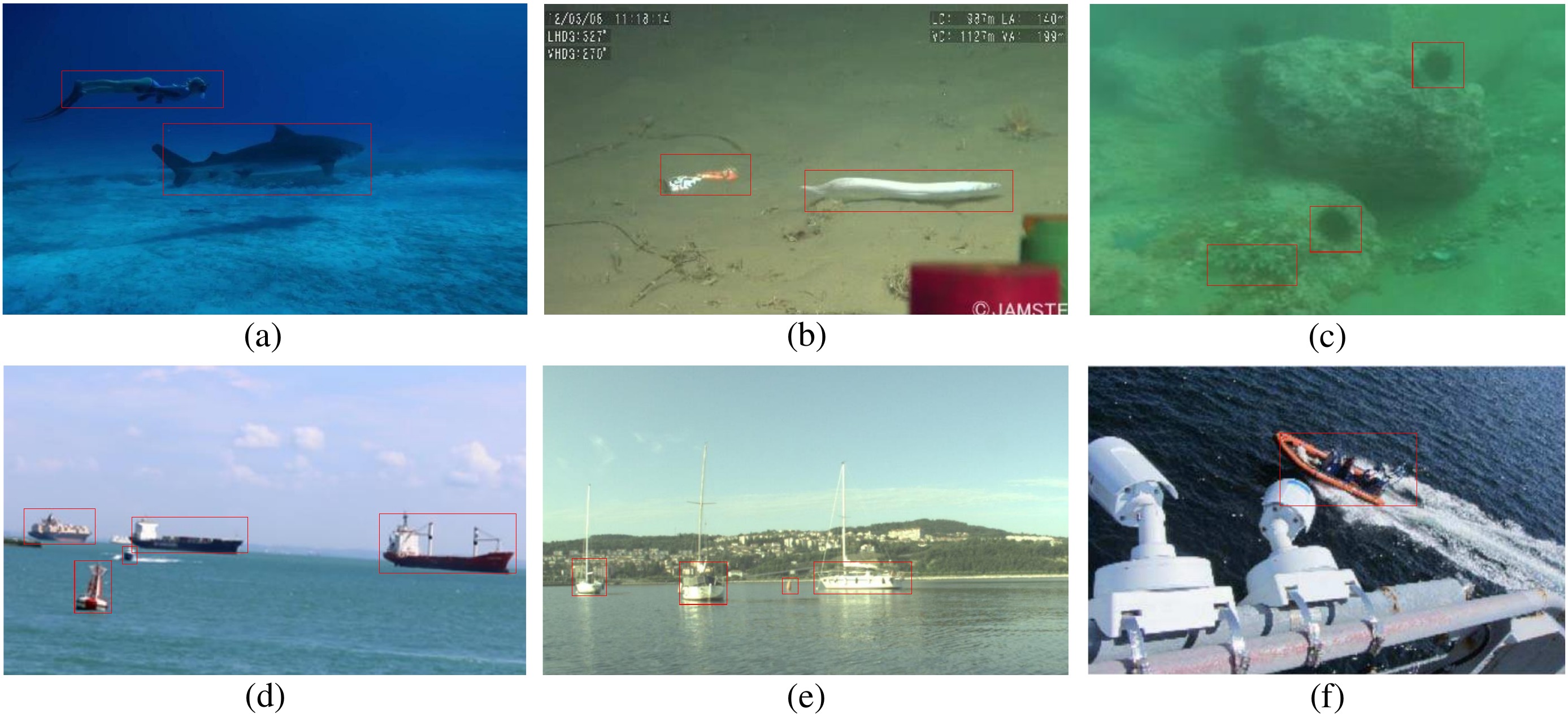}
\vspace{-3mm}
\caption{Examples of anomaly from videos captured (a-c) under water and (d-f) on the water surface.
(a) Divers, fish, and shipwrecks from the SUIM dataset~\cite{SUIM2020}. 
(b) Trash can and fish from the TrashCan dataset~\cite{hong2020trashcan}.
(c) Sea cucumber and sea urchin from the UUD dataset~\cite{UUD}.
(d) Large vessels, small boats, and Ocean Buoy from the Singapore Marine Detection dataset~\cite{SMD}.
(e) Yacht and marine debris from the Marine Obstacle Detection dataset~\cite{MODS2022}.
(f) A boat that suddenly accelerates across a ship from the IPATCH dataset~\cite{ipatch}.
}
\label{uuvusv}
\end{figure*}

We identify a list of datasets that can serve for MC-VAD, especially in the context of automation of maritime monitoring and navigation security, for tasks such as detecting anomalous vessel behaviors~\cite{ipatch} and surface obstacles~\cite{kristan2015fast, SMD, modd2, MODS2022} (Fig.~\ref{uuvusv}d-f).
These datasets feature complex scenarios and contain a wide range of targets, making them suitable for training and testing MC-VAD models.

The Marine Obstacle Detection Dataset (MODD1)~\cite{kristan2015fast} is a pioneering small-scale dataset that consists of 12 video sequences, with 4,454 fully annotated frames of image resolution 640$\times$480 pixels. 
The video sequences are recorded from multiple unmanned platforms (mostly from small USVs) over several months, under different weather conditions, and at different times of the day. 
The dataset contains situations that pose a danger to USV, such as approaching obstacles and collision threats. 
Subsequently, \cite{modd2} upgraded MODD1 to MODD2 that contains 28 video sequences with 11,675 frames at a resolution of 1278$\times$958 pixels. 
These datasets include various challenging conditions, such as abrupt motion change of USVs, environmental reflections in water, and sun glitter.

The IPATCH dataset~\cite{ipatch} is designed for maritime anomaly detection, with a specific focus on preventing pirate activities.
IPATCH consists of 24 videos, captured by multiple cameras stationed on a large tugboat.
The dataset can support three stages of video analysis: low-level analysis for object detection and tracking, mid-level analysis for simple abnormal event detection, and high-level analysis for complex threat event detection.
The types of events include simple ones such as boats speeding up, loitering, or moving around a vessel, as well as more complex events that involve real threats, such as abnormal approaching to the vessel of a boat that concludes with the skiff staying at the vessel starboard.

The SMD~\cite{SMD} dataset is collected using Canon 70D cameras and Electro-Optical sensors for marine object detection.
The dataset consists of 70 on-shore videos and 11 on-board videos, captured by a stationary camera on-shore and a mobile camera on a vessel, respectively. 
All the videos in the dataset are of high-definition (1080$\times$1920 pixels). 
Moreover, the dataset features various environmental conditions, such as before sunrise, sunrise, noon, evening, haze, and rain.

Finally, the MODS dataset~\cite{MODS2022} serves for marine object detection in order to avoid obstacles during the autonomous operation of small-sized USVs. 
It contains 81K stereo images synchronized with an on-board IMU, with over 60K objects annotated. 
MODS covers a broad range of obstacle appearances and types, including various dynamic obstacles such as boats, buoys, swimmers, and swans, as well as static obstacles such as shores and wharves. 
MODS training set is a combination of the SMD~\cite{SMD}, MODD1~\cite{kristan2015fast}, and MODD2~\cite{modd2} datasets.
The dataset presents real-world challenges such as different weather conditions, water surface ripples, and sunlight reflections on the water.

\begin{figure*}[t]
\centering
\includegraphics[width=1\textwidth]{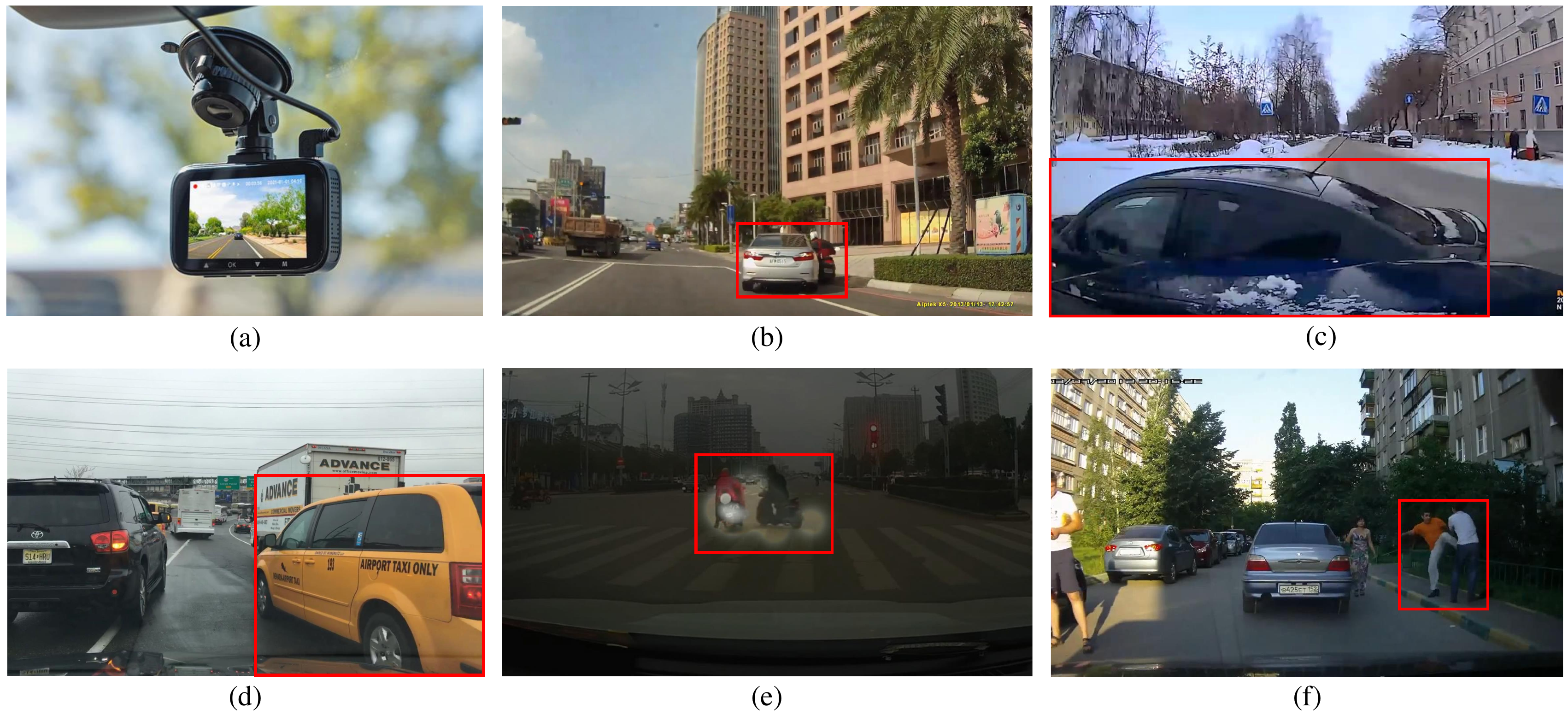}
\vspace{-7mm}
\caption{Examples of anomalous events captured by dashcams mounted on cars.
(a) Image of a dashcam mounted on a car's windshield.
(b) Non-ego traffic accident (collision) from the DAD dataset \cite{DAD}.
(c) Ego-involved traffic accident (collision) from the DoTA dataset \cite{DoTA}.
(d) Traffic anomalous behavior (close merge) from the HTA dataset \cite{HTA}.
(e) SMI-RED250 infrared eye-tracker provides extra driver attention information (white dot) for crowd-sourced traffic accident videos in the DADA-2000 dataset \cite{dada2000}.
(f) Real-life events-dyadic interactions (fighting) from the RE-DID dataset \cite{RE-DID}.
}
\label{dashcam}
\end{figure*}

\subsection{By platforms on the ground}

In this section, we present datasets that are collected by mobile platforms that operate on the ground, such as wheeled robots and autonomous cars.
Among all MC-VAD datasets, those with moving cameras on the ground constitute the largest portion of datasets overall, including indoor scenes, urban streets, and farmland scenarios.
Most of the ground datasets are designed for safe autonomous navigation where anomalies are typically potential hazards to be avoided. 
For example, some of these datasets are designed for detecting traffic accidents~\cite{DAD, HEVI, A3D, dada2000, HTA, CCD, DoTA}, obstacles on urban roads~\cite{pinggera2016lost, singh2020lidar, roadanomal}, rail transit irregularities~\cite{7533104, railanomaly,he2021obstacle}, hazards in agricultural settings~\cite{kragh2017fieldsafe}, and detection in hazardous environments~\cite{mantegazza2022outlier}.
Some of them are utilized for patrolling or surveillance where anomalies are events of interest that need to be reported, such as abandoned object detection~\cite{VDAO} and violent behavior detection~\cite{RE-DID}.
Because RGB cameras are nowadays inexpensive, they are often used to capture data, in particular dashcams are rather popular and commonly installed in vehicles.
Dashcams serve as the main video contributors for road monitoring and anomalies.
The primary challenges in dealing with dashcam datasets involve unpredictable high-speed camera motion, low resolution, reflections on car windows, occlusions, and changing weather and traffic conditions.

The Re-DID dataset~\cite{RE-DID} is a collection of 30 YouTube videos for the study of detecting fighting events by identifying the area associated with interpersonal space (Fig.~\ref{dashcam}f). Among the videos, 25 of them are recorded using dashcams, while the remaining ones are taken by mobile phones. 
Re-DID includes 73 different fight instances under different lighting (day, night), weather conditions (sunny, rainy), camera views (wide angle, fish-eye, zoomed view), and moving and static scenes. 
Re-DID also contains annotations of subjects' bounding boxes, temporal windows where the interaction occurs, and interpersonal space regions.

The DAD dataset~\cite{DAD} is crowd sourced from dashcam footage (Fig.~\ref{dashcam}a,b).
It comprises 678 accident videos from six major cities in Taiwan, featuring a variety of collisions involving motorbikes and cars as primary participants. 
Each positive clip in the dataset contains the moment of the accident within the final ten frames, while a negative clip includes no accident. 
Compared to datasets such as KITTI~\cite{kitti2012} that is captured by high-quality cameras in normal driving scenarios, the DAD dataset presents more complex road scenes and diverse accident scenarios.

The A3D dataset~\cite{A3D} is another dashcam dataset containing 1.5K road anomaly event videos captured from different vehicles in East Asia.
These videos are categorized into 18 types of traffic accidents, and feature various weather conditions (e.g.~sunny, rainy, snowy), driving scenarios (e.g.~urban or rural), and participants (e.g.~cars, motorcycles, pedestrians and animals). 
The start and end times of each anomaly are annotated: the start time is defined as the point the accident becomes unavoidable, and the end time is defined as the point where the participants resume their normal activity.  
However, the video quality is generally low, making anomaly detection, object detection, and ego-motion estimation challenging.

The DADA-2000 dataset~\cite{dada2000} mainly serves for the study of predicting driver attention in accident scenarios. 
It contains 2K dashcam video clips on 54 kinds of accidents along with the maps of fixations, saccade scan path, and focusing time to represent driver attention (Fig.~\ref{dashcam}e).

The HTA dataset~\cite{HTA} targets the challenge of identifying abnormal traffic patterns on highways using dashcam footage. 
HTA is derived from the Berkeley DeepDrive dataset~\cite{BDD100K}, which includes 100K high-resolution (1280$\times$720 pixels, 30 FPS) dashcam videos shot by vehicles in New York and the Bay Area.
HTA features 389 curated videos, where normality is characterized by the motion of vehicles that do not disrupt the motion of the vehicle with the dashcam, or where the movement of other vehicles remains relatively self-similar (Fig.~\ref{dashcam}d).
Anomalies are categorized as speeding vehicles, close merging events, halted vehicles, vehicle accidents, and speeding motorcycles. 
Each of these instances depicts a scenario where a human driver would typically react with caution.

\begin{figure*}[t]
\centering
\includegraphics[width=1\textwidth]{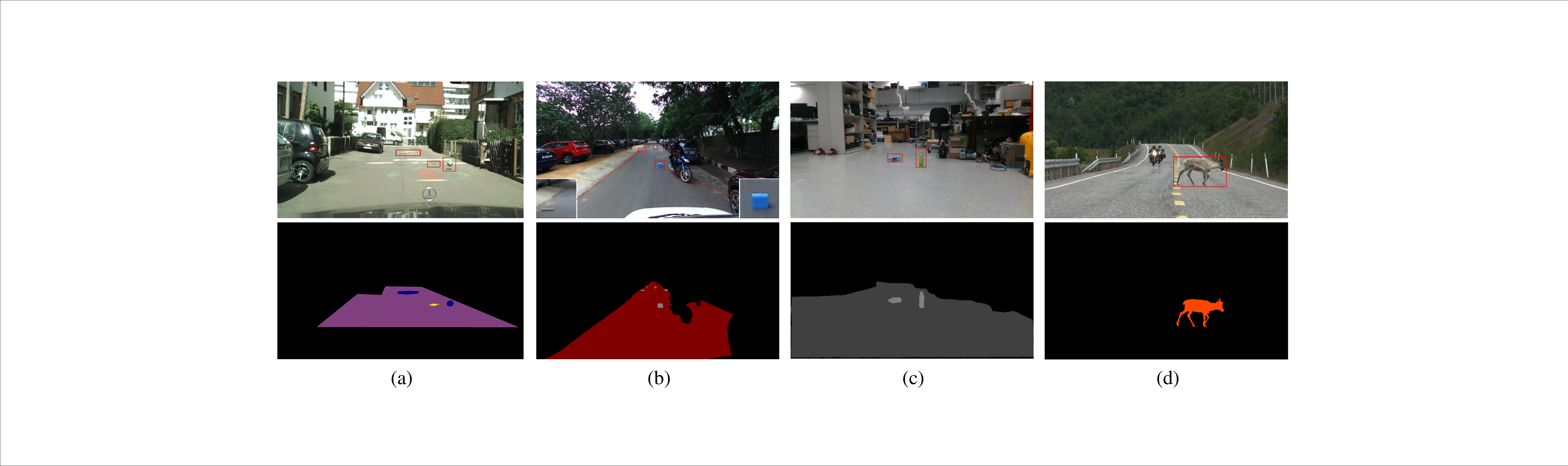}
\vspace{-7mm}
\caption{Examples of anomaly (obstacle) detection performed on videos captured from unmanned ground vehicles.
(first row) RGB image with overlaid bounding boxes indicating the location of the anomaly.
(second row) Ground-truth segmentation masks containing anomalous objects.
(a) Static and dynamic obstacles from the Lost\&Found dataset \cite{pinggera2016lost}.
The ground-truth mask shows the free space in purple, a pallet and a ball in blue, and a piece of wood in yellow.
(b) Small hazards (boxes and stones) from the Small Obstacle dataset \cite{singh2020lidar}.
The ground-truth mask shows the road in red, off-road in black, and obstacles in other colors.
(c) Static obstacles (bottles) from the Ground Mobile Robot Perception dataset \cite{8786197}.
The ground-truth mask shows the road in dark gray, the rest of the environment in black, and obstacles in light gray.
(d) Dynamic obstacle (animal) from the RoadAnomaly dataset \cite{roadanomal}. 
The ground-truth mask shows the obstacle in red.}
\label{obstacles}
\end{figure*}

The CCD dataset~\cite{CCD} contains both abnormal and normal dashcam videos. 
It includes 1.5K traffic accident video clips and 3K normal dashcam videos from BDD100K~\cite{BDD100K}. 
The CCD dataset is larger than the DAD and A3D datasets, and contains richer annotations including environmental attributes, details about whether ego-vehicles are involved, and descriptions of the reasons for each accident.

The DoTA dataset~\cite{DoTA}, an enhancement of the A3D dataset, comprises 4,677 anomaly videos that are annotated for temporal, spatial, and categorical attributes (Fig.~\ref{dashcam}c). 
The anomalies in the dataset are categorized into 18 types, including different kinds of collisions, instances of vehicles going out of control, turning or crossing situations, interactions between vehicles and pedestrians, vehicles and objects, and so on. 
Each category is further divided into ego anomalies and non-ego anomalies.
The richness of the categorical annotations in DoTA has been leveraged to evaluate video action detection and online action detection methodologies.
Other dashcam video datasets similar to DoTA and designed for MC-VAD include the datasets introduced by \cite{GCNVAD, HEVI,d2city, truckVAD}.

There are also datasets available for detecting various types of obstacles and unexpected objects, both in indoor environments with patrolling robots or in outdoor environments, e.g. roads or fields, with autonomous vehicles. 

The VDAO dataset~\cite{VDAO} is collected with a camera equipped on a iRobot Roomba for automatic detection of abandoned objects in cluttered industrial environments (Fig.~\ref{VDAO}). The dataset comprises 6 multi-object, 54 single-object, and 4 no-object (reference) videos, collected from two different cameras. 
Various abandoned objects, such as different boxes, bags, and coats, are considered as anomalies. 
Factors like cluttered environments, object occlusions, reflections, and scene shadows in the videos may pose challenges to effective detection.

The Hazards \& Robots dataset~\cite{mantegazza2022outlier} is another indoor video dataset, captured from the front-facing RGB camera equipped on a wheeled robot, Robomaster S1. 
The robot patrols in indoor environments (e.g.~offices and university buildings) with natural and artificial illumination. 
The dataset is composed of 132,838 RGB frames with a resolution of 512$\times$512 pixels and includes 16 categories of anomalies. 
These range from subtle objects like screws on the floor to more noticeable ones like passing humans, as well as hazardous events (e.g.~water puddles to avoid) and harmless objects (e.g.~hanging pieces of cellophane). 
This dataset provides hazardous situations that are common in reality and difficult to discern.

Obstacle detection on the road ahead is an essential component for both fully autonomous vehicles and advanced driver assistance systems.
As illustrated in Fig.~\ref{obstacles}, obstacles that may appear on the road can be of various types and sizes. 
It is also challenging to detect small objects in complex and cluttered outdoor scenes due to distant objects, varying road surface appearances, and illumination changes. 
In addition to monocular RGB cameras, some of these datasets are collected using stereo cameras~\cite{pinggera2016lost}, radar~\cite{singh2020lidar}, or multi-sensor setups \cite{kragh2017fieldsafe}.

\begin{figure*}[t]
\centering
\includegraphics[width=1\textwidth]{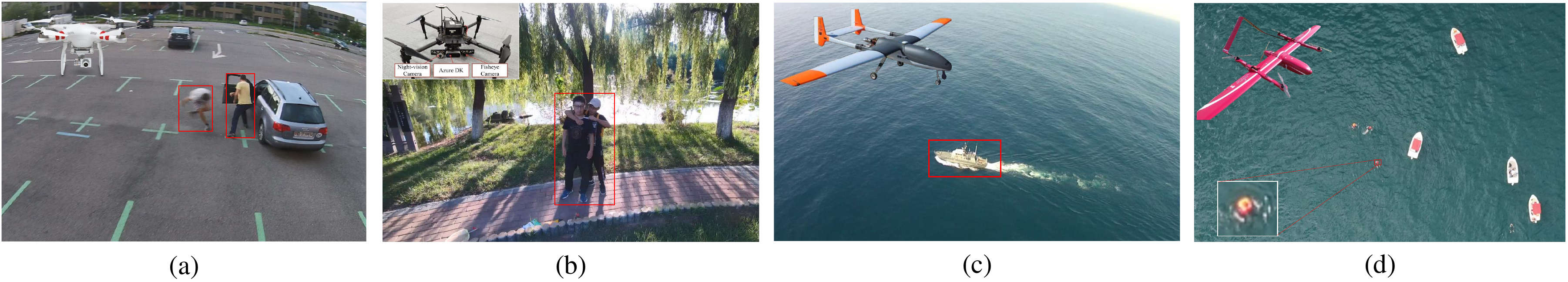}
\vspace{-7mm}
\caption{Examples of anomalies on videos captured from aerial platforms.
(a) A drone over the parking lot caught a man in yellow robbing a car from the minidrone dataset \cite{minidrone}. 
(b) A drone with multi-modal camera records a person in the act of hijacking from the UAV-human dataset \cite{li2021uav}. 
(c) A fast-moving boat on the sea from the Seagull dataset \cite{seagull}. 
(d) People in life jackets floating on the water from the SeadroneSee dataset \cite{seadronesee}.}
\label{uav}
\end{figure*}

The Lost \& Found dataset~\cite{pinggera2016lost} is collected to facilitate the study of detecting small obstacles or debris on the road for autonomous driving.
The dataset is comprised of 112 stereo video sequences, recorded from 13 different challenging street scenarios, featuring 37 different obstacle types. 
The selected scenarios represent particular challenges including irregular road profiles, far object distances, different road surface appearance
and illumination changes. 
The dataset provides broad annotations of free-space areas and detailed annotations of obstacles on the road. 
The anomalies include various small hazards and lost cargo such as pallets, boxes, and different roadblocks (Fig.~\ref{obstacles}a). 

The Road Anomaly dataset~\cite{roadanomal} is designed for detecting anomalous objects such as animals, rocks, lost tires, trash cans, and construction equipment, which are located on or near the road (Fig.~\ref{obstacles}d). 
The dataset contains 60 images, each with a resolution of 1280$\times$720 pixels, and includes per-pixel annotations of these unexpected objects. 
However, annotations are provided only for objects belonging to the abnormal class in the images, without labeling anomalies of other categories present in the scene.

RGB-D cameras can measure both the photometric properties of objects and the physical distance to their surfaces, therefore they can enhance a robot's ability to detect anomalous objects on the road, such as bumps and irregularities.
The GMRP dataset~\cite{GMRPD} comprises 3,896 pairs of RGB-D images captured using an Intel RealSense D415 camera, specifically designed for road anomaly detection in scenarios pertaining to autonomous navigation of general ground robots, such as robotic wheelchairs. 
Different from most obstacle detection datasets for autonomous cars, this dataset focuses on a broader spectrum of ground robot scenarios, encompassing 18 road anomalies across 30 different contexts, including indoor and outdoor environments like corridors and sidewalks. 
Anomalies are defined as areas rising more than five centimeters above the surface of the drivable area, including objects like traffic cones, plastic bottles, and stones (Fig.~\ref{obstacles}c).

Because LiDAR has a longer detection range than RGB-D cameras, it can identify obstacles at greater distances, and its detection capabilities are more robust to light and color variations.
The Small Obstacle dataset, introduced by Singh et al.~\cite{singh2020lidar}, includes 2,927 monocular RGB images synchronized with sparse LiDAR information from a Velodyne Puck 16 and aligned with odometry data.
The dataset annotations are divided into three categories, i.e.~road, off-road, and small obstacles. 
The external parameters for LiDAR-Camera calibration are also included.

The FieldSAFE dataset~\cite{kragh2017fieldsafe} is a multi-modal dataset that is designed for obstacle detection in agricultural scenarios.
The dataset is obtained using multiple sensors mounted on a tractor during grass-mowing activities.
The sensor pack includes a stereo camera, a thermal camera, a web camera, a 360$^\circ$ camera, a LiDAR, and a radar.
The dataset includes both static and moving obstacles such as humans, mannequin dolls, rocks, barrels, buildings, and vehicles. 
All obstacles are accompanied by ground-truth labels and geographic coordinates. 
As the dataset is collected using the Robot Operating System (ROS), hence its release is in the ROS bag format.

Lastly, the large-scale, first-person, multimodal X-MAS dataset~\cite{XMAS} consists of more than 500K image pairs, collected from a UGV navigating in outdoor environments. 
This dataset incorporates infrared, RGB, thermal, depth, and LiDAR sensors.
X-MAS covers a diverse array of challenging environmental conditions, such as varying weather patterns, including daytime, nighttime, rainy, and foggy conditions. 
The dataset is dedicated to human behavior monitoring tasks, with a particular focus on anomalous scenarios, such as falls, convulsions, and drunkenness. 
The dataset provides extensive annotations for tasks like object detection, tracking, action classification, and labeling of normal and abnormal scenes.

\subsection{By platforms in the sky}

In this section, we cover datasets that are obtained from cameras equipped on UAVs, such as rotorcraft~\cite{minidrone,li2021uav,Eyeinsky2018} or fixed-wing~\cite{seagull,seadronesee}, that feature various scenes such as cities, farmlands, and marine areas. 
Similar to surveillance videos, this type of datasets typically offers a third-person perspective from the top view, thus with a broader field of view and fewer occlusions.
However, as the distance between the camera and the monitoring areas is often very large, these datasets typically present challenges in detecting small objects as shown in Fig.~\ref{uav}.
Many of the datasets captured from flying UAVs are designed for anomalous human behavior analysis.
 
The MDV dataset~\cite{minidrone} comprises 38 videos recorded by a drone flying at low altitude in a parking lot. 
The content of this dataset falls into three categories: normal, suspicious, and illicit behaviors. 
Normal content captures people walking, entering cars, and parking vehicles. 
Whereas suspicious content portrays individuals acting in questionable ways. Illicit behaviors primarily include improper parking, theft, and physical altercations.

The AVI dataset~\cite{Eyeinsky2018} includes 2K images captured from a drone at heights varying from 2 to 8 meters. This dataset is mainly used for human violent action recognition and pose estimation, and it features five violent action classes, such as punching and kicking, carried out by 25 individuals aged between 18 and 25 years. 
The images are annotated with 14 key points for each human to aid pose estimation. 
In general, the task of detecting individual action anomalies in these aerial images presents significant challenges due to factors such as lighting variations, shadows, low resolution, and blur effects.

The UAV-Human dataset~\cite{li2021uav} is a large-scale multi-modal dataset that is designed for human behavior analysis. 
It comprises 22,476 UAV videos, each with three types of sensors: Azure DK, fisheye camera, and night-vision camera. Thus, this multi-modal dataset includes RGB, depth, IR, fisheye, night-vision, and skeleton sequences. 
The dataset features 119 distinct subjects and 155 different activity categories, encompassing over 26 anomalous types, such as vomiting, kicking, punching, stabbing someone with a knife, and holding someone hostage. 
The dataset exhibits extensive variations in environments, time periods, and diverse challenging weather conditions.

UAV-captured datasets also serves for MC-VAD in marine environments.
The Seagull dataset~\cite{seagull} comprises 19 video sequences that feature various types of objects observed in marine environments, such as cargo ships, small boats, sailing yachts, and leaked floating oil. These objects are recorded using visible light, Long Wave Infra-red (LWIR), and hyperspectral sensors. 
The flight altitude for surveillance is between 150 and 300 meters above the ocean surface to ensure a broad coverage. 
The objects captured are small and the varying lighting conditions along with sea surface reflections make object detection challenging.

The Seadronesee dataset~\cite{seadronesee} encompasses over 54K frames of 20 test subjects in open water, captured under various lighting conditions across multiple days. 
The annotation categories include swimmer (person in water without a life jacket), floater (person in water with a life jacket), life jacket, swimmer (person on a boat without a life jacket), floater (person on a boat with a life jacket), and boats. 
Seadronesee can be used for the study of several related tasks, including object detection, single-object tracking, and multi-object tracking.

\begin{table*}
\caption{Summary of relevant MC-VAD methods. 
Each class is organized according to traditional image processing methods (such as geometry) and deep learning-based methods.}
\vspace{-3mm}
\label{tab3}
\tabcolsep 3pt
\resizebox{\textwidth}{!}{%
\begin{tabular}{p{3cm}p{8cm}p{2cm}p{4cm}}
\toprule
Methods & Descriptions & Traditional methods & Deep learning-based methods \\
\midrule
Semantic segmentation & Based on scene context, and often used for scene-level appearance anomaly detection such as obstacle detection, and abandoned object detection 
& \cite{pinggera2016lost}
\cite{singh2020lidar}
\cite{8809907}
& 
\cite{mcmahon2017multimodal}
\cite{8639077}
\cite{8793588}
\cite{roadanomal}
\cite{ohgushi2020road}
\cite{xia2020synthesize}
\cite{vzust2022temporal}
\cite{GMRPD}
\\
Reference frames & Based on reference videos for each specific scene, and often used in known environments with minor background changes & 
\cite{TIP2010}
\cite{7533104}
\cite{VDAO18MSSP}
\cite{VDAO18TCS}
\cite{VDAO20TIP}
&
\cite{patrolrobot16,patrolrobot17}
\cite{Zaham}
\\
Background separation & Based on foreground extraction or spatial attention for eliminating background noise, and employed for detecting anomalies of pedestrians and vehicles
& 
\cite{PAMI2001}
\cite{6393573} 
&
\cite{ICRA17}
\cite{Eyeinsky2018}
\cite{8575414}
\cite{ionescu2019object}
\cite{morais2019learning}
\cite{liu2019exploring}
\cite{coppola2020social}
\cite{9093633}
\cite{9157616}
\cite{srivastava2021recognizing}
\cite{georgescu2021anomaly}
\cite{he2021obstacle} \\

Object trajectories & Based on object detection and tracking, and trajectory prediction and analysis &  &
\cite{DAD}
\cite{alahi2016social}
\cite{8578539}
\cite{yagi2018future}
\cite{HEVI}
\cite{rasouli2019pie}
\cite{malla2020titan}
\cite{qiu2022egocentric}
\cite{DoTA}
\cite{9714213}\\
Object interactions & Based on spatio-temporal contextual clues such as object location relationships to detect object-to-object and object-to-scene interactions & \cite{RE-DID}
&
\cite{CCD}
\cite{9304822,9341018}
\cite{9197057}
\cite{9672160}
\cite{fang2022traffic}
\cite{10068772}
\cite{10030193}
\cite{9423525}
\\
\bottomrule
\end{tabular}
}
\end{table*}

\subsection{Discussion}\label{sec:dataset_discussion}

We provided a thorough coverage on the datasets that can of use for MC-VAD research. Note that not all datasets mentioned in this section are specifically designed for anomaly detection. Some are originally designed for tasks such as object detection or semantic segmentation. 
However, they possess typical MC-VAD characteristics, including complex dynamic backgrounds, camera motion, and multiple detection categories, that align with the definition of MC-VAD provided in the introduction. 
As illustrated in Tabs.~\ref{tab2} \& \ref{tab1}, datasets targeting different application domains exhibit significant variations in their characteristics.

Underwater datasets primarily focus on object detection and semantic segmentation tasks~\cite{SUIM2020}.
These datasets often present challenges related to motion blur, low underwater clarity, and limited field of view.
On the other hand, datasets recorded on the water surface primarily focus on object detection tasks~\cite{MODS2022}.
The typical challenges associated with these datasets are diverse weather conditions and sea surface reflection.
Datasets captured by cameras moving on the ground generally offer a first-person perspective, being recorded from dashcams and collected from the internet.
One of the main tasks associated with such datasets is the detection of road accidents~\cite{DoTA}.
These datasets often present challenges due to complex and dynamic backgrounds, unpredictable camera motion, and significant occlusions.
Additionally, these datasets may include multiple sensory modalities, which could enhance the accuracy of anomaly detection.
Datasets captured by UAVs typically offer a high third-person viewpoint. 
In the task of human behavior analysis~\cite{li2021uav}, most existing UAV datasets are designed for target detection and/or tracking, leading to a scarcity of abnormal samples. 
Datasets that do feature abnormal behaviors only contain a few instances~\cite{minidrone,Eyeinsky2018}, or the abnormal behaviors are insufficiently realistic~\cite{li2021uav}, thereby making them impractical for MC-VAD studies.

Finally, we observe an unbalanced distribution in the available datasets, a disparity attributable to differing interest levels within their respective application domains. 
For example, within the intelligent transport domain, there exists a significant quantity of real-world datasets, obtained either from dashcams or cameras mounted on vehicles. Conversely, in other domains, such as marine environments, the availability of datasets is considerably less.
The infrequency of real anomalous videos in certain application domains also hampers respective research and development efforts. 
Despite growing demand for urban surveillance with moving cameras, real-world anomalous events such as crime, violence, and fires are challenging to capture with cameras mounted on mobile robots.

\section{Approaches}\label{sec:Approaches}

We categorize MC-VAD approaches into \textit{scene-level}, which focus on detecting frame-level anomalies, and \textit{scenario-level}, which focus on detecting temporal-level anomalies.

\textit{Scene-level} anomalies are the result of deviations from normal patterns that can be detected in a single video frame. 
This category includes appearance-only anomalies, or collective anomalies that do not require temporal information. 
Appearance-only anomalies can be the existence of an unknown object not seen during training or a known object in an unusual location.
Examples of these include novel objects in underwater exploration, e.g.~sunken ships or statues~\cite{zhou2022discovering}, or a bottle in the middle of the road~\cite{GMRPD}. 
Collective anomalies can be crowds in public places or events with people flowing in certain directions (e.g.~parades) \cite{Lawal2017, ullah2021multi, felemban2021deep}.
Crowd counting and density estimation techniques can be exploited for detecting anomalies in these types of scenarios~\cite{gouiaa2021advances, ptak2022board}.

\textit{Scenario-level} anomalies result from unusual movements or unexpected events that necessitate temporal information for detection.
We further categorize scenario-level anomalies into short-term motion anomalies, long-term trajectory anomalies, and contextual anomalies.
Short-term motion anomalies refer to unusual object movements over a short duration, identified by analyzing optical flow and local information of moving objects. 
An example might be a car suddenly stopping at an intersection due to a collision~\cite{HTA}.
Long-term trajectory anomalies pertaining to unusual object trajectories over a longer duration, such as a person suspiciously wandering within a parking lot~\cite{minidrone}.
Detecting such anomalies can necessitate long-term object tracking, which poses a challenge even with static cameras~\cite{Beery_2020_CVPR}, and more so with moving cameras~\cite{wang2016avss,wang2023you}.
Indeed, we note a paucity of literature addressing long-term trajectory anomalies.
Contextual anomalies are not exclusively based on the motion or trajectory of objects but rather on the overarching context in which objects interact and behave.
Methods to detect contextual anomalies may involve analyzing object location relationships, object interactions, or other aspects of semantic scene understanding to identify abnormal patterns.
Moreover, collective anomalies can also form part of scenario-level anomalies as they might result from crowd intersections.
These anomalies can be detected based on the analysis of individual motions, such as oscillatory features, trajectory features \cite{khan2019congestion, basalamah2023deep}, or the temporal accumulation of intersecting motions \cite{Poiesi2015}.

\begin{figure*}[t]
    \centering
    \includegraphics[width=.9\textwidth]{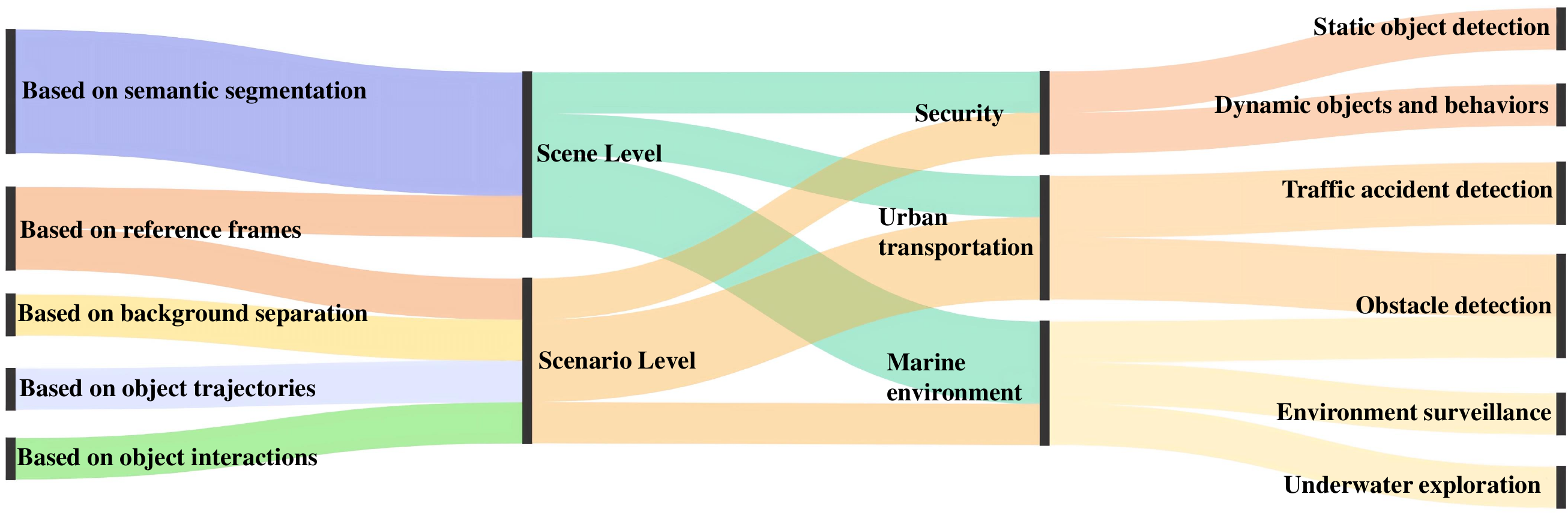}
    \vspace{-3mm}
    \caption{Summary of the correspondence between MC-VAD approaches, anomaly levels, field of application and specific tasks. 
    The line width indicates the quantity of work in each category.}
    \label{sankey}
\end{figure*}

MC-VAD methods can be further classified into five categories based on the type of anomaly (Tab.~\ref{tab3}). 
Two out of these five categories belong to scene-level anomaly detection methods that are semantic segmentation-based and reference frame-based. 
The other three belong to scenario-level anomaly detection methods that are background separation-based, object trajectory-based, and interaction-based. 
Among them, the reference frame-based method can also handle some scenario-level anomalies.
Tabs.~\ref{tab4}-\ref{tab8} provide a summary of relevant methods within these categories, specifying the datasets each method employs for evaluation.
In order to provide a comprehensive overview of our analysis, we present a categorization of methods based on anomaly level, application domains, and tasks in Fig.~\ref{sankey}.
It is worth noting that both scene- and scenario-level anomalies are present across all three application domains, which include security, urban transportation, and marine environments. 
However, the proportion of research works significantly varies among different fields, as indicated by the line width in Fig.~\ref{sankey}.
In the following sections, we provide a detailed description of the pertinent methods within each of these categories.

\subsection{Based on semantic segmentation}

Semantic segmentation entails the process of assigning a specific category label, chosen from a predefined set of possibilities, to each pixel in an image. 
VAD approaches that utilize semantic segmentation are predominantly used for detecting scene-level anomalies, such as obstacle detection, across a range of application domains.

In a marine environment, the work by Cane and Ferryman~\cite{8639077} evaluates three deep segmentation networks, namely ENet~\cite{paszke2016enet}, ESPNet~\cite{mehta2018espnet}, and SegNet~\cite{badrinarayanan2017segnet}, for the detection of typical objects such as large ships, speedboats, sailing boats, and buoys. 
These segmentation networks are trained on the ADE20K \cite{zhou2017scene} ocean dataset in a supervised manner and are tested on four maritime surveillance datasets, namely MODD1 \cite{kristan2015fast}, SMD \cite{SMD}, IPATCH \cite{ipatch}, and Seagull \cite{seagull}. 
Although the deep segmentation networks have shown promising results in terms of marine object detection and segmentation, marine environments present a unique challenge due to frequent lighting condition variations, leading to numerous false positives.
Recent work in marine obstacle detection incorporates temporal information to eliminate false positives caused by sun reflection \cite{vzust2022temporal}.
Multimodal information can also be leveraged to mitigate undesired false detections.
The method proposed by Arain et al.~\cite{8793588} combines feature-based stereo matching with learning-based segmentation to produce a more robust obstacle map. 
This method enhances the binary obstacle map with vision-based distance classification (near, mid, and far) and the depth information from sparse stereo matching, achieving an improved obstacle detection with fewer false positives.

In urban scenarios, multiple modalities are often exploited alongside visual inputs to improve anomaly detection accuracy. 
For example, stereo vision can be used for detecting small obstacles on the road. 
The detection task can be formulated as a statistical hypothesis testing problem on image data applied to the segmentation map of the road \cite{pinggera2016lost}.
LiDAR can provide additional context for monocular segmentation networks in the form of a confidence map, thereby improving detection performance. 
The method proposed by Singh et al.~\cite{singh2020lidar} employs the Hausdorff distance for calibration refinement over extrinsic parameters, achieving accurate calibration between LiDAR and the camera.
The discrepancy between abnormal and normal regions can be enhanced by fusing multi-modal and multi-scale features \cite{GMRPD}.

Recent methods also combine semantic segmentation with generative models for anomaly detection~\cite{roadanomal,ohgushi2020road,xia2020synthesize}. 
The main intuition behind this is as follows: since semantic segmentation models are trained with known classes, when they are tested with images containing unknown classes, these unknown regions would be inaccurately segmented (misclassified as other known classes, either foreground or background or over-segmented) in the predicted semantic map. 
If one resynthesizes the image based on the predicted semantic map, the regions of the unknown classes in the generated image would exhibit a large discrepancy compared to the original test image. 
This discrepancy can be exploited for anomaly detection \cite{roadanomal}. 
Several recent works follow this framework for anomaly detection. 
For example, the method presented by Di Biase et al.~\cite{9578249} introduces a method for integrating different uncertainty measures, such as softmax entropy, softmax difference, and perceptual differences, to improve the discrepancy in differentiating between the input and generated images.
Moreover, the work by Vojir et al.~\cite{10030141} further exploits the property that the anomalous object is not from a class that could be modeled and is also dissimilar to normal objects in appearance. 
This method employs an embedding bottleneck (a small embedding network that takes the backbone features and transforms them into an embedding vector) to model the normal class in an explicit embedding space using augmented training data by synthesized anomalies. 
It also introduces image-conditioned distance features that allow known class identification in a nearest-neighbor manner on-the-fly, and an inpainting module to model the uniqueness of detected anomalies. 
This effectively reduces false positives by filtering out regions that are similar to their neighborhood.

In summary, segmentation-based methods exploit pixel-level labels for the detection and localization of scene-level anomalies, particularly for tasks such as obstacle detection and the recognition of anomalous or novel objects. 
These methods primarily focus on single image frames, not utilizing temporal information, which can lead to inaccurate detections in cases where cameras move under various lighting conditions and complex backgrounds.
Additionally, training these methods may require a substantial amount of annotated datasets, necessitating efforts associated with dataset collection and annotation.
Tab.~\ref{tab4} provides a summary of these methods.

\begin{table*}
\caption{Relevant methods based on semantic segmentation.
For each method we indicate the dataset used for evaluation and whether the dataset is publicly available.}
\vspace{-3mm}
\label{tab4}
\tabcolsep 3pt
\resizebox{\textwidth}{!}{%
\begin{tabular}{p{1.5cm}p{4cm}p{3.5cm}p{4cm}p{7cm}}
\toprule
References & Descriptions & Domain & Dataset/availability & Remarks \\
\midrule
\cite{pinggera2016lost} & 
Detecting small obstacles in road scenes using stereo vision & 
Obstacle detection & 
Lost\&Found/$\checkmark$ & 
Traditional multi-view geometry-based method to operate in real-time. 
Small obstacles are defined as objects with a height between 5cm and 25cm above the ground. \\
\cite{8639077} & 
Detecting marine objects based on semantic segmentation & 
Maritime surveillance & 
ADE20k/$\checkmark$ & 
Test sequences are selected in MODD, SMD, IPATCH, and Seagull datasets with different viewpoints (from water surface to high altitude). \\
\cite{8793588} &
Detecting underwater obstacles based on the combination of feature-based stereo matching and learning-based segmentation &
Underwater obstacle detection &
John Brewer Reef Dataset/$\times$ &
Obstacles are defined as any object (e.g.~seafloor, coral) that hinders robot navigation. \\
\cite{roadanomal} &
Detecting road anomalies based on semantic image synthesis & 
Obstacle detection & 
Lost\&Found/$\checkmark$, \newline Road Anomaly/$\checkmark$ &
The problem of detecting unknown classes is formulated as identifying poorly-synthesized image regions. \\
\cite{singh2020lidar} &
Detecting small road obstacles based on multiple modalities (LiDAR and monocular RGB) &
Obstacle detection & 
Small Obstacle Dataset/$\checkmark$ &
The method can detect small obstacles (as small as 15cm high) within the range of 50 meters. \\
\cite{xia2020synthesize} &
Detecting anomalies based on the difference between a generated synthetic image from the segmentation map and the input image & 
Failure detection and anomaly segmentation &
Cityscapes/$\checkmark$, \newline StreetHazards Dataset/$\checkmark$, \newline Pancreatic Tumor Segmentation Dataset/$\checkmark$ &
Evaluation carried out in both autonomous driving and medical imaging applications. \\
\cite{vzust2022temporal} & 
Detecting marine obstacles based on temporal context-assisted segmentation & 
Marine obstacle detection & 
Mastr1478/$\checkmark$, MODS/$\checkmark$ &
The temporal information in consecutive frames is used to reduce the influence of water turbulence and sunlight flicker on obstacle segmentation. \\
\cite{GMRPD} &
Detecting road anomalies based on dynamic data-fusion model &
Obstacle detection &
GMRP/$\checkmark$, KITTI Road/$\checkmark$, and KITTI Semantic Segmentation Dataset/$\checkmark$ &
Includes a benchmark for large-scale mobile robot road anomaly detection. \\
\bottomrule
\end{tabular}%
}
\end{table*}

\subsection{Based on reference frames}

\begin{figure}[t]
    \centering
    \includegraphics[width=1\columnwidth]{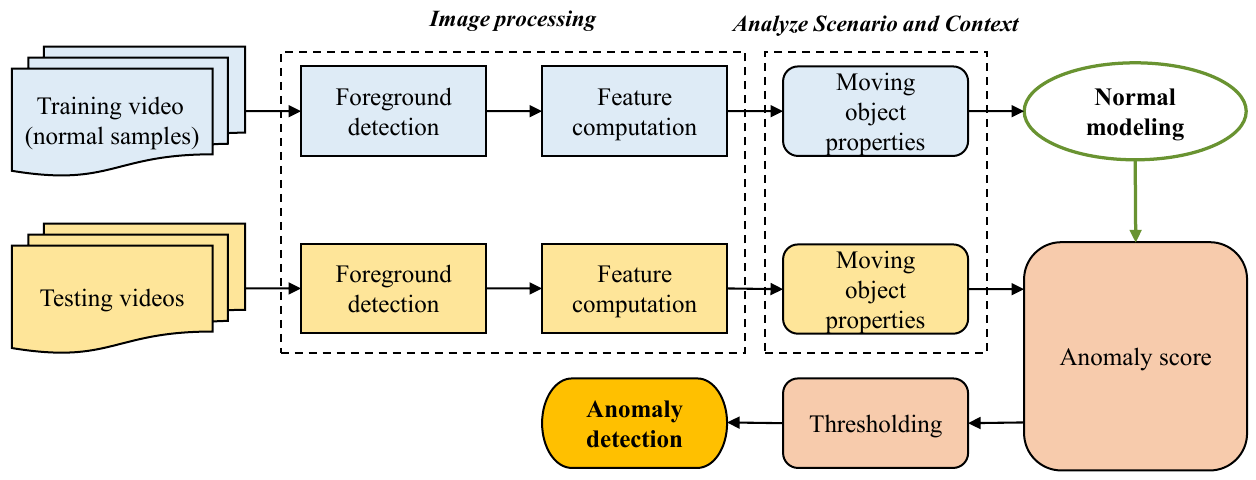}
    \vspace{-5mm}
    \caption{General pipeline of approaches based on reference frames for MC-VAD, including video sequence and geometric alignment before frame comparison. 
    Reference and target videos are input into the pipeline at the same time, and video alignment is carried out after preprocessing, including temporal alignment and spatial alignment. 
    The temporal alignment involves matching between frames in the video sequence.
    Spatial alignment involves pixel-level matching in a single frame. 
    Matched images are post-processed to eliminate false positives, and the abnormal regions are extracted by contrast to achieve anomaly detection.}
    \label{reference-based}
\end{figure}

Approaches based on reference frames are typically developed for detecting anomalies in specific or known environments, where normality is well-defined, such as abandoned object detection in industrial settings~\cite{VDAO, VDAO18MSSP, VDAO18TCS, VDAO20TIP}, obstacle detection on railway tracks~\cite{7533104}, or robot patrolling \cite{TIP2010, Zaham, patrolrobot16, patrolrobot17}. 
These approaches typically capture and record normal images of the environment beforehand to detect any anomalies that deviate from the recorded normal scenes. 
Fig.~\ref{reference-based} illustrates how a typical reference frame-based method works: it detects abnormal objects by aligning the reference video sequence with the target video sequence. 
The reference video is captured by a moving camera when the scene is in its normal state, i.e.~no suspicious object is present in the scene, while the target video is captured by a camera moving along the same route that may contain unexpected objects. 
As the background and viewpoint vary due to the camera's ego-motion \cite{yazdi2018new}, methods in MC-VAD require video pre-processing to spatio-temporally align videos in order to locate the differences among the frames of the target videos and those of the reference videos.

Reference frame-based methods can be used for detecting both scene-level and scenario-level anomalies, depending on the application scenario. 
At the scene level, temporal information is primarily used for frame alignment, while the anomaly itself is independent of time. 
Conversely, scenario-level anomalies are dependent on temporal information. 
For example, abandoned luggage in an airport may not be considered anomalous upon initial detection, but if it remains unattended for an extended period, it may raise security concerns and be classified as an anomaly~\cite{luna2018abandoned}.

The work by Kong et al.~\cite{TIP2010} follows such an alignment-detection framework for detecting non-flat abandoned objects. The process begins with a rough alignment achieved using GPS information to find corresponding frame pairs. 
Then, four steps are carried out:
(i) Inter-sequence geometric alignment based on homographies~\cite{bookhartley2003multiple}, which is computed by a modified random sample consensus (m-RANSAC)~\cite{andrew2001multiple} with Scale-Invariant Feature Transform (SIFT) descriptors~\cite{lowe2004distinctive}. 
This step aims to find all potential suspicious areas.
(ii) Intra-sequence geometric alignment is then employed to eliminate false alarms caused by high objects. 
The parameter selection is determined by the speed of the camera's ego motion.
(iii) A local appearance comparison is performed between two aligned intrasequence frames to eliminate false alarms in flat areas.
(iv) Lastly, a temporal filtering step is used to confirm the existence of anomalous objects.

There also exist methods that do not require temporal alignment between the test and reference video sequences. They rely solely on video content analysis without using other modalities, such as GPS. 
For instance, the method proposed by Thomaz et al.~\cite{VDAO18TCS} employs sparse video decomposition, which has been used in SC-VAD~\cite{candes2011robust, bian2015bi}, for detecting anomalies in moving-camera videos without video synchronization. 
This work represents both the reference and target videos as a low-rank projection on a union of subspaces with a sparse residual term that is used for the detection of anomalies.
Such methods assume that the camera trajectories during the target and reference video acquisitions are similar. 
However, in real-world scenarios, this condition might not be satisfied. 
To tackle this problem, the paper by Jardim et al.~\cite{VDAO20TIP} presents a domain-transformable sparse representation for moving camera videos (mcDTSR), where an additional non-linear domain-transformation term is introduced to cope with the camera motion difference between reference and target acquisition~\cite{peng2012rasl}. 
The introduced domain transformation reduces the anomaly detection error as it allows a better alignment between the reference video frame and the target video frame.

The work by Lawson et al.~\cite{patrolrobot16} involves a patrol robot that autonomously detects aberrant objects along a fixed path. 
Both the reference and target videos are divided into patches using a fixed-size grid for each frame. 
A dictionary is then formed to represent the current environment, with features extracted from patches using AlexNet. 
As the patrol can cover various scenes where the anomaly can be defined differently, a deep network model~\cite{placeCNN} is used to recognize the robot's environment to condition the dictionary look-up. 
In a subsequent work~\cite{patrolrobot17}, a generative model, DCGAN~\cite{DCGAN}, is trained using these patches, and the bottleneck features of the generated patches and actual patches are compared to detect anomalies.

In summary, reference frame-based methods can serve for anomaly detection within specific environments. 
They are particularly effective in cases where scene changes are minimal, such as unmanned workshops or industrial parks. 
In these scenarios, autonomous patrolling can be used for anomaly detection, as normal conditions can be predefined and recorded in reference frames. 
Initially, the accuracy of these methods significantly relied on the alignment accuracy between reference frames and target frames. 
However, advancements in deep learning techniques have gradually reduced this dependency, expanding their applicability across various scenarios.
Tab.~\ref{tab5} provides a summary of these methods.

\begin{table*}
\caption{Relevant methods based on reference frames. For each method we indicate the dataset used for evaluation and whether the dataset is publicly available.}
\vspace{-3mm}
\label{tab5}
\tabcolsep 3pt
\resizebox{\textwidth}{!}{%
\begin{tabular}{p{1.5cm}p{5cm}p{2.5cm}p{3cm}p{7.5cm}}
\toprule
References & Descriptions & Domain & Dataset/availability & Remarks\\
\midrule
\cite{TIP2010} & Detecting non-flat abandoned object with a moving camera by using the geometric alignment &
Abandoned object detection & 
Proprietary dataset/$\times$ & 
RANSAC-based homography alignment to detect suspicious areas, an intra-sequence geometric alignment to remove false alarms caused by high objects, a local appearance comparison to remove false alarms in flat areas and a temporal filtering step to confirm suspicious objects.
 \\
\cite{patrolrobot16,patrolrobot17} &
Locating abnormal objects for robot patrol, a scene dictionary is constructed by clustering deep neural network features &
Anomalous object detection &
Proprietary dataset/$\times$ &
Image patches extracted with a fixed-sized grid and processed by a deep network. 
Deep features of the patch are incorporated into the environmental dictionary through K-means clustering. \\
\cite{7533104} &
Detecting obstacles using image subtraction between the input and reference train frontal view &
Railway tracks obstacle detection &
Proprietary dataset/$\times$ &
Temporal alignment to find frame-by-frame correspondences of current and reference sequence, spatial alignment to register pixel-wise image, and subtraction methods are applied to detect obstacles. \\
\cite{VDAO18TCS} &
Detecting anomalies in video sequences obtained from slow-moving cameras based on a family of sparse decomposition algorithms &
Abandoned object detection &
VDAO/$\checkmark$ &
Representation of video data as a low-rank projection on a union of subspaces plus a sparse residue term. \\
\cite{VDAO20TIP} & 
Detecting anomalies in video sequences from moving cameras through matrix factorization based on sparse representation &
Abandoned object detection &
VDAO/$\checkmark$ &
Geometric domain-transformations are used to model the effect over the samples caused by the camera movement and rotation, a transformation term in the sparse representation procedure is included for optimization. \\
\cite{Zaham} & 
Detecting anomalies for moving surveillance robots based on a change detection model using a Siamese network. &
Anomalous object detection & 
Proprietary dataset/$\times$ &
Geo-tag, camera Pitch and camera Yaw (GPY) information are used for normalcy definition image retrieval. Human feedback enables the system to learn continuously. \\
\bottomrule
\end{tabular}
}
\end{table*}

\subsection{Based on background separation}

\begin{figure}[t]
    \centering
    \includegraphics[width=1\columnwidth]{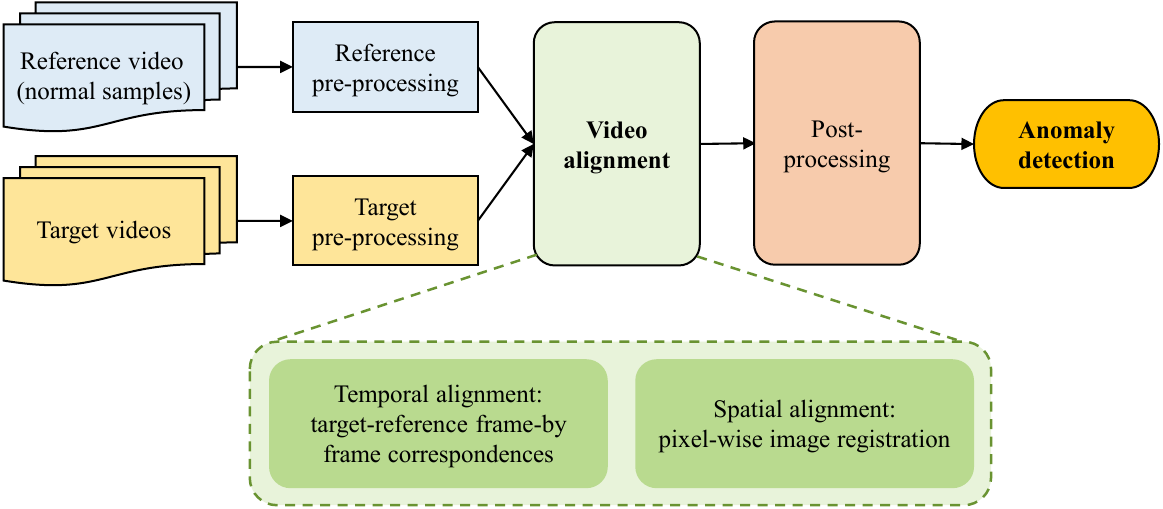}
    \vspace{-5mm}
    \caption{The general framework of background separation-based approaches for MC-VAD.
    Dashed lines indicate that in-box steps may require additional considerations under multi-scenario conditions. In the training phase, the training video containing only normal samples is input, the moving objects are extracted through foreground detection, and the features of moving objects are calculated. Without scene and context analysis, the normal model is established by directly calculating the features of moving objects in a single scene. In the testing phase, the testing video containing abnormal samples is input, processed in the same way as in the training phase, and the features of moving objects are calculated based on the normal model. The resulting deviation is considered the anomaly score for anomaly detection.}
    \label{detection-based}
\end{figure}

Approaches based on background separation focus on analyzing the motion of foreground objects \cite{ICRA17}, or employ spatial attention mechanisms~\cite{ullah2021attention, DAD, zhou2019attention} to detect abnormal behaviors. 
Such methods can help mitigate the background-bias phenomenon \cite{liu2019exploring}, i.e.~deep neural networks are inclined towards learning the background information, rather than the patterns of anomalies, for the recognition of abnormal behaviors.
In the context of SC-VAD, the background is typically static, thus motion-based foreground detection using pixel-level differences between foreground and background can be employed. 
However, in MC-VAD, motion-based foreground detection is no longer applicable as the cameras are in motion, introducing a dynamically changing background.
Therefore, techniques that use object detection to separate the foreground from the background are more common \cite{ren2015faster,he2017mask,redmon2018yolov3}.
Depending on the application, these methods can analyze foreground regions independently \cite{ICRA17,srivastava2021recognizing,9706619}, or jointly when interactions among objects are being considered \cite{Eyeinsky2018,coppola2020social,noghre2023understanding}. Fig.~\ref{detection-based} illustrates the general framework of background separation-based approaches in MC-VAD. 

Typical VAD methods tend to attribute anomalies to the actions of moving objects~\cite{UCSD}. 
As such, after detecting these objects, they extract key features from object bounding boxes such as height, width, velocity, direction of motion, and relative distances \cite{PAMI2001}. 
To further enhance the comprehension of activities, additional features like human posture can be also used \cite{Eyeinsky2018,li2021uav}.

\begin{table*}
\caption{Relevant methods based on background separation. 
For each method we indicate the dataset used for evaluation and whether the dataset is publicly available.}
\vspace{-3mm}
\label{tab6}
\tabcolsep 3pt
\resizebox{\textwidth}{!}{%
\begin{tabular}{p{1.5cm}p{5cm}p{2.5cm}p{4cm}p{7.5cm}}
\toprule
References & Descriptions & Domain
& Dataset/availability & Remarks\\
\midrule
\cite{PAMI2001} &
Detecting anomalies through behavior analysis of moving objects captured from airborne moving platforms &
Aerial surveillance &
Proprietary dataset/$\times$ & 
Moving regions in the sequence are detected and tracked. 
Ego-motion estimation is used to compensate for the motion of the observer.\\
\cite{ICRA17} &
Detecting abnormal human behavior based on real-world environments of autonomous moving robots &
Abnormal behavior detection &
UCF101/$\checkmark$, HMDB/$\checkmark$, Non-biased Background Dataset/$\times$, Moving Camera Dataset/$\times$ & 
Human action in unconstrained videos is located through generic action region proposals. 
Motion features are extracted from action proposals by a ConvNet framework \\
\cite{Eyeinsky2018} &
Identification of violent individuals using proposed real-time drone surveillance system &
Violence detection &
Aerial Violent Individual/$\times$ & 
Feature pyramid network is used to detect the humans. 
ScatterNet Hybrid Deep Learning network is proposed for human pose estimation. 
Orientations between limbs are used to identify violent individuals. \\
\cite{coppola2020social} &
Detecting potential hazards by identifying human social interactions in continuous RGB-D video sequences obtained by service robots &
Behavior understanding &
Social Activity Dataset/$\checkmark$ & 
Extract skeleton and compute features from RGB-D sequences. 
Individual poses, movements, and spatial relations of features are used to classify and detect social activities such as fights, talk, and handshake. \\
\cite{srivastava2021recognizing} &
Recognizing human violent action using drones from different heights for an unconstrained environment &
Violence detection &
Proprietary dataset/$\times$ & 
Extract key-point and generate 2D skeletons for the person. 
Support Vector Machine and Random Forest are used for classification to recognize actions. \\
\bottomrule
\end{tabular}
}
\end{table*}

During the model learning phase, a normal model is constructed to represent a normal situation by incorporating scene analysis and contextual information while extracting features. 
In the case of a single scene, a singular normal model can be directly established \cite{A3D}.
In the case of multiple scenes, the relationship between different features and corresponding scenes is typically required \cite{PAMI2001,ICRA17}.
At inference time, moving object features (such as appearance, pose and motion trajectories) are extracted, and the deviation between these features and those extracted from normal model is used for anomaly detection. 
These deviations are represented can be the prediction error and reconstruction error. 
Additionally, methods such as clustering or Support Vector Machine (SVM) can also be employed for feature classification to detect anomalies.

Recognizing and understanding human activity is essential for a wide variety of MC-VAD applications from security surveillance purposes~\cite{Eyeinsky2018} to having safe and collaborative interaction between humans and robots in shared workspaces~\cite{collarobot17}.
The work by \cite{ICRA17} leverages the concept of motion boundaries~\cite{7298873} to select action regions (i.e.~interested area only contains human activity) to eliminate irrelevant regions. Convolutional neural networks are utilized to extract features within the action regions, followed by action classification using SVM. 
During the model learning phase, the model requires identifying the place as well as the action being performed by the human. 
Then the prior probabilities of actions occurring in different scenes are considered to determine whether an action falls under abnormal behavior.

Due to the articulated structure of the human body, which enables complex movements across multiple degrees of freedom, obtaining key features solely from bounding boxes may not effectively capture this advanced movement, thus pose estimation algorithms are employed to extract key points and skeletal features. These features are then used for feature computation, offering a more concise and compact representation compared to pixel-based methods. Consequently, this approach has garnered considerable attention in the field of anomaly detection in recent times~\cite{morais2019learning,9157616,luo2021normal,noghre2023understanding}.
The work by Singh et al.~\cite{Eyeinsky2018} and Srivastava et al.~\cite{srivastava2021recognizing, srivastava2022uav} achieves violent individual identification in an unmanned aerial vehicle surveillance system by utilizing human detection and pose estimation algorithms to compute key point features. 
Coppola et al.~\cite{coppola2020social} extract skeleton and compute features from continuous RGB-D video sequences captured by a mobile service robot to detect human interactive behaviors such as handshake, help walking, fighting and talking.

In summary, background-based methods in MC-VAD require object detection in order to detect foreground objects. 
In traffic scenes, anomalies are often associated with accidents, which can be detected by analyzing motion patterns.
In security surveillance scenes, direct analysis of skeletal features in human bodies, as opposed to frame-level detection, allows for the inclusion of more compact and valuable information. 
It is important to note that these methods inherently involve some level of prior knowledge regarding the specific types of anomalies present in the application. 
Therefore, they are better suited for specific applications or scenarios. 
These methods heavily rely on the accuracy of foreground detection. 
In situations where occlusions or crowds are present, the accuracy of anomaly detection can be affected. Tab.~\ref{tab6} provides a summary of these methods.

\begin{table*}
\caption{Relevant methods based on object trajectories. 
For each method we indicate the dataset used for evaluation and whether the dataset is publicly available.}
\vspace{-3mm}
\label{tab7}
\tabcolsep 3pt
\resizebox{\textwidth}{!}{%
\begin{tabular}{p{1.5cm}p{4.7cm}p{3.8cm}p{2.5cm}p{7.5cm}}
\toprule
References & Descriptions & Domain &Dataset/availability &Remarks\\
\midrule
\cite{DAD} &
Anticipating accidents in dashcam videos based on a Dynamic-Spatial-Attention (DSA) Recurrent Neural Network (RNN) &
Traffic accident anticipation &
DAD/$\checkmark$ &
Early work that uses crowdsourced dashcam accident datasets for traffic accident anticipation. \\
\cite{8578539} &
Accident prevention through joint prediction of on-board camera ego-motion and pedestrian trajectories &
Trajectories prediction &
Cityscapes/$\checkmark$ & 
Model consists of two specialized streams, the odometry stream to predict future vehicle odometry sequence, and the bounding box estimation stream to predict odometry distributions conditioned to pedestrian trajectories. 
A Bayesian RNN encoder-decoder architecture is used to compute model epistemic and aleatoric uncertainties. \\
\cite{HEVI} &
Localizing future vehicles by simultaneously predicting location and scale from egocentric views &
Trajectory prediction &
HEV-I/$\checkmark$ &
Multi-stream recurrent neural network (RNN) encoder-decoder model to capture both object location and scale, and pixel-level observations to predict vehicle locations. \\
\cite{DoTA} &
Detecting traffic video anomalies based on future object localization & 
Traffic accident detection &
DoTA/$\checkmark$ &
Large scale benchmark for traffic anomaly detection is proposed including a new dataset (DoTA), evaluation metrics (spatial-temporal area under curve), and online video action detection methods. \\
\cite{qiu2022egocentric} &
Forecasting egocentric human trajectory with a wearable camera and multi-modal fusion  &
Trajectory prediction &
TISS/$\checkmark$ &
Three different modalities (past ego trajectory, past trajectories of nearby people, and the environment such as the scene semantics or the depth of the scene) to forecast the trajectory of the ego-camera. \\
\cite{9714213} &
Detecting accidents based on spatio-temporal feature encoding with a multilayer neural network &
Traffic accident detection &
DoTA/$\checkmark$ &
Coarse detection based on temporal coding locates potential accident frames, and fine detection based on spatial coding detects accident frames. \\
\bottomrule
\end{tabular}
}
\end{table*}

\subsection{Based on object trajectories}

Approaches based on object trajectories focus on constructing a normal model using trajectories derived from collections of videos that only include normal behaviors or events \cite{DAD,8578539,HEVI,A3D,DoTA,qiu2022egocentric,9714213}. 
The trajectory of an object in a test video is then compared with those in the normal model, where any significant deviation is considered as an anomaly. 

Object trajectories are estimated from image sequences captured by the camera, and thus, motion patterns are determined relative to the camera reference system. 
In SC-VAD, where the camera reference system is static, the detected motion is purely that of the moving object, and clustering techniques can be employed to detect these deviations~\cite{Anjum2008, khan2016analyzing, trajectory2017}.
However, in MC-VAD, the measured trajectories from the camera also include the camera's ego motion. It is necessary to differentiate between the motion of the camera and that of the objects of interest in order to only model the normality of object trajectories \cite{HEVI}.
To account for camera ego-motion, some studies use visual odometry to predict the ego-motion, and they incorporate a trajectory prediction module into the framework~\cite{PAMI2001, 8578539, A3D, DoTA}.

Road scenes viewed from aerial moving cameras can be analyzed by using the method proposed by Medioni et al.~\cite{PAMI2001}.
This method initially estimates the drone ego-motion to register consecutive frames, then it carries out multi-object detection and tracking to locate moving objects. 
Given that motion alone is not a sufficient indicator of anomalous activities, the method also takes into consideration the interactions between trajectories.
These trajectory patterns are further combined with user-provided information, such as geospatial context and goal context, to classify and recognize context-aware anomalous activities.

Traffic accidents can be detected from vehicle dashcams using the method proposed by Yao et al.~\cite{A3D}.
This method leverages the past motion trajectory of moving objects to predict their current positions.
These predictions are then compared with the observed frames to identify and flag any unexpected discrepancies. 
As object motion alone is insufficient for identifying anomalies, this method employs a combination of optical flow, moving object bounding box, and camera ego-motion for trajectory prediction. 
Data for the bounding box is supplied by object detection \cite{he2017mask} and multi-object tracking \cite{wojke2017deepsort} techniques, while information about camera motion is obtained via camera ego-motion estimation \cite{mur2017orb}.
The same authors extended this work in \cite{DoTA}, which presents a method for traffic VAD based on an ego-motion Recurrent Neural Network (RNN) encoder-decoder to predict future odometry of the ego-vehicle, and a two-stream RNN encoder-decoder incorporating predicted ego-motion into future object bounding box predictions. 
The same work also introduces a new large-scale traffic VAD dataset named DoTA, with temporal, spatial, and categorical annotations, a new spatio-temporal evaluation metric for VAD, called STAUC, and a comprehensive benchmark of state-of-the-art VAD, video action recognition, and online action detection methods.

Pedestrian motion in traffic scenes can be predicted from vehicle dashcams by using the method proposed by Bhattacharyya et al.~\cite{8578539}.
This method is a two-stream model for long-term prediction of both person bounding box and the vehicle odometry. The bounding box prediction stream, implemented as a Bayesian recurrent neural network encoder-decoder, follows a Bayesian formulation to predict odometry-conditioned distributions over pedestrian trajectories. Such Bayesian formulation also enables the modeling of both epistemic (model) and aleatoric (observation) uncertainties for the bounding box prediction.

In summary, these methods leverage object trajectories for anomaly detection, reducing the dependency on pixel-level differences and background changes. Yet, the performance of the final task, whether it is behavior prediction or analysis, is closely linked to the performance of other processing modules, such as object detection, object tracking, and ego-motion/odometry estimation. 
Tab.~\ref{tab7} provides a summary of these methods.

\begin{table*}
\caption{Relevant methods based on object interactions. For each method we indicate the dataset used for evaluation and whether the dataset is publicly available.}
\vspace{-3mm}
\label{tab8}
\tabcolsep 3pt
\resizebox{\textwidth}{!}{%
\begin{tabular}{p{1.5cm}p{5cm}p{2.5cm}p{3.5cm}p{7.5cm}}
\toprule
References & Descriptions & Horizon & Dataset/availability & Remarks\\
\midrule

\cite{RE-DID} &
Detecting violent behaviors, such as fighting, through the detection and localization of dynamic human interactions in real videos  &
Violence detection &
RE-DID/$\checkmark$ &
Urban fight scene dataset containing camera motion is presented. 
Visual information is extracted in the area associated with the interpersonal space to detect fights. \\
\cite{CCD} &
Anticipating traffic accidents based on uncertainty spatio-temporal relational learning &
Traffic accident anticipation &
CCD/$\checkmark$ &
Graph convolution and recurrent networks are used for relational feature learning. 
Bayesian neural networks are leveraged to address the intrinsic variability of latent relational representations. \\
\cite{9341018} &
Understanding vehicle behavior based on multi-relational graph convolutional network &
Behavior understanding &
Honda Driving Dataset/$\checkmark$,\newline Apollo/$\checkmark$, KITTI/$\checkmark$  &
Dynamic scenes are decomposed into multiple relation interaction graphs. 
Detected lane markings and poles are used as information. \\
\cite{9672160} &
Predicting future collisions in real-world scenario setups using scene graphs &
Traffic accident anticipation &
Honda Driving dataset/$\checkmark$,\newline Scene-graph/$\checkmark$, DoTA/$\checkmark$&
Spatio-temporal scene-graph embedding methodology with Graph Neural Network and Long Short-Term Memory are used for accident anticipation. \\
\cite{fang2022traffic} &
Detecting traffic accidents using appearance, motion and context consistency learning in driving scenarios &
Traffic accident detection &
A3D/$\checkmark$, DADA-2000/$\checkmark$ & 
Fusion technique is used to integrate appearance, trajectories, interactions, and contextual features of driving scenes, which are then processed by graph convolutional networks. \\
\cite{10068772} & 
Anticipating accidents based on a Graph and Spatio-temporal Continuity framework &
Traffic accident anticipation &
DAD/$\checkmark$, A3D/$\checkmark$ &
Graph convolution and recurrent networks are used for feature learning. 
Bayesian neural networks are used to model latent relational representations. \\
\bottomrule
\end{tabular}
}
\end{table*}

\subsection{Based on object interactions}

Analyzing object interactions and spatial relationships can also provide a basis for detecting anomalies. 
Features like relative distance, speed, direction, and interaction patterns between objects can be analyzed to identify abnormal events that deviate from normal behavioral patterns. 
For instance, in traffic surveillance scenarios, a sudden decrease in the distance between two vehicles could indicate a potential collision. 
In the context of the recent COVID-19 pandemic, the interpersonal distance between individuals could be used as a measure to flag violations of social distancing rules~\cite{aghaei2021single, morerio2021icip}. 
Furthermore, interpersonal interactions can also be analyzed to detect abnormal behaviors such as physical violence or fighting~\cite{RE-DID}.
Because interactions are the result of individual motions, approaches based on object interactions often require multi-object tracking as a pre-processing step.

Traffic accident anticipation problems can be addressed through the analysis of vehicle interactions, employing spatial relationship learning with a Graph Convolutional Network (GCN), and temporal relationship modeling with a Recurrent Neural Network \cite{CCD}.
A Bayesian Deep Neural Network can be leveraged to estimate the epistemic uncertainty associated with these predictions.
Alternatively, the same problem can be conceptualized as identifying irregular object movements and sudden changes in spatial relations between different vehicles across sequential frames in dashcam videos \cite{fang2022traffic}.
This approach accomplishes its goal by predicting the visual scene context using a GCN, taking into account the temporal frame consistency, temporal object location consistency, and the spatio-temporal relationship consistency among the participants on the road.

Occlusions in moving camera scenarios can often result in fragmented trajectories or identity switches \cite{Poiesi2013, Poiesi2015a}.
To address this challenge, a graph-based method can be deployed to reconstruct the spatio-temporal states of occluded objects by learning their spatio-temporal relationships~\cite{10068772}.
Object trajectories can then be aggregated over time to facilitate the detection of interactions.

As deep learning-based method output predictions are often difficult to explain, the work by Doshi and Yilmaz~\cite{10030193} introduces an approach for detecting video anomalies grounded in scene graphs.
This approach does not only monitor objects independently but also monitors the interactions between objects to detect anomalous events and provide explanations for their root causes.
Furthermore, this explainable methodology facilitates cross-domain adaptability, allowing for transfer learning in different surveillance scenarios.

In summary, methods based on object interactions in MC-VAD offer the distinctive advantage of providing an additional layer of contextual information through the analysis of relationships between objects. 
These methods are capable of detecting anomalies that might be overlooked by other approaches and often synergize effectively with methods that focus on object trajectories.
When compared with frame-level anomaly detection methods, these techniques can also enable the explainability of anomaly detection.
Moreover, they face challenges such as the limited availability of contextual anomaly datasets and the potential for occlusions in moving camera scenarios to adversely affect the accuracy of anomaly detection. 
Tab.~\ref{tab8} provides a summary of these methods.

\section{Outlook}\label{sec:outlook}

MC-VAD is a recently emerging direction that has gained momentum in recent years. 
Although numerous methods have been developed, they are often scattered. 
In the previous section, we established an intuitive taxonomy framework and interconnected previous research efforts. 
In this section, we summarize the limitations of these works, discuss promising research trends, and explore potential directions for the future.

\subsection{Limitations of existing works}\label{sec:limitations}

Existing MC-VAD algorithms are not without their limitations. 
Firstly, they are often restricted to detecting only a handful of specific anomaly categories, thus curtailing their versatility across a broad spectrum of anomaly types. 
Secondly, they frequently suffer from poor generalizability, implying the need for retraining on each unique target scene, which subsequently increases the computational load on the algorithm. 
They are also particularly vulnerable to a variety of external factors such as changes in illumination, obstructions, and complex background scenarios. 
These factors often result in frequent false positives or false negatives, consequently lowering the precision and reliability of the detection process. 
These combined limitations likely serve as the principal reasons why MC-VAD algorithms have proven challenging to implement on a wide scale.

Specifically, most semantic segmentation methods are only capable of detecting appearance anomalies at the scene level, such as the presence of abnormal objects. They are limited in their ability to handle other types of scene-level anomalies that involve motion and span across consecutive frames. Furthermore, these methods often fail to effectively utilize temporal information, resulting in the loss of abnormal targets in certain frames due to occlusion, light reflections, and other interfering factors. The discarded abnormal information in these frames can be key for tasks that require real-time performance, such as obstacle detection. Recent works have begun exploring the improvement of anomaly segmentation by considering temporal context and reducing false positives.

Methods based on reference frames are limited by the requirement of reference videos that do not contain any anomalies. They are prone to false positives when the scene undergoes significant changes or when normal objects that were not present in the reference frames appear frequently. When the camera undergoes severe motion, it greatly affects the alignment of frames and consequently reduces the accuracy of anomaly detection. However, the methods based on background separation, object trajectory and object interaction are affected by the effects of low-level algorithms such as detection and tracking, and have high requirements for data quality.

\subsection{Recent research trends and future directions}\label{sec:recent_trends_fut_direct}

Based on our analysis of existing works, video anomaly detection in dynamic scenes with moving cameras is a rapidly evolving research field with great potential for improving traffic safety, environmental surveillance, and autonomous vehicle technology. However, due to the inherent challenges of MC-VAD and the limitations of current works, there are still some future research directions that we deem to be interesting to be explored by the community.

\setlist{nolistsep}
\begin{itemize}[leftmargin=*]

\item[\textbullet]\textit{Datasets and benchmark}: Collecting and labeling anomalous events for MC-VAD datasets remains to be challenging due to the rarity and diversity of anomalies in the real world, with even reduced chances acquired by moving cameras. Increasing the deployment of mobile robots for dataset capturing is an option, yet the data quantity is limited. Thus it is crucial to encourage dataset acquisition via various ways including simulation environments~\cite{ubnormal,GridNet,zhou2022discovering,9672160} using game engines and crowd-sourcing~\cite{UCF,DAD,CCD,DoTA} via platforms such as Youtube. In the former, synthetic anomaly data can be generated using game engines in different scenarios, accompanied by precise pixel-level annotations. The latter can provide a large number of real data. 

Along with the datasets, establishing relevant benchmark evaluation metrics will also allow researchers to effectively compare different algorithms and techniques, driving the development of better-performing methods. 
Although most VAD methods calculate anomaly scores for each frame and evaluate performance using area under the curve (AUC) of receiver operating characteristic (ROC) curves, this metric only measures the effectiveness of temporal localization of anomalies while overlooking spatial detection accuracy. This simplistic and incomplete evaluation metric makes it challenging to assess the true capabilities of algorithms and lacks interpretability in terms of results. Recent research has recognized this limitation and proposed more comprehensive evaluation metrics to address these issues~\cite{ramachandra2020street,DoTA}.

\item[\textbullet]\textit{Multi-modal data fusion}: Combining data from multiple sources like RGB video, depth maps, LiDAR point clouds, and thermal images~\cite{Riz2023} can improve the accuracy and robustness of anomaly detection systems, especially to deal with the problems of low image quality and many external interference factors are common in MC-VAD. 
The XD-Violence~\cite{wu2020not} dataset introduced audio information and demonstrated the positive impact of audio-visual fusion on VAD. In practical applications of MC-VAD, the camera is usually closer to the location of anomalies, making it more likely to capture valuable audio information. However, only a few datasets in the field of MC-VAD contain multimodal data, which limits the development of relevant multimodal methods.
Therefore, research can focus on developing multimodal datasets and investigating better fusion strategies that maximize the benefits of each modality.

\item[\textbullet]\textit{Context-awareness}: Contextual information is crucial for MC-VAD, as detecting individual objects or actions without considering the context can be biased or even erroneous since anomalies are defined based on their contextual relevance. Appearance context can identify anomalies based on object appearance, semantic context can enhance understanding of scenes and objects, and temporal context can extract information that is not present in individual frames and help determine if events or actions should occur. Recent work has recognized the importance of context understanding in the field of computer vision and has conducted detailed research on this topic~\cite{WANG2023103646}. In the future, research should focus on incorporating various types of context extensively to broaden the coverage of MC-VAD algorithms in terms of scenes and types of anomalies, while achieving higher robustness.

\item[\textbullet]\textit{Robustness to environmental conditions}: As mobile platforms can move anywhere and anytime, improving the robustness of VAD systems to different environmental conditions like varying lighting, weather or road conditions, becomes extremely important to be applied in adversarial real-world applications. However, the most popular works such as UCSD~\cite{UCSD}, CUHK Avenue~\cite{CUHK} and ShanghaiTech~\cite{ShanghaiTech} simply formulate abnormal events through human simulation and do not contain diverse complex environmental conditions, which makes it difficult for well-trained models to detect real-world anomalies.

\item[\textbullet]\textit{Deep learning and generative models}: Developing or exploring advanced deep learning techniques and generative models to better capture the high complexity of real-world scenarios.
This has been recently shown particularly effective for image classification~\cite{Azizi2023}. In addition, some novel methods such as Vision Language Model (VLM) and diffusion models are utilized for video anomaly detection~\cite{osman2023exploring}. In particular, utilizing pre-trained large-scale VLM allows for a better understanding of abnormal situations in videos and generating corresponding descriptions. It combines the modeling of both image and language information while incorporating features such as multimodal fusion, contextual understanding, handling complex, and providing explanatory descriptions. It is worth noting that researchers have achieved state-of-the-art performance on multiple VAD datasets by using CLIP~\cite{clip2021} to extract deep features, only employing a simple framework~\cite{reiss2022attribute}. 
Additionally, recent work on vocabulary-free image classification can provide valuable prompts as cues for abnormal events under challenging conditions such as severe occlusion and low-quality images \cite{conti2023vocabulary}.

\item[\textbullet]\textit{Explainability and interpretability}: Most VAD methods are based on data-drive end-to-end neural networks. While being effective, the learned representations are often not interpretable, which inhibits the deployment in the real world for safety and security concerns~\cite{10030193}. Developing explainable and interpretable models~\cite{ramachandra2020street, DoTA} provides insights into the underlying reasoning for the detected anomalies, thus making the automatic system easier to be accepted by human end users.

\item[\textbullet]\textit{Real-time processing}: Due to the nature of application scenarios, MC-VAD has a higher requirement for real-time performance, as it often involves prevention and reaction in case of hazards. Thus, it is important to develop efficient algorithms for real-time video anomaly detection that can run on low-resource devices like embedded systems.
Only a few of the methods we analyzed are specifically designed for light-weight computing~\cite{VDAO,Eyeinsky2018}.
This will enable the integration of these systems into real-world applications.

\item[\textbullet]\textit{Continual learning and human collaboration}: Continual Learning (CL), as a special paradigm of machine learning, has gained increasing attention in the field of robotics, where the data distribution and learning objectives vary over time, or where all training data and objective criteria are never available at once~\cite{lesort2020continual}. In the context of MC-VAD, where abnormal events are sparse and the definition of anomalies may change over time, CL can enable anomaly detection systems to continuously adapt and improve, thus offering long-term reliability and accuracy in anomaly detection~\cite{9706619}. Furthermore, updating the definition of normality through human collaboration and preventing the system from repeatedly generating false alarms for the same anomaly is a promising direction for achieving long-term learning~\cite{Zaham}.

\item[\textbullet]\textit{Domain adaptation and transfer learning in open world}: Domain adaptation is still a challenging issue to address to enable VAD models to any new environments and scenarios in the open world. While this can be mitigated if a larger quantity of datasets is available, the advancement in the algorithm level should also be investigated to jointly address this challenge. 

\item[\textbullet]\textit{Privacy and ethics}: It is also important to address privacy and ethical concerns associated with VAD, particularly when it comes to surveillance and data collection in public spaces~\cite{marvel2021}. It is an interesting direction to explore in terms of how to balance the VAD performance when personally identifiable information is removed from the visual inputs, such as faces, poses or gaits of persons, and car-plates of vehicles.  

\end{itemize}
\section{Conclusions}\label{sec:conclusions}

The widespread adoption of mobile cameras and the increased demand for MC-VAD across various domains have propelled the advancement of relevant research at a rapid pace.
However, no review specifically summarizes the research activities and achievements on this particular topic. In this paper, we provide a comprehensive overview on MC-VAD including applications, datasets and methods, making significant contributions to position relevant research in the broad field of VAD.

As a high-level visual understanding task, VAD is often difficult to define and evaluate, resulting in a lack of unity across different works. This survey offers a clear definition of video anomaly detection for mobile cameras and discusses the new challenges, applications, and methods arising from dynamic scenes for MC-VAD. We summarize six types of tasks that span in three most crucial application fields and the corresponding 25 public datasets. Additionally, we categorize and highlight methods that are particularly effective or novel in the context of MC-VAD.

Furthermore, this survey classifies and reviews the types of anomalies included in various datasets. Existing VAD research primarily focuses on specific scenarios, which implicitly assume certain definitions and contextual cues for anomalies. While these assumptions might be clear to researchers in the same field, they may not be readily apparent to readers from other disciplines, leading to confusion. By consolidating and categorizing the anomalies in different datasets, this survey facilitates a more comprehensive understanding of VAD methods and promotes cross-domain communication in this research area. The findings also provide guidance for designing suitable methods for detecting particular types of anomalies. 

Lastly, as mentioned in the future research direction section (Sec.~\ref{sec:recent_trends_fut_direct}), there are still open challenges in the field, where the most urgent one we deem would be establishing proper benchmark datasets for existing applications and develop curated datasets to promote the advancement of the field.

\bibliographystyle{IEEEtran}
\bibliography{ref}

\begin{thebibliography}{100}
\providecommand{\url}[1]{#1}
\csname url@samestyle\endcsname
\providecommand{\newblock}{\relax}
\providecommand{\bibinfo}[2]{#2}
\providecommand{\BIBentrySTDinterwordspacing}{\spaceskip=0pt\relax}
\providecommand{\BIBentryALTinterwordstretchfactor}{4}
\providecommand{\BIBentryALTinterwordspacing}{\spaceskip=\fontdimen2\font plus
\BIBentryALTinterwordstretchfactor\fontdimen3\font minus
  \fontdimen4\font\relax}
\providecommand{\BIBforeignlanguage}[2]{{%
\expandafter\ifx\csname l@#1\endcsname\relax
\typeout{** WARNING: IEEEtran.bst: No hyphenation pattern has been}%
\typeout{** loaded for the language `#1'. Using the pattern for}%
\typeout{** the default language instead.}%
\else
\language=\csname l@#1\endcsname
\fi
#2}}
\providecommand{\BIBdecl}{\relax}
\BIBdecl

\bibitem{surveypami}
B.~Ramachandra, M.~J. Jones, and R.~R. Vatsavai, ``A survey of single-scene
  video anomaly detection,'' \emph{IEEE Transactions on Pattern Analysis and
  Machine Intelligence}, vol.~44, no.~5, pp. 2293--2312, 2022.

\bibitem{DAD}
F.-H. Chan, Y.-T. Chen, Y.~Xiang, and M.~Sun, ``Anticipating accidents in
  dashcam videos,'' in \emph{Proceedings of Asian Conference on Computer Vision
  (ACCV)}, 2016, pp. 136--153.

\bibitem{UCSD}
V.~Mahadevan, W.~Li, V.~Bhalodia, and N.~Vasconcelos, ``Anomaly detection in
  crowded scenes,'' in \emph{Proceedings of the IEEE Conference on Computer
  Vision and Pattern Recognition (CVPR)}, 2010, pp. 1975--1981.

\bibitem{ICRA17}
F.~Rezazadegan, S.~Shirazi, B.~Upcrofit, and M.~Milford, ``Action recognition:
  From static datasets to moving robots,'' in \emph{Proceedings of IEEE
  International Conference on Robotics and Automation (ICRA)}, 2017, pp.
  3185--3191.

\bibitem{zaheer2022generative}
M.~Zaheer, A.~Mahmood, M.~Khan, M.~Segu, F.~Yu, and S.~Lee, ``Generative
  cooperative learning for unsupervised video anomaly detection,'' in
  \emph{Proceedings of the IEEE Conference on Computer Vision and Pattern
  Recognition (CVPR)}, 2022, pp. 14\,744--14\,754.

\bibitem{UCF}
W.~Sultani, C.~Chen, and M.~Shah, ``Real-world anomaly detection in
  surveillance videos,'' in \emph{Proceedings of the IEEE Conference on
  Computer Vision and Pattern Recognition (CVPR)}, 2018, pp. 6479--6488.

\bibitem{7780455}
M.~Hasan, J.~Choi, J.~Neumann, A.~K. Roy-Chowdhury, and L.~S. Davis, ``Learning
  temporal regularity in video sequences,'' in \emph{Proceedings of the IEEE
  Conference on Computer Vision and Pattern Recognition (CVPR)}, 2016, pp.
  733--742.

\bibitem{liu2018ffp}
W.~Liu, W.~Luo, D.~Lian, and S.~Gao, ``Future frame prediction for anomaly
  detection--a new baseline,'' in \emph{Proceedings of the IEEE Conference on
  Computer Vision and Pattern Recognition (CVPR)}, 2018, pp. 6536--6545.

\bibitem{ShanghaiTech}
W.~Luo, W.~Liu, and S.~Gao, ``A revisit of sparse coding based anomaly
  detection in stacked rnn framework,'' in \emph{Proceedings of the IEEE
  International Conference on Computer Vision (ICCV)}, 2017, pp. 341--349.

\bibitem{Ravanbakhsh2017icip}
M.~Ravanbakhsh, M.~Nabi, E.~Sangineto, L.~Marcenaro, C.~Regazzoni, and N.~Sebe,
  ``Abnormal event detection in videos using generative adversarial nets,'' in
  \emph{Proceedings of IEEE International Conference on Image Processing
  (ICIP)}, 2017, pp. 1577--1581.

\bibitem{Sabokrou2017tip}
M.~Sabokrou, M.~Fayyaz, M.~Fathy, and R.~Klette, ``{Deep-Cascade}: Cascading
  {3D} deep neural networks for fast anomaly detection and localization in
  crowded scenes,'' \emph{IEEE Transactions on Image Processing}, vol.~26,
  no.~4, pp. 1992--2004, 2017.

\bibitem{Xia2015iccv}
Y.~Xia, X.~Cao, F.~Wen, G.~Hua, and J.~Sun, ``Learning discriminative
  reconstructions for unsupervised outlier removal,'' in \emph{Proceedings of
  IEEE International Conference on Computer Vision (ICCV)}, 2015, pp.
  1511--1519.

\bibitem{Zaigham2020cvpr}
M.~Zaigham~Zaheer, J.-H. Lee, M.~Astrid, and S.-I. Lee, ``Old is gold:
  Redefining the adversarially learned one-class classifier training
  paradigm,'' in \emph{Proceedings of IEEE Conference on Computer Vision and
  Pattern Recognition (CVPR)}, 2020, pp. 14\,171--14\,181.

\bibitem{zhang2016video}
Y.~Zhang, H.~Lu, L.~Zhang, X.~Ruan, and S.~Sakai, ``Video anomaly detection
  based on locality sensitive hashing filters,'' \emph{Pattern Recognition},
  vol.~59, pp. 302--311, 2016.

\bibitem{liu2019completeness}
D.~Liu, T.~Jiang, and Y.~Wang, ``Completeness modeling and context separation
  for weakly supervised temporal action localization,'' in \emph{Proceedings of
  the IEEE Conference on Computer Vision and Pattern Recognition (CVPR)}, 2019,
  pp. 1298--1307.

\bibitem{liu2019weakly}
Z.~Liu, L.~Wang, Q.~Zhang, W.~Tang, N.~Zheng, and G.~Hua, ``Weakly supervised
  temporal action localization through contrast based evaluation networks,''
  \emph{IEEE Transactions on Pattern Analysis and Machine Intelligence},
  vol.~44, no.~9, pp. 5886--5902, 2022.

\bibitem{narayan20193c}
S.~Narayan, H.~Cholakkal, F.~Khan, and L.~Shao, ``3c-net: Category count and
  center loss for weakly-supervised action localization,'' in \emph{Proceedings
  of the IEEE International Conference on Computer Vision (ICCV)}, 2019, pp.
  8679--8687.

\bibitem{shou2018autoloc}
Z.~Shou, H.~Gao, L.~Zhang, K.~Miyazawa, and S.~Chang, ``Autoloc:
  Weakly-supervised temporal action localization in untrimmed videos,'' in
  \emph{Proceedings of the European Conference on Computer Vision (ECCV)},
  2018, pp. 154--171.

\bibitem{wang2017untrimmednets}
L.~Wang, Y.~Xiong, D.~Lin, and L.~Van~Gool, ``Untrimmednets for weakly
  supervised action recognition and detection,'' in \emph{Proceedings of the
  IEEE Conference on Computer Vision and Pattern Recognition (CVPR)}, 2017, pp.
  4325--4334.

\bibitem{yu2019temporal}
T.~Yu, Z.~Ren, Y.~Li, E.~Yan, N.~Xu, and J.~Yuan, ``Temporal structure mining
  for weakly supervised action detection,'' in \emph{Proceedings of the IEEE
  International Conference on Computer Vision (ICCV)}, 2019, pp. 5522--5531.

\bibitem{tian2021weakly}
Y.~Tian, G.~Pang, Y.~Chen, R.~Singh, J.~Verjans, and G.~Carneiro,
  ``Weakly-supervised video anomaly detection with robust temporal feature
  magnitude learning,'' in \emph{Proceedings of the IEEE International
  Conference on Computer Vision (ICCV)}, 2021, pp. 4975--4986.

\bibitem{purwanto2021dance}
D.~Purwanto, Y.~Chen, and W.~Fang, ``Dance with self-attention: A new look of
  conditional random fields on anomaly detection in videos,'' in
  \emph{Proceedings of the IEEE International Conference on Computer Vision
  (ICCV)}, 2021, pp. 173--183.

\bibitem{zaheer2020claws}
M.~Zaheer, A.~Mahmood, M.~Astrid, and S.~Lee, ``Claws: Clustering assisted
  weakly supervised learning with normalcy suppression for anomalous event
  detection,'' in \emph{Proceedings of European Conference on Computer Vision
  (ECCV)}, 2020, pp. 358--376.

\bibitem{zaheer2020self}
M.~Zaheer, A.~Mahmood, H.~Shin, and S.~Lee, ``A self-reasoning framework for
  anomaly detection using video-level labels,'' \emph{IEEE Signal Processing
  Letters}, vol.~27, pp. 1705--1709, 2020.

\bibitem{ubnormal}
A.~Acsintoae, A.~Florescu, M.-I. Georgescu, T.~Mare, P.~Sumedrea, R.~T.
  Ionescu, F.~S. Khan, and M.~Shah, ``Ubnormal: New benchmark for supervised
  open-set video anomaly detection,'' in \emph{Proceedings of the IEEE
  Conference on Computer Vision and Pattern Recognition (CVPR)}, 2022, pp.
  20\,143--20\,153.

\bibitem{CUHK}
C.~Lu, J.~Shi, and J.~Jia, ``Abnormal event detection at 150 fps in matlab,''
  in \emph{Proceedings of the IEEE International Conference on Computer Vision
  (ICCV)}, 2013, pp. 2720--2727.

\bibitem{survey2021multimedia}
P.~Kumari, A.~K. Bedi, and M.~Saini, ``Multimedia datasets for anomaly
  detection: A survey,'' \emph{arXiv preprint arXiv:2112.05410}, 2021.

\bibitem{raghavendra2006unusual}
R.~Raghavendra, A.~Bue, and M.~Cristani, ``Unusual crowd activity dataset of
  university of minnesota,'' 2006.

\bibitem{Subway}
A.~Adam, E.~Rivlin, I.~Shimshoni, and D.~Reinitz, ``Robust real-time unusual
  event detection using multiple fixed-location monitors,'' \emph{IEEE
  Transactions on Pattern Analysis and Machine Intelligence}, vol.~30, no.~3,
  pp. 555--560, 2008.

\bibitem{ramachandra2020street}
B.~Ramachandra and M.~Jones, ``Street scene: A new dataset and evaluation
  protocol for video anomaly detection,'' in \emph{Proceedings of the IEEE
  Winter Conference on Applications of Computer Vision (WACV)}, 2020, pp.
  2569--2578.

\bibitem{Poiesi2015}
F.~Poiesi and A.~Cavallaro, ``Predicting and recognizing human interactions in
  public spaces,'' \emph{Journal of Real-Time Image Processing}, vol.~10,
  no.~4, pp. 785--803, 2015.

\bibitem{minidrone}
M.~Bonetto, P.~Korshunov, G.~Ramponi, and T.~Ebrahimi, ``Privacy in mini-drone
  based video surveillance,'' in \emph{Proceedings of IEEE International
  Conference on Image Processing (ICIP)}, 2015, pp. 2464--2469.

\bibitem{li2021uav}
T.~Li, J.~Liu, W.~Zhang, Y.~Ni, W.~Wang, and Z.~Li, ``Uav-human: A large
  benchmark for human behavior understanding with unmanned aerial vehicles,''
  in \emph{Proceedings of the IEEE Conference on Computer Vision and Pattern
  Recognition (CVPR)}, 2021, pp. 16\,266--16\,275.

\bibitem{seagull}
R.~Ribeiro, G.~Cruz, J.~Matos, and A.~Bernardino, ``A data set for airborne
  maritime surveillance environments,'' \emph{IEEE Transactions on Circuits and
  Systems for Video Technology}, vol.~29, no.~9, pp. 2720--2732, 2019.

\bibitem{seadronesee}
L.~A. Varga, B.~Kiefer, M.~Messmer, and A.~Zell, ``Seadronessee: A maritime
  benchmark for detecting humans in open water,'' in \emph{Proceedings of IEEE
  Winter Conference on Applications of Computer Vision (WACV)}, 2022, pp.
  3686--3696.

\bibitem{chakravarty2007anomaly}
P.~Chakravarty, A.~M. Zhang, R.~Jarvis, and L.~Kleeman, ``Anomaly detection and
  tracking for a patrolling robot,'' in \emph{Proceedings of Australasian
  Conference on Robotics and Automation (ACRA)}, 2007.

\bibitem{gehring2021anymal}
C.~Gehring, P.~Fankhauser, L.~Isler, R.~Diethelm, S.~Bachmann, M.~Potz,
  L.~Gerstenberg, and M.~Hutter, ``Anymal in the field: Solving industrial
  inspection of an offshore hvdc platform with a quadrupedal robot,'' in
  \emph{Field and Service Robotics}, 2021, pp. 247--260.

\bibitem{pinggera2016lost}
P.~Pinggera, S.~Ramos, S.~Gehrig, U.~Franke, C.~Rother, and R.~Mester, ``Lost
  and found: detecting small road hazards for self-driving vehicles,'' in
  \emph{Proceedings of IEEE/RSJ International Conference on Intelligent Robots
  and Systems (IROS)}, 2016, pp. 1099--1106.

\bibitem{kragh2017fieldsafe}
M.~F. Kragh, P.~Christiansen, M.~S. Laursen, M.~Larsen, K.~A. Steen, O.~Green,
  H.~Karstoft, and R.~N. J{\o}rgensen, ``Fieldsafe: dataset for obstacle
  detection in agriculture,'' \emph{Sensors}, vol.~17, no.~11, p. 2579, 2017.

\bibitem{singh2020lidar}
A.~Singh, A.~Kamireddypalli, V.~Gandhi, and K.~M. Krishna, ``Lidar guided small
  obstacle segmentation,'' in \emph{Proceedings of IEEE/RSJ International
  Conference on Intelligent Robots and Systems (IROS)}, 2020, pp. 8513--8520.

\bibitem{roadanomal}
K.~Lis, K.~K. Nakka, P.~Fua, and M.~Salzmann, ``Detecting the unexpected via
  image resynthesis,'' in \emph{Proceedings of IEEE International Conference on
  Computer Vision (ICCV)}, 2019, pp. 2152--2161.

\bibitem{SMD}
D.~K. Prasad, D.~Rajan, L.~Rachmawati, E.~Rajabally, and C.~Quek, ``Video
  processing from electro-optical sensors for object detection and tracking in
  a maritime environment: A survey,'' \emph{IEEE Transactions on Intelligent
  Transportation Systems}, vol.~18, no.~8, pp. 1993--2016, 2017.

\bibitem{MODS2022}
B.~Bovcon, J.~Muhovič, D.~Vranac, D.~Mozetič, J.~Perš, and M.~Kristan,
  ``Mods—a usv-oriented object detection and obstacle segmentation
  benchmark,'' \emph{IEEE Transactions on Intelligent Transportation Systems},
  vol.~23, no.~8, pp. 13\,403--13\,418, 2022.

\bibitem{UUD}
C.~Liu, Z.~Wang, S.~Wang, T.~Tang, Y.~Tao, C.~Yang, H.~Li, X.~Liu, and X.~Fan,
  ``A new dataset, poisson gan and aquanet for underwater object grabbing,''
  \emph{IEEE Transactions on Circuits and Systems for Video Technology},
  vol.~32, no.~5, pp. 2831--2844, 2022.

\bibitem{zhou2022discovering}
Y.~Zhou, B.~Li, J.~Wang, E.~Rocco, and Q.~Meng, ``Discovering unknowns:
  Context-enhanced anomaly detection for curiosity-driven autonomous underwater
  exploration,'' \emph{Pattern Recognition}, vol. 131, p. 108860, 2022.

\bibitem{HEVI}
Y.~Yao, M.~Xu, C.~Choi, D.~J. Crandall, E.~M. Atkins, and B.~Dariush,
  ``Egocentric vision-based future vehicle localization for intelligent driving
  assistance systems,'' in \emph{Proceedings of International Conference on
  Robotics and Automation (ICRA)}, 2019, pp. 9711--9717.

\bibitem{A3D}
Y.~Yao, M.~Xu, Y.~Wang, D.~J. Crandall, and E.~M. Atkins, ``Unsupervised
  traffic accident detection in first-person videos,'' in \emph{Proceedings of
  IEEE/RSJ International Conference on Intelligent Robots and Systems (IROS)},
  2019, pp. 273--280.

\bibitem{dada2000}
J.~Fang, D.~Yan, J.~Qiao, J.~Xue, H.~Wang, and S.~Li, ``Dada-2000: Can driving
  accident be predicted by driver attentionƒ analyzed by a benchmark,'' in
  \emph{Proceedings of IEEE Intelligent Transportation Systems Conference
  (ITSC)}, 2019, pp. 4303--4309.

\bibitem{CCD}
W.~Bao, Q.~Yu, and Y.~Kong, ``Uncertainty-based traffic accident anticipation
  with spatio-temporal relational learning,'' in \emph{Proceedings of ACM
  International Conference on Multimedia}, 2020, pp. 2682--2690.

\bibitem{DoTA}
Y.~Yao, X.~Wang, M.~Xu, Z.~Pu, Y.~Wang, E.~Atkins, and D.~Crandall, ``Dota:
  Unsupervised detection of traffic anomaly in driving videos,'' \emph{IEEE
  Transactions on Pattern Analysis and Machine Intelligence}, pp. 1--1, 2022.

\bibitem{Eyeinsky2018}
A.~Singh, D.~Patil, and S.~Omkar, ``Eye in the sky: Real-time drone
  surveillance system (dss) for violent individuals identification using
  scatternet hybrid deep learning network,'' in \emph{Proceedings of IEEE
  Conference on Computer Vision and Pattern Recognition Workshops (CVPRW)},
  2018, pp. 1710--17\,108.

\bibitem{ipatch}
L.~Patino, T.~Nawaz, T.~Cane, and J.~Ferryman, ``Pets 2017: Dataset and
  challenge,'' in \emph{Proceedings of IEEE Conference on Computer Vision and
  Pattern Recognition Workshops (CVPRW)}, 2017, pp. 2126--2132.

\bibitem{HTA}
H.~Singh, E.~M. Hand, and K.~Alexis, ``Anomalous motion detection on highway
  using deep learning,'' in \emph{Proceedings of IEEE International Conference
  on Image Processing (ICIP)}, 2020, pp. 1901--1905.

\bibitem{TIP2010}
H.~Kong, J.-Y. Audibert, and J.~Ponce, ``Detecting abandoned objects with a
  moving camera,'' \emph{IEEE Transactions on Image Processing}, vol.~19,
  no.~8, pp. 2201--2210, 2010.

\bibitem{poiesi2016detection}
F.~Poiesi and A.~Cavallaro, ``Detection of fast incoming objects with a moving
  camera.'' in \emph{Proceedings of the British Machine Vision Conference
  (BMVC)}, 2016.

\bibitem{kristan2015fast}
M.~Kristan, V.~Kenk, S.~Kova{\v{c}}i{\v{c}}, and J.~Per{\v{s}}, ``Fast
  image-based obstacle detection from unmanned surface vehicles,'' \emph{IEEE
  Transactions on Cybernetics}, vol.~46, no.~3, pp. 641--654, 2015.

\bibitem{modd2}
B.~Bovcon, J.~Per{\v{s}}, M.~Kristan \emph{et~al.}, ``Stereo obstacle detection
  for unmanned surface vehicles by imu-assisted semantic segmentation,''
  \emph{Robotics and Autonomous Systems}, vol. 104, pp. 1--13, 2018.

\bibitem{mantegazza2022challenges}
D.~Mantegazza, A.~Giusti, L.~Gambardella, A.~Rizzoli, and J.~Guzzi,
  ``Challenges in visual anomaly detection for mobile robots,'' \emph{arXiv
  preprint arXiv:2209.10995}, 2022.

\bibitem{deepvad2021}
J.~Ren, F.~Xia, Y.~Liu, and I.~Lee, ``Deep video anomaly detection:
  Opportunities and challenges,'' in \emph{Proceedings of International
  Conference on Data Mining Workshops (ICDMW)}, 2021, pp. 959--966.

\bibitem{surveyAD}
G.~Pang, C.~Shen, L.~Cao, and A.~V.~D. Hengel, ``Deep learning for anomaly
  detection: A review,'' \emph{ACM Computing Surveys (CSUR)}, vol.~54, no.~2,
  pp. 1--38, 2021.

\bibitem{AIreview22}
S.~Chandrakala, K.~Deepak, and G.~Revathy, ``Anomaly detection in surveillance
  videos: a thematic taxonomy of deep models, review and performance
  analysis,'' \emph{Artificial Intelligence Review}, pp. 1--50, 2022.

\bibitem{popoola2012video}
O.~P. Popoola and K.~Wang, ``Video-based abnormal human behavior
  recognition—a review,'' \emph{IEEE Transactions on Systems, Man, and
  Cybernetics, Part C (Applications and Reviews)}, vol.~42, no.~6, pp.
  865--878, 2012.

\bibitem{Spagnolo2014}
P.~Spagnolo, P.~Mazzeo, and C.~Distante, \emph{Human Behavior Understanding in
  Networked Sensing}.\hskip 1em plus 0.5em minus 0.4em\relax Springer, 2014.

\bibitem{surveyHAR}
P.~Pareek and A.~Thakkar, ``A survey on video-based human action recognition:
  recent updates, datasets, challenges, and applications,'' \emph{Artificial
  Intelligence Review}, vol.~54, no.~3, pp. 2259--2322, 2021.

\bibitem{10038646}
D.~Xiao, M.~Dianati, W.~G. Geiger, and R.~Woodman, ``Review of graph-based
  hazardous event detection methods for autonomous driving systems,''
  \emph{IEEE Transactions on Intelligent Transportation Systems}, pp. 1--19,
  2023.

\bibitem{liu2023generalized}
Y.~Liu, D.~Yang, Y.~Wang, J.~Liu, and L.~Song, ``Generalized video anomaly
  event detection: Systematic taxonomy and comparison of deep models,''
  \emph{arXiv preprint arXiv:2302.05087}, 2023.

\bibitem{ullah2023comprehensive}
F.~U.~M. Ullah, M.~S. Obaidat, A.~Ullah, K.~Muhammad, M.~Hijji, and S.~W. Baik,
  ``A comprehensive review on vision-based violence detection in surveillance
  videos,'' \emph{ACM Computing Surveys}, vol.~55, no.~10, pp. 1--44, 2023.

\bibitem{mumtaz2023overview}
N.~Mumtaz, N.~Ejaz, S.~Habib, S.~M. Mohsin, P.~Tiwari, S.~S. Band, and
  N.~Kumar, ``An overview of violence detection techniques: current challenges
  and future directions,'' \emph{Artificial Intelligence Review}, vol.~56,
  no.~5, pp. 4641--4666, 2023.

\bibitem{gowsikhaa2014automated}
D.~Gowsikhaa, S.~Abirami, and R.~Baskaran, ``Automated human behavior analysis
  from surveillance videos: a survey,'' \emph{Artificial Intelligence Review},
  vol.~42, pp. 747--765, 2014.

\bibitem{surveyoldpeople}
A.~Lentzas and D.~Vrakas, ``Non-intrusive human activity recognition and
  abnormal behavior detection on elderly people: A review,'' \emph{Artificial
  Intelligence Review}, vol.~53, no.~3, pp. 1975--2021, 2020.

\bibitem{santhosh2020anomaly}
K.~K. Santhosh, D.~P. Dogra, and P.~P. Roy, ``Anomaly detection in road traffic
  using visual surveillance: A survey,'' \emph{ACM Computing Surveys (CSUR)},
  vol.~53, no.~6, pp. 1--26, 2020.

\bibitem{breitenstein2021corner}
J.~Breitenstein, J.-A. Term{\"o}hlen, D.~Lipinski, and T.~Fingscheidt, ``Corner
  cases for visual perception in automated driving: some guidance on detection
  approaches,'' \emph{arXiv preprint arXiv:2102.05897}, 2021.

\bibitem{nguyen2022state}
K.~Nguyen, C.~Fookes, S.~Sridharan, Y.~Tian, F.~Liu, X.~Liu, and A.~Ross, ``The
  state of aerial surveillance: A survey,'' \emph{arXiv preprint
  arXiv:2201.03080}, 2022.

\bibitem{khan2019survey}
S.~D. Khan and H.~Ullah, ``A survey of advances in vision-based vehicle
  re-identification,'' \emph{Computer Vision and Image Understanding}, vol.
  182, pp. 50--63, 2019.

\bibitem{VDAO}
A.~F. {Da Silva}, L.~A. Thomaz, G.~Carvalho, M.~T. Nakahata, E.~Jardim,
  J.~F.~L. de~Oliveira, E.~A.~B. da~Silva, S.~L. Netto, G.~Freitas, and R.~R.
  Costa, ``An annotated video database for abandoned-object detection in a
  cluttered environment,'' in \emph{Proceedings of International
  Telecommunications Symposium (ITS)}, 2014, pp. 1--5.

\bibitem{mur2017orb}
R.~Mur-Artal and J.~D. Tard{\'o}s, ``Orb-slam2: An open-source slam system for
  monocular, stereo, and rgb-d cameras,'' \emph{IEEE Transactions on Robotics},
  vol.~33, no.~5, pp. 1255--1262, 2017.

\bibitem{D3vo}
N.~Yang, L.~von Stumberg, R.~Wang, and D.~Cremers, ``D3vo: Deep depth, deep
  pose and deep uncertainty for monocular visual odometry,'' in
  \emph{Proceedings of the IEEE Conference on Computer Vision and Pattern
  Recognition (CVPR)}, 2020, pp. 1278--1289.

\bibitem{9741458}
Y.~Gao, F.~Tian, J.~Li, Z.~Fang, S.~Al-Rubaye, W.~Song, and Y.~Yan, ``Joint
  optimization of depth and ego-motion for intelligent autonomous vehicles,''
  \emph{IEEE Transactions on Intelligent Transportation Systems}, vol.~24,
  no.~7, pp. 7390--7403, 2023.

\bibitem{backgroundbias}
M.~Tian, S.~Yi, H.~Li, S.~Li, X.~Zhang, J.~Shi, J.~Yan, and X.~Wang,
  ``Eliminating background-bias for robust person re-identification,'' in
  \emph{Proceedings of the IEEE Conference on Computer Vision and Pattern
  Recognition (CVPR)}, 2018, pp. 5794--5803.

\bibitem{zhou2019attention}
J.~T. Zhou, L.~Zhang, Z.~Fang, J.~Du, X.~Peng, and Y.~Xiao, ``Attention-driven
  loss for anomaly detection in video surveillance,'' \emph{IEEE Transactions
  on Circuits and Systems for Video Technology}, vol.~30, no.~12, pp.
  4639--4647, 2019.

\bibitem{chapel2020moving}
M.-N. Chapel and T.~Bouwmans, ``Moving objects detection with a moving camera:
  A comprehensive review,'' \emph{Computer Science Review}, vol.~38, p. 100310,
  2020.

\bibitem{erez2022deep}
G.~Erez, R.~S. Weber, and O.~Freifeld, ``A deep moving-camera background
  model,'' in \emph{European Conference on Computer Vision}.\hskip 1em plus
  0.5em minus 0.4em\relax Springer, 2022, pp. 177--194.

\bibitem{RE-DID}
P.~Rota, N.~Conci, N.~Sebe, and J.~M. Rehg, ``Real-life violent social
  interaction detection,'' in \emph{Proceedings of IEEE International
  Conference on Image Processing (ICIP)}, 2015, pp. 3456--3460.

\bibitem{GMRPD}
H.~Wang, R.~Fan, Y.~Sun, and M.~Liu, ``Dynamic fusion module evolves drivable
  area and road anomaly detection: A benchmark and algorithms,'' \emph{IEEE
  Transactions on Cybernetics}, vol.~52, no.~10, pp. 10\,750--10\,760, 2022.

\bibitem{XMAS}
D.~Noh, C.~Sung, T.~Uhm, W.~Lee, H.~Lim, J.~Choi, K.~Lee, D.~Hong, D.~Um,
  I.~Chung, H.~Shin, M.~Kim, H.-R. Kim, S.~Baek, and H.~Myung, ``X-mas:
  Extremely large-scale multi-modal sensor dataset for outdoor surveillance in
  real environments,'' \emph{IEEE Robotics and Automation Letters}, vol.~8,
  no.~2, pp. 1093--1100, 2023.

\bibitem{filonenko2016unattended}
A.~Filonenko, K.-H. Jo \emph{et~al.}, ``Unattended object identification for
  intelligent surveillance systems using sequence of dual background
  difference,'' \emph{IEEE Transactions on Industrial Informatics}, vol.~12,
  no.~6, pp. 2247--2255, 2016.

\bibitem{luna2018abandoned}
E.~Luna, J.~C. San~Miguel, D.~Ortego, and J.~M. Mart{\'\i}nez, ``Abandoned
  object detection in video-surveillance: survey and comparison,''
  \emph{Sensors}, vol.~18, no.~12, p. 4290, 2018.

\bibitem{jo2017cumulative}
K.-H. Jo \emph{et~al.}, ``Cumulative dual foreground differences for illegally
  parked vehicles detection,'' \emph{IEEE Transactions on Industrial
  Informatics}, vol.~13, no.~5, pp. 2464--2473, 2017.

\bibitem{Zaham}
M.~Z. Zaheer, A.~Mahmood, M.~H. Khan, M.~Astrid, and S.-I. Lee, ``An anomaly
  detection system via moving surveillance robots with human collaboration,''
  in \emph{Proceedings of IEEE International Conference on Computer Vision
  Workshops (ICCVW)}, 2021, pp. 2595--2601.

\bibitem{VDAO18TCS}
L.~A. Thomaz, E.~Jardim, A.~F. da~Silva, E.~A.~B. da~Silva, S.~L. Netto, and
  H.~Krim, ``Anomaly detection in moving-camera video sequences using principal
  subspace analysis,'' \emph{IEEE Transactions on Circuits and Systems I:
  Regular Papers}, vol.~65, no.~3, pp. 1003--1015, 2018.

\bibitem{VDAO18MSSP}
M.~T. Nakahata, L.~A. Thomaz, A.~F. da~Silva, E.~A. da~Silva, and S.~L. Netto,
  ``Anomaly detection with a moving camera using spatio-temporal codebooks,''
  \emph{Multidimensional Systems and Signal Processing}, vol.~29, no.~3, pp.
  1025--1054, 2018.

\bibitem{VDAO20TIP}
E.~Jardim, L.~A. Thomaz, E.~A.~B. da~Silva, and S.~L. Netto,
  ``Domain-transformable sparse representation for anomaly detection in
  moving-camera videos,'' \emph{IEEE Transactions on Image Processing},
  vol.~29, pp. 1329--1343, 2020.

\bibitem{patrolrobot16}
W.~Lawson, L.~Hiatt, and K.~Sullivan, ``Detecting anomalous objects on mobile
  platforms,'' in \emph{Proceedings of IEEE Conference on Computer Vision and
  Pattern Recognition Workshops (CVPRW)}, 2016, pp. 1426--1433.

\bibitem{patrolrobot17}
W.~Lawson, E.~Bekele, and K.~Sullivan, ``Finding anomalies with generative
  adversarial networks for a patrolbot,'' in \emph{Proceedings of IEEE
  Conference on Computer Vision and Pattern Recognition Workshops (CVPRW)},
  2017, pp. 484--485.

\bibitem{PAMI2001}
G.~Medioni, I.~Cohen, F.~Bremond, S.~Hongeng, and R.~Nevatia, ``Event detection
  and analysis from video streams,'' \emph{IEEE Transactions on Pattern
  Analysis and Machine Intelligence}, vol.~23, no.~8, pp. 873--889, 2001.

\bibitem{MDV2018}
J.~Henrio and T.~Nakashima, ``Anomaly detection in videos recorded by drones in
  a surveillance context,'' in \emph{2018 IEEE International Conference on
  Systems, Man, and Cybernetics (SMC)}, 2018, pp. 2503--2508.

\bibitem{MDV2020}
A.~Chriki, H.~Touati, H.~Snoussi, and F.~Kamoun, ``Uav-based surveillance
  system: an anomaly detection approach,'' in \emph{Proceedings of IEEE
  Symposium on Computers and Communications (ISCC)}, 2020, pp. 1--6.

\bibitem{MDV2021}
------, ``Deep learning and handcrafted features for one-class anomaly
  detection in uav video,'' \emph{Multimedia Tools and Applications}, vol.~80,
  no.~2, pp. 2599--2620, 2021.

\bibitem{srivastava2021recognizing}
A.~Srivastava, T.~Badal, A.~Garg, A.~Vidyarthi, and R.~Singh, ``Recognizing
  human violent action using drone surveillance within real-time proximity,''
  \emph{Journal of Real-Time Image Processing}, vol.~18, no.~5, pp. 1851--1863,
  2021.

\bibitem{lin2017feature}
T.-Y. Lin, P.~Doll{\'a}r, R.~Girshick, K.~He, B.~Hariharan, and S.~Belongie,
  ``Feature pyramid networks for object detection,'' in \emph{Proceedings of
  the IEEE Conference on Computer Vision and Pattern Recognition (CVPR)}, 2017,
  pp. 2117--2125.

\bibitem{srivastava2022uav}
A.~Srivastava, T.~Badal, P.~Saxena, A.~Vidyarthi, and R.~Singh, ``Uav
  surveillance for violence detection and individual identification,''
  \emph{Automated Software Engineering}, vol.~29, no.~1, p.~28, 2022.

\bibitem{d2city}
Z.~Che, G.~Li, T.~Li, B.~Jiang, X.~Shi, X.~Zhang, Y.~Lu, G.~Wu, Y.~Liu, and
  J.~Ye, ``D$^2$-city: A large-scale dashcam video dataset of diverse traffic
  scenarios,'' \emph{arXiv preprint arXiv:1904.01975}, 2019.

\bibitem{truckVAD}
S.~Haresh, S.~Kumar, M.~Z. Zia, and Q.-H. Tran, ``Towards anomaly detection in
  dashcam videos,'' in \emph{Proceedings of IEEE Intelligent Vehicles Symposium
  (IV)}, 2020, pp. 1407--1414.

\bibitem{national2021traffic}
{National Center for Statistics and Analysis}, ``Traffic safety facts 2019: A
  compilation of motor vehicle crash data,'' \emph{National Highway Traffic
  Safety Administration}, 2021.

\bibitem{obstacle2009}
A.~Ess, B.~Leibe, K.~Schindler, and L.~van Gool, ``Moving obstacle detection in
  highly dynamic scenes,'' in \emph{Proceedings of IEEE International
  Conference on Robotics and Automation (ICRA)}, 2009, pp. 56--63.

\bibitem{8786197}
H.~Wang, Y.~Sun, and M.~Liu, ``Self-supervised drivable area and road anomaly
  segmentation using rgb-d data for robotic wheelchairs,'' \emph{IEEE Robotics
  and Automation Letters}, vol.~4, no.~4, pp. 4386--4393, 2019.

\bibitem{kitti2012}
A.~Geiger, P.~Lenz, and R.~Urtasun, ``Are we ready for autonomous driving? the
  kitti vision benchmark suite,'' in \emph{Proceedings of the IEEE Conference
  on Computer Vision and Pattern Recognition (CVPR)}, 2012, pp. 3354--3361.

\bibitem{cordts2016cityscapes}
M.~Cordts, M.~Omran, S.~Ramos, T.~Rehfeld, M.~Enzweiler, R.~Benenson,
  U.~Franke, S.~Roth, and B.~Schiele, ``The cityscapes dataset for semantic
  urban scene understanding,'' in \emph{Proceedings of the IEEE Conference on
  Computer Vision and Pattern Recognition (CVPR)}, 2016, pp. 3213--3223.

\bibitem{sirimanne2019review}
S.~N. Sirimanne, J.~Hoffman, W.~Juan, R.~Asariotis, M.~Assaf, G.~Ayala,
  H.~Benamara, D.~Chantrel, J.~Hoffmann, A.~Premti \emph{et~al.}, ``Review of
  maritime transport 2019,'' in \emph{United Nations conference on trade and
  development, Geneva, Switzerland}, 2019.

\bibitem{8911242}
D.~K. Prasad, H.~Dong, D.~Rajan, and C.~Quek, ``Are object detection assessment
  criteria ready for maritime computer vision?'' \emph{IEEE Transactions on
  Intelligent Transportation Systems}, vol.~21, no.~12, pp. 5295--5304, 2020.

\bibitem{kiefer20231st}
B.~Kiefer, M.~Kristan, J.~Per{\v{s}}, L.~{\v{Z}}ust, F.~Poiesi, F.~Andrade,
  A.~Bernardino, M.~Dawkins, J.~Raitoharju, Y.~Quan \emph{et~al.}, ``1st
  workshop on maritime computer vision (macvi) 2023: Challenge results,'' in
  \emph{Proceedings of the IEEE Winter Conference on Applications of Computer
  Vision}, 2023, pp. 265--302.

\bibitem{hong2020trashcan}
J.~Hong, M.~Fulton, and J.~Sattar, ``Trashcan: A semantically-segmented dataset
  towards visual detection of marine debris,'' \emph{arXiv preprint
  arXiv:2007.08097}, 2020.

\bibitem{SUIM2020}
M.~J. Islam, C.~Edge, Y.~Xiao, P.~Luo, M.~Mehtaz, C.~Morse, S.~S. Enan, and
  J.~Sattar, ``Semantic segmentation of underwater imagery: Dataset and
  benchmark,'' in \emph{Proceedings of International Conference on Intelligent
  Robots and Systems (IROS)}, 2020, pp. 1769--1776.

\bibitem{mantegazza2022outlier}
D.~Mantegazza, A.~Giusti, L.~M. Gambardella, and J.~Guzzi, ``An outlier
  exposure approach to improve visual anomaly detection performance for mobile
  robots,'' \emph{IEEE Robotics and Automation Letters}, vol.~7, no.~4, pp.
  11\,354--11\,361, 2022.

\bibitem{8917818}
C.~Li, C.~Guo, W.~Ren, R.~Cong, J.~Hou, S.~Kwong, and D.~Tao, ``An underwater
  image enhancement benchmark dataset and beyond,'' \emph{IEEE Transactions on
  Image Processing}, vol.~29, pp. 4376--4389, 2020.

\bibitem{berman2020underwater}
D.~Berman, D.~Levy, S.~Avidan, and T.~Treibitz, ``Underwater single image color
  restoration using haze-lines and a new quantitative dataset,'' \emph{IEEE
  Transactions on Pattern Analysis and Machine Intelligence}, 2020.

\bibitem{7301727}
D.~D. Bloisi, L.~Iocchi, A.~Pennisi, and L.~Tombolini, ``Argos-venice boat
  classification,'' in \emph{Proceedings of IEEE International Conference on
  Advanced Video and Signal Based Surveillance (AVSS)}, 2015, pp. 1--6.

\bibitem{9381638}
Y.~Cheng, M.~Jiang, J.~Zhu, and Y.~Liu, ``Are we ready for unmanned surface
  vehicles in inland waterways? the usvinland multisensor dataset and
  benchmark,'' \emph{IEEE Robotics and Automation Letters}, vol.~6, no.~2, pp.
  3964--3970, 2021.

\bibitem{Cheng_2021_ICCV}
Y.~Cheng, J.~Zhu, M.~Jiang, J.~Fu, C.~Pang, P.~Wang, K.~Sankaran, O.~Onabola,
  Y.~Liu, D.~Liu, and Y.~Bengio, ``Flow: A dataset and benchmark for floating
  waste detection in inland waters,'' in \emph{Proceedings of the IEEE
  International Conference on Computer Vision (ICCV)}, October 2021, pp.
  10\,953--10\,962.

\bibitem{liu2021efficient}
J.~Liu, H.~Li, J.~Luo, S.~Xie, and Y.~Sun, ``Efficient obstacle detection based
  on prior estimation network and spatially constrained mixture model for
  unmanned surface vehicles,'' \emph{Journal of Field Robotics}, vol.~38,
  no.~2, pp. 212--228, 2021.

\bibitem{7533104}
H.~Mukojima, D.~Deguchi, Y.~Kawanishi, I.~Ide, H.~Murase, M.~Ukai, N.~Nagamine,
  and R.~Nakasone, ``Moving camera background-subtraction for obstacle
  detection on railway tracks,'' in \emph{Proceedings of International
  Conference on Image Processing (ICIP)}, 2016, pp. 3967--3971.

\bibitem{railanomaly}
Y.~Wang, Z.~Yu, and L.~Zhu, ``Intrusion detection for high-speed railways based
  on unsupervised anomaly detection models,'' \emph{Applied Intelligence},
  vol.~53, no.~7, pp. 8453--8466, 2023.

\bibitem{he2021obstacle}
D.~He, Z.~Zou, Y.~Chen, B.~Liu, X.~Yao, and S.~Shan, ``Obstacle detection of
  rail transit based on deep learning,'' \emph{Measurement}, vol. 176, p.
  109241, 2021.

\bibitem{BDD100K}
F.~Yu, H.~Chen, X.~Wang, W.~Xian, Y.~Chen, F.~Liu, V.~Madhavan, and T.~Darrell,
  ``Bdd100k: A diverse driving dataset for heterogeneous multitask learning,''
  in \emph{Proceedings of the IEEE Conference on Computer Vision and Pattern
  Recognition (CVPR)}, June 2020.

\bibitem{GCNVAD}
R.~Herzig, E.~Levi, H.~Xu, H.~Gao, E.~Brosh, X.~Wang, A.~Globerson, and
  T.~Darrell, ``Spatio-temporal action graph networks,'' in \emph{2019 IEEE
  International Conference on Computer Vision Workshop (ICCVW)}, 2019, pp.
  2347--2356.

\bibitem{8809907}
R.~Fan, U.~Ozgunalp, B.~Hosking, M.~Liu, and I.~Pitas, ``Pothole detection
  based on disparity transformation and road surface modeling,'' \emph{IEEE
  Transactions on Image Processing}, vol.~29, pp. 897--908, 2020.

\bibitem{mcmahon2017multimodal}
S.~McMahon, N.~S{\"u}nderhauf, B.~Upcroft, and M.~Milford, ``Multimodal trip
  hazard affordance detection on construction sites,'' \emph{IEEE Robotics and
  Automation Letters}, vol.~3, no.~1, pp. 1--8, 2017.

\bibitem{8639077}
T.~Cane and J.~Ferryman, ``Evaluating deep semantic segmentation networks for
  object detection in maritime surveillance,'' in \emph{Proceedings of IEEE
  International Conference on Advanced Video and Signal Based Surveillance
  (AVSS)}, 2018, pp. 1--6.

\bibitem{8793588}
B.~Arain, C.~McCool, P.~Rigby, D.~Cagara, and M.~Dunbabin, ``Improving
  underwater obstacle detection using semantic image segmentation,'' in
  \emph{Proceedings of International Conference on Robotics and Automation
  (ICRA)}, 2019, pp. 9271--9277.

\bibitem{ohgushi2020road}
T.~Ohgushi, K.~Horiguchi, and M.~Yamanaka, ``Road obstacle detection method
  based on an autoencoder with semantic segmentation,'' in \emph{Proceedings of
  the Asian Conference on Computer Vision (ACCV)}, 2020.

\bibitem{xia2020synthesize}
Y.~Xia, Y.~Zhang, F.~Liu, W.~Shen, and A.~L. Yuille, ``Synthesize then compare:
  Detecting failures and anomalies for semantic segmentation,'' in
  \emph{Proceedings of European Conference on Computer Vision (ECCV)}, 2020,
  pp. 145--161.

\bibitem{vzust2022temporal}
L.~{\v{Z}}ust and M.~Kristan, ``Temporal context for robust maritime obstacle
  detection,'' in \emph{Proceedings of IEEE/RSJ International Conference on
  Intelligent Robots and Systems (IROS)}, 2022, pp. 6340--6346.

\bibitem{6393573}
J.~A. Uribe, L.~Fonseca, and J.~F. Vargas, ``Video based system for railroad
  collision warning,'' in \emph{Proceedings of IEEE International Carnahan
  Conference on Security Technology (ICCST)}, 2012, pp. 280--285.

\bibitem{8575414}
J.~Wei, J.~Zhao, Y.~Zhao, and Z.~Zhao, ``Unsupervised anomaly detection for
  traffic surveillance based on background modeling,'' in \emph{Proceedings of
  IEEE Conference on Computer Vision and Pattern Recognition Workshops
  (CVPRW)}, 2018, pp. 129--1297.

\bibitem{ionescu2019object}
R.~T. Ionescu, F.~S. Khan, M.-I. Georgescu, and L.~Shao, ``Object-centric
  auto-encoders and dummy anomalies for abnormal event detection in video,'' in
  \emph{Proceedings of the IEEE Conference on Computer Vision and Pattern
  Recognition (CVPR)}, 2019, pp. 7842--7851.

\bibitem{morais2019learning}
R.~Morais, V.~Le, T.~Tran, B.~Saha, M.~Mansour, and S.~Venkatesh, ``Learning
  regularity in skeleton trajectories for anomaly detection in videos,'' in
  \emph{Proceedings of the IEEE Conference on Computer Vision and Pattern
  Recognition (CVPR)}, 2019, pp. 11\,996--12\,004.

\bibitem{liu2019exploring}
K.~Liu and H.~Ma, ``Exploring background-bias for anomaly detection in
  surveillance videos,'' in \emph{Proceedings of ACM International Conference
  on Multimedia}, 2019, pp. 1490--1499.

\bibitem{coppola2020social}
C.~Coppola, S.~Cosar, D.~R. Faria, and N.~Bellotto, ``Social activity
  recognition on continuous rgb-d video sequences,'' \emph{International
  Journal of Social Robotics}, vol.~12, pp. 201--215, 2020.

\bibitem{9093633}
R.~Rodrigues, N.~Bhargava, R.~Velmurugan, and S.~Chaudhuri, ``Multi-timescale
  trajectory prediction for abnormal human activity detection,'' in
  \emph{Proceedings of IEEE Winter Conference on Applications of Computer
  Vision (WACV)}, 2020, pp. 2615--2623.

\bibitem{9157616}
A.~Markovitz, G.~Sharir, I.~Friedman, L.~Zelnik-Manor, and S.~Avidan, ``Graph
  embedded pose clustering for anomaly detection,'' in \emph{Proceedings of
  IEEE Conference on Computer Vision and Pattern Recognition (CVPR)}, 2020, pp.
  10\,536--10\,544.

\bibitem{georgescu2021anomaly}
M.-I. Georgescu, A.~Barbalau, R.~T. Ionescu, F.~S. Khan, M.~Popescu, and
  M.~Shah, ``Anomaly detection in video via self-supervised and multi-task
  learning,'' in \emph{Proceedings of the IEEE Conference on Computer Vision
  and Pattern Recognition (CVPR)}, 2021, pp. 12\,742--12\,752.

\bibitem{alahi2016social}
A.~Alahi, K.~Goel, V.~Ramanathan, A.~Robicquet, L.~Fei-Fei, and S.~Savarese,
  ``Social lstm: Human trajectory prediction in crowded spaces,'' in
  \emph{Proceedings of the IEEE Conference on Computer Vision and Pattern
  Recognition (CVPR)}, 2016, pp. 961--971.

\bibitem{8578539}
A.~Bhattacharyya, M.~Fritz, and B.~Schiele, ``Long-term on-board prediction of
  people in traffic scenes under uncertainty,'' in \emph{Proceedings of IEEE
  Conference on Computer Vision and Pattern Recognition (CVPR)}, 2018, pp.
  4194--4202.

\bibitem{yagi2018future}
T.~Yagi, K.~Mangalam, R.~Yonetani, and Y.~Sato, ``Future person localization in
  first-person videos,'' in \emph{Proceedings of the IEEE Conference on
  Computer Vision and Pattern Recognition (CVPR)}, 2018, pp. 7593--7602.

\bibitem{rasouli2019pie}
A.~Rasouli, I.~Kotseruba, T.~Kunic, and J.~K. Tsotsos, ``Pie: A large-scale
  dataset and models for pedestrian intention estimation and trajectory
  prediction,'' in \emph{Proceedings of the IEEE International Conference on
  Computer Vision (ICCV)}, 2019, pp. 6262--6271.

\bibitem{malla2020titan}
S.~Malla, B.~Dariush, and C.~Choi, ``Titan: Future forecast using action
  priors,'' in \emph{Proceedings of the IEEE Conference on Computer Vision and
  Pattern Recognition (CVPR)}, 2020, pp. 11\,186--11\,196.

\bibitem{qiu2022egocentric}
J.~Qiu, L.~Chen, X.~Gu, F.~P.-W. Lo, Y.-Y. Tsai, J.~Sun, J.~Liu, and B.~Lo,
  ``Egocentric human trajectory forecasting with a wearable camera and
  multi-modal fusion,'' \emph{IEEE Robotics and Automation Letters}, vol.~7,
  no.~4, pp. 8799--8806, 2022.

\bibitem{9714213}
Z.~Zhou, X.~Dong, Z.~Li, K.~Yu, C.~Ding, and Y.~Yang, ``Spatio-temporal feature
  encoding for traffic accident detection in vanet environment,'' \emph{IEEE
  Transactions on Intelligent Transportation Systems}, vol.~23, no.~10, pp.
  19\,772--19\,781, 2022.

\bibitem{9304822}
S.~Mylavarapu, M.~Sandhu, P.~Vijayan, K.~M. Krishna, B.~Ravindran, and
  A.~Namboodiri, ``Towards accurate vehicle behaviour classification with
  multi-relational graph convolutional networks,'' in \emph{Proceedings of IEEE
  Intelligent Vehicles Symposium (IV)}, 2020, pp. 321--327.

\bibitem{9341018}
------, ``Understanding dynamic scenes using graph convolution networks,'' in
  \emph{Proceedings of IEEE/RSJ International Conference on Intelligent Robots
  and Systems (IROS)}, 2020, pp. 8279--8286.

\bibitem{9197057}
C.~Li, Y.~Meng, S.~H. Chan, and Y.-T. Chen, ``Learning {3D}-aware egocentric
  spatial-temporal interaction via graph convolutional networks,'' in
  \emph{Proceedings of IEEE International Conference on Robotics and Automation
  (ICRA)}, 2020, pp. 8418--8424.

\bibitem{9672160}
A.~V. Malawade, S.-Y. Yu, B.~Hsu, D.~Muthirayan, P.~P. Khargonekar, and
  M.~A.~A. Faruque, ``Spatiotemporal scene-graph embedding for autonomous
  vehicle collision prediction,'' \emph{IEEE Internet of Things Journal},
  vol.~9, no.~12, pp. 9379--9388, 2022.

\bibitem{fang2022traffic}
J.~Fang, J.~Qiao, J.~Bai, H.~Yu, and J.~Xue, ``Traffic accident detection via
  self-supervised consistency learning in driving scenarios,'' \emph{IEEE
  Transactions on Intelligent Transportation Systems}, vol.~23, no.~7, pp.
  9601--9614, 2022.

\bibitem{10068772}
T.~Wang, K.~Chen, G.~Chen, B.~Li, Z.~Li, Z.~Liu, and C.~Jiang, ``Gsc: A graph
  and spatio-temporal continuity based framework for accident anticipation,''
  \emph{IEEE Transactions on Intelligent Vehicles}, pp. 1--13, 2023.

\bibitem{10030193}
K.~Doshi and Y.~Yilmaz, ``Towards interpretable video anomaly detection,'' in
  \emph{Proceedings of IEEE Winter Conference on Applications of Computer
  Vision (WACV)}, 2023, pp. 2654--2663.

\bibitem{9423525}
S.-Y. Yu, A.~V. Malawade, D.~Muthirayan, P.~P. Khargonekar, and M.~A.~A.
  Faruque, ``Scene-graph augmented data-driven risk assessment of autonomous
  vehicle decisions,'' \emph{IEEE Transactions on Intelligent Transportation
  Systems}, vol.~23, no.~7, pp. 7941--7951, 2022.

\bibitem{Lawal2017}
I.~Lawal, F.~Poiesi, D.~Anguita, and A.~Cavallaro, ``Support vector motion
  clustering,'' \emph{IEEE Transactions on Circuits and Systems for Video
  Technology}, vol.~27, no.~11, pp. 2395--2408, 2017.

\bibitem{ullah2021multi}
H.~Ullah, I.~U. Islam, M.~Ullah, M.~Afaq, S.~D. Khan, and J.~Iqbal,
  ``Multi-feature-based crowd video modeling for visual event detection,''
  \emph{Multimedia Systems}, vol.~27, pp. 589--597, 2021.

\bibitem{felemban2021deep}
E.~Felemban, S.~D. Khan, A.~Naseer, F.~Ur~Rehman, and S.~Basalamah, ``Deep
  trajectory classification model for congestion detection in human crowds.''
  \emph{Computers, Materials \& Continua}, vol.~68, no.~1, 2021.

\bibitem{gouiaa2021advances}
R.~Gouiaa, M.~A. Akhloufi, and M.~Shahbazi, ``Advances in convolution neural
  networks based crowd counting and density estimation,'' \emph{Big Data and
  Cognitive Computing}, vol.~5, no.~4, p.~50, 2021.

\bibitem{ptak2022board}
B.~Ptak, D.~Pieczy{\'n}ski, M.~Piechocki, and M.~Kraft, ``On-board crowd
  counting and density estimation using low altitude unmanned aerial
  vehicles—looking beyond beating the benchmark,'' \emph{Remote Sensing},
  vol.~14, no.~10, p. 2288, 2022.

\bibitem{Beery_2020_CVPR}
S.~Beery, G.~Wu, V.~Rathod, R.~Votel, and J.~Huang, ``Context r-cnn: Long term
  temporal context for per-camera object detection,'' in \emph{Proceedings of
  the IEEE Conference on Computer Vision and Pattern Recognition (CVPR)}, June
  2020.

\bibitem{wang2016avss}
Y.~Wang and A.~Cavallaro, ``Prioritized target tracking with active
  collaborative cameras,'' in \emph{Proceedings of IEEE International
  Conference on Advanced Video and Signal Based Surveillance (AVSS)}, 2016, pp.
  131--137.

\bibitem{wang2023you}
X.~Wang, J.~He, C.~Fu, T.~Meng, and M.~Huang, ``You only need two detectors to
  achieve multi-modal 3d multi-object tracking,'' \emph{arXiv preprint
  arXiv:2304.08709}, 2023.

\bibitem{khan2019congestion}
S.~D. Khan, ``Congestion detection in pedestrian crowds using oscillation in
  motion trajectories,'' \emph{Engineering Applications of Artificial
  Intelligence}, vol.~85, pp. 429--443, 2019.

\bibitem{basalamah2023deep}
S.~Basalamah, S.~D. Khan, E.~Felemban, A.~Naseer, and F.~U. Rehman, ``Deep
  learning framework for congestion detection at public places via learning
  from synthetic data,'' \emph{Journal of King Saud University-Computer and
  Information Sciences}, vol.~35, no.~1, pp. 102--114, 2023.

\bibitem{paszke2016enet}
A.~Paszke, A.~Chaurasia, S.~Kim, and E.~Culurciello, ``Enet: A deep neural
  network architecture for real-time semantic segmentation,'' \emph{arXiv
  preprint arXiv:1606.02147}, 2016.

\bibitem{mehta2018espnet}
S.~Mehta, M.~Rastegari, A.~Caspi, L.~Shapiro, and H.~Hajishirzi, ``Espnet:
  Efficient spatial pyramid of dilated convolutions for semantic
  segmentation,'' in \emph{Proceedings of the European Conference on Computer
  Vision (ECCV)}, 2018, pp. 552--568.

\bibitem{badrinarayanan2017segnet}
V.~Badrinarayanan, A.~Kendall, and R.~Cipolla, ``Segnet: A deep convolutional
  encoder-decoder architecture for image segmentation,'' \emph{IEEE
  Transactions on Pattern Analysis and Machine Intelligence}, vol.~39, no.~12,
  pp. 2481--2495, 2017.

\bibitem{zhou2017scene}
B.~Zhou, H.~Zhao, X.~Puig, S.~Fidler, A.~Barriuso, and A.~Torralba, ``Scene
  parsing through ade20k dataset,'' in \emph{Proceedings of the IEEE Conference
  on Computer Vision and Pattern Recognition (CVPR)}, 2017, pp. 633--641.

\bibitem{9578249}
G.~Di~Biase, H.~Blum, R.~Siegwart, and C.~Cadena, ``Pixel-wise anomaly
  detection in complex driving scenes,'' in \emph{Proceedings of IEEE
  Conference on Computer Vision and Pattern Recognition (CVPR)}, 2021, pp.
  16\,913--16\,922.

\bibitem{10030141}
T.~Vojíř and J.~Matas, ``Image-consistent detection of road anomalies as
  unpredictable patches,'' in \emph{Proceedings of the IEEE Winter Conference
  on Applications of Computer Vision (WACV)}, 2023, pp. 5480--5489.

\bibitem{yazdi2018new}
M.~Yazdi and T.~Bouwmans, ``New trends on moving object detection in video
  images captured by a moving camera: A survey,'' \emph{Computer Science
  Review}, vol.~28, pp. 157--177, 2018.

\bibitem{bookhartley2003multiple}
R.~Hartley and A.~Zisserman, \emph{Multiple view geometry in computer
  vision}.\hskip 1em plus 0.5em minus 0.4em\relax Cambridge university press,
  2003.

\bibitem{andrew2001multiple}
A.~M. Andrew, ``Multiple view geometry in computer vision,'' \emph{Kybernetes},
  vol.~30, no. 9/10, pp. 1333--1341, 2001.

\bibitem{lowe2004distinctive}
D.~G. Lowe, ``Distinctive image features from scale-invariant keypoints,''
  \emph{International Journal of Computer Vision}, vol.~60, no.~2, pp. 91--110,
  2004.

\bibitem{candes2011robust}
E.~J. Cand{\`e}s, X.~Li, Y.~Ma, and J.~Wright, ``Robust principal component
  analysis?'' \emph{Journal of the ACM (JACM)}, vol.~58, no.~3, pp. 1--37,
  2011.

\bibitem{bian2015bi}
X.~Bian and H.~Krim, ``Bi-sparsity pursuit for robust subspace recovery,'' in
  \emph{Proceedings of IEEE International Conference on Image Processing
  (ICIP)}, 2015, pp. 3535--3539.

\bibitem{peng2012rasl}
Y.~Peng, A.~Ganesh, J.~Wright, W.~Xu, and Y.~Ma, ``Rasl: Robust alignment by
  sparse and low-rank decomposition for linearly correlated images,''
  \emph{IEEE Transactions on Pattern Analysis and Machine Intelligence},
  vol.~34, no.~11, pp. 2233--2246, 2012.

\bibitem{placeCNN}
B.~Zhou, A.~Lapedriza, J.~Xiao, A.~Torralba, and A.~Oliva, ``Learning deep
  features for scene recognition using places database,'' in \emph{Advances in
  Neural Information Processing Systems}, vol.~27, 2014.

\bibitem{DCGAN}
I.~J. Goodfellow, J.~Shlens, and C.~Szegedy, ``Explaining and harnessing
  adversarial examples,'' \emph{arXiv preprint arXiv:1412.6572}, 2014.

\bibitem{ullah2021attention}
M.~Ullah, M.~Mudassar~Yamin, A.~Mohammed, S.~Daud~Khan, H.~Ullah, and
  F.~Alaya~Cheikh, ``Attention-based lstm network for action recognition in
  sports,'' \emph{Electronic Imaging}, vol.~33, no.~6, pp. 302--1, 2021.

\bibitem{ren2015faster}
S.~Ren, K.~He, R.~Girshick, and J.~Sun, ``Faster r-cnn: Towards real-time
  object detection with region proposal networks,'' \emph{Advances in neural
  information processing systems}, vol.~28, 2015.

\bibitem{he2017mask}
K.~He, G.~Gkioxari, P.~Doll{\'a}r, and R.~Girshick, ``Mask r-cnn,'' in
  \emph{Proceedings of the IEEE International Conference on Computer Vision
  (ICCV)}, 2017, pp. 2961--2969.

\bibitem{redmon2018yolov3}
J.~Redmon and A.~Farhadi, ``Yolov3: An incremental improvement,'' \emph{arXiv
  preprint arXiv:1804.02767}, 2018.

\bibitem{9706619}
K.~Doshi and Y.~Yilmaz, ``Rethinking video anomaly detection - a continual
  learning approach,'' in \emph{2022 IEEE Winter Conference on Applications of
  Computer Vision (WACV)}, 2022, pp. 3036--3045.

\bibitem{noghre2023understanding}
G.~A. Noghre, A.~D. Pazho, V.~Katariya, and H.~Tabkhi, ``Understanding the
  challenges and opportunities of pose-based anomaly detection,'' \emph{arXiv
  preprint arXiv:2303.05463}, 2023.

\bibitem{collarobot17}
B.~Hayes and J.~A. Shah, ``Interpretable models for fast activity recognition
  and anomaly explanation during collaborative robotics tasks,'' in
  \emph{Proceedings of IEEE International Conference on Robotics and Automation
  (ICRA)}, 2017, pp. 6586--6593.

\bibitem{7298873}
P.~Weinzaepfel, J.~Revaud, Z.~Harchaoui, and C.~Schmid, ``Learning to detect
  motion boundaries,'' in \emph{Proceedings of the IEEE Conference on Computer
  Vision and Pattern Recognition (CVPR)}, 2015, pp. 2578--2586.

\bibitem{luo2021normal}
W.~Luo, W.~Liu, and S.~Gao, ``Normal graph: Spatial temporal graph
  convolutional networks based prediction network for skeleton based video
  anomaly detection,'' \emph{Neurocomputing}, vol. 444, pp. 332--337, 2021.

\bibitem{Anjum2008}
N.~Anjum and A.~Cavallaro, ``Multifeature object trajectory clustering for
  video analysis,'' \emph{IEEE Transactions on Circuits and Systems for Video
  Technology}, vol.~18, no.~11, pp. 1555--1564, 2008.

\bibitem{khan2016analyzing}
S.~D. Khan, S.~Bandini, S.~Basalamah, and G.~Vizzari, ``Analyzing crowd
  behavior in naturalistic conditions: Identifying sources and sinks and
  characterizing main flows,'' \emph{Neurocomputing}, vol. 177, pp. 543--563,
  2016.

\bibitem{trajectory2017}
S.~Coşar, G.~Donatiello, V.~Bogorny, C.~Garate, L.~O. Alvares, and
  F.~Brémond, ``Toward abnormal trajectory and event detection in video
  surveillance,'' \emph{IEEE Transactions on Circuits and Systems for Video
  Technology}, vol.~27, no.~3, pp. 683--695, 2017.

\bibitem{wojke2017deepsort}
N.~Wojke, A.~Bewley, and D.~Paulus, ``Simple online and realtime tracking with
  a deep association metric,'' in \emph{Proceedings of IEEE International
  Conference on Image Processing (ICIP)}, 2017, pp. 3645--3649.

\bibitem{aghaei2021single}
M.~Aghaei, M.~Bustreo, Y.~Wang, G.~Bailo, P.~Morerio, and A.~Del~Bue, ``Single
  image human proxemics estimation for visual social distancing,'' in
  \emph{Proceedings of the IEEE Winter Conference on Applications of Computer
  Vision (WACV)}, 2021, pp. 2785--2795.

\bibitem{morerio2021icip}
P.~Morerio, M.~Bustreo, Y.~Wang, and A.~D. Bue, ``End-to-end pairwise human
  proxemics from uncalibrated single images,'' in \emph{Proceedings of IEEE
  International Conference on Image Processing (ICIP)}, 2021, pp. 3058--3062.

\bibitem{Poiesi2013}
F.~Poiesi, R.~Mazzon, and A.~Cavallaro, ``Multi-target tracking on confidence
  maps: An application to people tracking,'' \emph{Computer Vision and Image
  Understanding}, vol. 117, no.~10, pp. 1257--1272, 2013.

\bibitem{Poiesi2015a}
F.~Poiesi and A.~Cavallaro, ``Ieee transactions on circuits and systems for
  video technology,'' \emph{Computer Vision and Image Understanding}, vol.~25,
  no.~4, pp. 623--637, 2015.

\bibitem{GridNet}
I.~Bozcan, J.~Le~Fevre, H.~X. Pham, and E.~Kayacan, ``Gridnet: Image-agnostic
  conditional anomaly detection for indoor surveillance,'' \emph{IEEE Robotics
  and Automation Letters}, vol.~6, no.~2, pp. 1638--1645, 2021.

\bibitem{Riz2023}
L.~Riz, A.~Caraffa, M.~Bortolon, M.~L. Mekhalfi, D.~Boscaini, A.~Moura,
  J.~Antunes, A.~Dias, H.~Silva, A.~Leonidou \emph{et~al.}, ``The monet
  dataset: Multimodal drone thermal dataset recorded in rural scenarios,'' in
  \emph{Proceedings of the IEEE Conference on Computer Vision and Pattern
  Recognition (CVPR)}, 2023, pp. 2545--2553.

\bibitem{wu2020not}
P.~Wu, J.~Liu, Y.~Shi, Y.~Sun, F.~Shao, Z.~Wu, and Z.~Yang, ``Not only look,
  but also listen: Learning multimodal violence detection under weak
  supervision,'' in \emph{Proceedings of European Conference on Computer Vision
  (ECCV)}, 2020, pp. 322--339.

\bibitem{WANG2023103646}
X.~Wang and Z.~Zhu, ``Context understanding in computer vision: A survey,''
  \emph{Computer Vision and Image Understanding}, vol. 229, p. 103646, 2023.

\bibitem{Azizi2023}
S.~Azizi, S.~Kornblith, C.~Saharia, M.~Norouzi, and D.~J.~Fleet, ``Synthetic
  data from diffusion models improves imagenet classification,''
  \emph{arXiv:2304.08466}, 2023.

\bibitem{osman2023exploring}
A.~Osman~Tur, N.~Dall'Asen, C.~Beyan, and E.~Ricci, ``Exploring diffusion
  models for unsupervised video anomaly detection,'' \emph{arXiv e-prints}, pp.
  arXiv--2304, 2023.

\bibitem{clip2021}
A.~Radford, J.~W. Kim, C.~Hallacy, A.~Ramesh, G.~Goh, S.~Agarwal, G.~Sastry,
  A.~Askell, P.~Mishkin, J.~Clark \emph{et~al.}, ``Learning transferable visual
  models from natural language supervision,'' in \emph{International Conference
  on Machine Learning}, 2021, pp. 8748--8763.

\bibitem{reiss2022attribute}
T.~Reiss and Y.~Hoshen, ``Attribute-based representations for accurate and
  interpretable video anomaly detection,'' \emph{arXiv preprint
  arXiv:2212.00789}, 2022.

\bibitem{conti2023vocabulary}
A.~Conti, E.~Fini, M.~Mancini, P.~Rota, Y.~Wang, and E.~Ricci,
  ``Vocabulary-free image classification,'' \emph{arXiv preprint
  arXiv:2306.00917}, 2023.

\bibitem{lesort2020continual}
T.~Lesort, V.~Lomonaco, A.~Stoian, D.~Maltoni, D.~Filliat, and
  N.~D{\'\i}az-Rodr{\'\i}guez, ``Continual learning for robotics: Definition,
  framework, learning strategies, opportunities and challenges,''
  \emph{Information fusion}, vol.~58, pp. 52--68, 2020.

\bibitem{marvel2021}
D.~Bajovic, A.~Bakhtiarnia, G.~Bravos, A.~Brutti, F.~Burkhardt, D.~Cauchi,
  A.~Chazapis, C.~Cianco, N.~Dall’Asen, V.~Delic, C.~Dimou, D.~Djokic,
  A.~Escobar-Molero, L.~Esterle, F.~Eyben, E.~Farella, T.~Festi, A.~Geromitsos,
  G.~Giakoumakis, G.~Hatzivasilis, S.~Ioannidis, A.~Iosifidis, T.~Kallipolitou,
  G.~Kalogiannis, A.~Kiousi, D.~Kopanaki, M.~Marazakis, S.~Markopoulou,
  A.~Muscat, F.~Paissan, T.~P. Lobo, D.~Pavlovic, T.~P. Raptis, E.~Ricci,
  B.~Saez, F.~Sahito, K.~Scerri, B.~Schuller, N.~Simic, G.~Spanoudakis,
  A.~Tomasi, A.~Triantafyllopoulos, L.~Valerio, J.~Villazán, Y.~Wang,
  A.~Xuereb, and J.~Zammit, ``Marvel: Multimodal extreme scale data analytics
  for smart cities environments,'' in \emph{Proceedings of International Balkan
  Conference on Communications and Networking (BalkanCom)}, 2021, pp. 143--147.

\end{thebibliography}

\end{document}